\documentclass[acmtog, nonacm]{acmart}

\usepackage{wrapfig}
\usepackage{booktabs} %
\usepackage{subcaption}
\usepackage{makecell}
\usepackage{listings}
\usepackage{soul}
\usepackage{multirow}
\usepackage{xspace}

\definecolor{graytext}{RGB}{130,130,130}
\newcommand{\esc}[1]{{\color{violet}#1}}

\newcommand{\ykc}[1]{{\color{brown}[\textbf{YK:} #1]}}
\newcommand{\hec}[1]{{\color{teal}[\textbf{HE:} #1]}}
\newcommand{\esn}[1]{{\color{red}[\textbf{NOTE ES:} #1]}}

\newcommand{\ignorethis}[1]{}

\def\benchmarkName{\texttt{CompoundPrompts}} %
\def\agentName{XXXXXXXXXXX}
\def\benchmarkTotal{$540$}
\def\methodName{InstanceGen}

\def\pass{$\color{green}{\checkmark}$}
\def\fail{$\color{red}{\pmb{\mathsf{X}}}$}

\newbox\jsavebox
\newcommand{\jsubfig}[2]{%
	\sbox\jsavebox{#1}%
	\parbox[t]{\wd\jsavebox}{\centering\usebox\jsavebox\\#2}%
	}

\definecolor{count_color}{rgb}{1, 0.5, 0.0}
\definecolor{attribute_color}{rgb}{0.85, 0.0, 0.95}
\definecolor{spatial_color}{rgb}{0.3, 0.7, 1.0}

\newcommand{\counttext}[1]{\textbf{\textcolor{count_color}{#1}}}
\newcommand{\attrtext}[1]{\textbf{\textcolor{attribute_color}{#1}}}
\newcommand{\spatialtext}[1]{\textbf{\textcolor{spatial_color}{#1}}}

\copyrightyear{2025}
\acmYear{2025}
\setcopyright{cc}
\setcctype{by}
\acmConference[SIGGRAPH Conference Papers '25]{Special Interest Group on Computer Graphics and Interactive Techniques Conference Conference Papers }{August 10--14, 2025}{Vancouver, BC, Canada}
\acmBooktitle{Special Interest Group on Computer Graphics and Interactive Techniques Conference Conference Papers (SIGGRAPH Conference Papers '25), August 10--14, 2025, Vancouver, BC, Canada}
\acmDOI{10.1145/3721238.3730613}
\acmISBN{979-8-4007-1540-2/2025/08}

\begin{document}
\title{InstanceGen: Image Generation with Instance-level Instructions
}

\author{Etai Sella}
\affiliation{%
  \institution{Tel Aviv University}
  \city{Tel Aviv}
  \country{Israel}
}
\affiliation{%
  \institution{Meta AI}
  \city{London}
  \country{UK}
}
\email{etaisella@gmail.com}
\orcid{0009-0002-2119-9808}

\author{Yanir Kleiman}
\affiliation{%
  \institution{Meta AI}
  \city{London}
  \country{UK}
}
\email{yanirk@gmail.com}
\orcid{0009-0005-4459-6891}

\author{Hadar Averbuch-Elor}
\affiliation{%
  \institution{Cornell University}
  \city{New York}
  \state{New York}
  \country{USA}
}
\email{hadarelor@cornell.edu}
\orcid{0000-0003-3476-0940}

\begin{abstract}
Despite rapid advancements in the capabilities of generative models, pretrained text-to-image models still struggle in capturing the semantics conveyed by complex prompts that compound multiple objects and instance-level attributes.
Consequently, we are witnessing growing interests in integrating additional structural constraints, %
typically in the form of coarse bounding boxes, to better guide the generation process in such challenging cases.
In this work, we take the idea of structural guidance a step further by making the observation that contemporary image generation models can directly provide a plausible \emph{fine-grained} structural initialization. We propose a technique that couples this image-based structural guidance with LLM-based instance-level instructions, yielding output images that adhere to all parts of the text prompt, including object counts, instance-level attributes, and spatial relations between instances. 
Additionally, we contribute  
\benchmarkName, a benchmark composed of complex prompts with three difficulty levels in which object instances are progressively compounded with attribute descriptions and spatial relations. Extensive experiments demonstrate that our method significantly surpasses the performance of prior models, particularly over complex multi-object and multi-attribute use cases.

\end{abstract}

\ccsdesc[500]{Computing methodologies~Artificial intelligence}
\ccsdesc[500]{Computing methodologies~Computer vision}

\keywords{Diffusion Models, Image Generation}

\begin{teaserfigure}

\begin{center}

\begin{minipage}{0.98\linewidth}
	\centering

	\begin{minipage}{0.16\linewidth}
	    \centering
        \textbf{Ours}

	\end{minipage}
	\begin{minipage}{0.16\linewidth}
	    \centering
            Emu

	\end{minipage}
	\begin{minipage}{0.16\linewidth}
	    \centering
            Flux1-dev

	\end{minipage}
	\begin{minipage}{0.16\linewidth}
	    \centering
            SDXL

	\end{minipage}
	\begin{minipage}{0.16\linewidth}
	    \centering
            Bounded Attention

	\end{minipage}
	\begin{minipage}{0.16\linewidth}
	    \centering
            Attention Refocusing

	\end{minipage}

	\begin{minipage}[t]{0.16\linewidth}
	    \centering
	    \includegraphics[width=\linewidth]{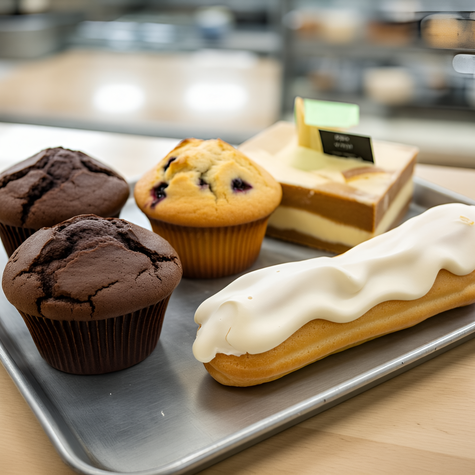}
	    
            \vspace*{-1mm}
            
            \small \counttext{$\blacksquare$} \attrtext{$\blacksquare$}
            
	    \vspace{1mm}
            
	\end{minipage}
	\begin{minipage}[t]{0.16\linewidth}
	    \centering
            \includegraphics[width=\linewidth]{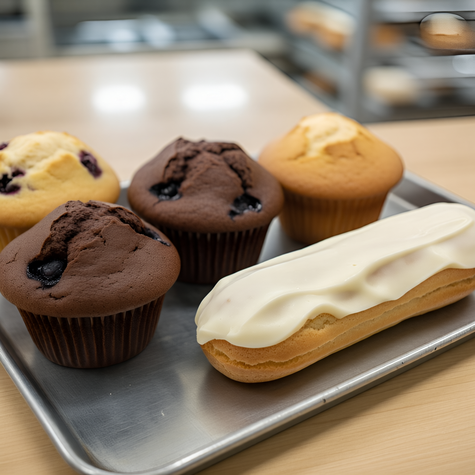}
	    
            \vspace*{-1mm}
            
            \small \counttext{$\square$} \attrtext{$\blacksquare$}
            
	    \vspace{1mm}
            
	\end{minipage}
	\begin{minipage}[t]{0.16\linewidth}
	    \centering
	    \includegraphics[width=\linewidth]{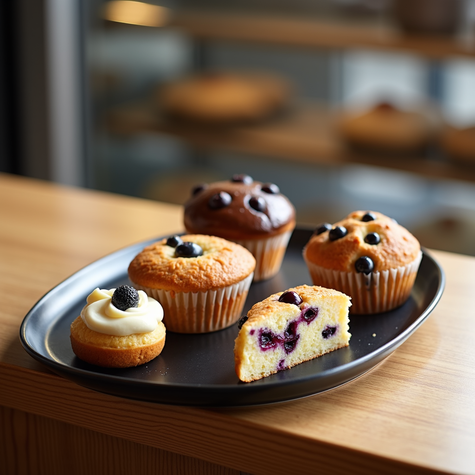}
	    
            \vspace*{-1mm}
            
            \small \counttext{$\square$} \attrtext{$\square$}
	    
            \vspace{1mm}

	\end{minipage}
	\begin{minipage}[t]{0.16\linewidth}
	    \centering
	    \includegraphics[width=\linewidth]{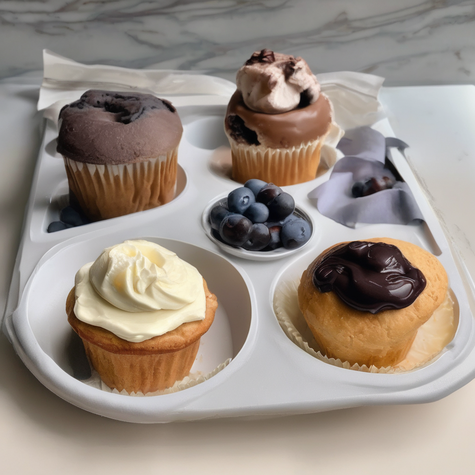}
	    
            \vspace*{-1mm}
            
            \small \counttext{$\square$} \attrtext{$\square$}
            
	    \vspace{1mm}

	\end{minipage}
	\begin{minipage}[t]{0.16\linewidth}

            \centering
	    \includegraphics[width=\linewidth]{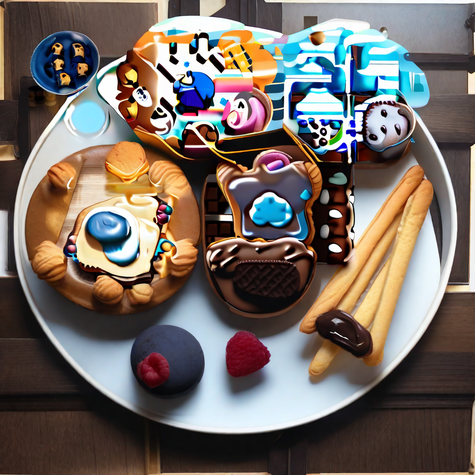}
	    
            \vspace*{-1mm}
            
            \small \counttext{$\square$} \attrtext{$\square$}
	    
            \vspace{1mm}

	\end{minipage}
	\begin{minipage}[t]{0.16\linewidth}

            \centering
	    \includegraphics[width=\linewidth]{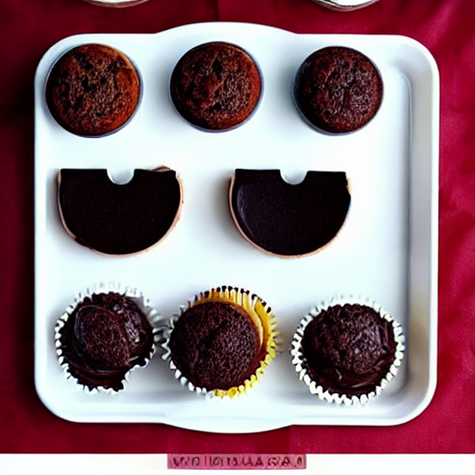}
	    
            \vspace*{-1mm}
            
            \small \counttext{$\square$} \attrtext{$\square$}
	    
            \vspace{1mm}

	\end{minipage}
    \normalsize{\emph{``a tray with \counttext{three} muffins, 
                    \counttext{a} vanilla éclair, 
                    and \counttext{a} piece of cake on it,
                    \attrtext{one} muffin is \attrtext{blueberry},
                    the other \attrtext{two} are \attrtext{chocolate}.''}}
    \vspace{3mm}

	\begin{minipage}{0.16\linewidth}
	    \centering
	    \includegraphics[width=\linewidth]{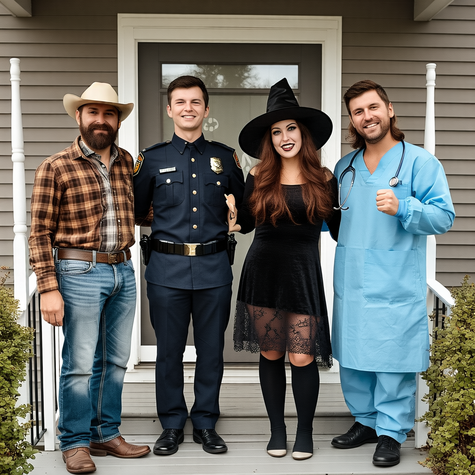}
	    
            \vspace*{-1mm}
            
            \small \counttext{$\blacksquare$} \attrtext{$\blacksquare$} \spatialtext{$\blacksquare$} 
            
	    \vspace{1mm}

	\end{minipage}
	\begin{minipage}{0.16\linewidth}
	    \centering
	    \includegraphics[width=\linewidth]{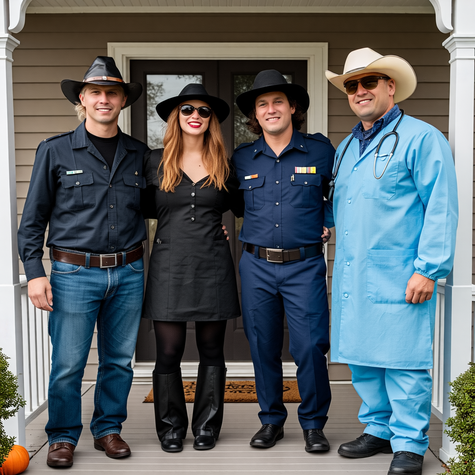}
	    
            \vspace*{-1mm}
            
            \small \counttext{$\blacksquare$} \attrtext{$\square$} \spatialtext{$\square$} 
            
	    \vspace{1mm}

	\end{minipage}
	\begin{minipage}{0.16\linewidth}
	    \centering
	    \includegraphics[width=\linewidth]{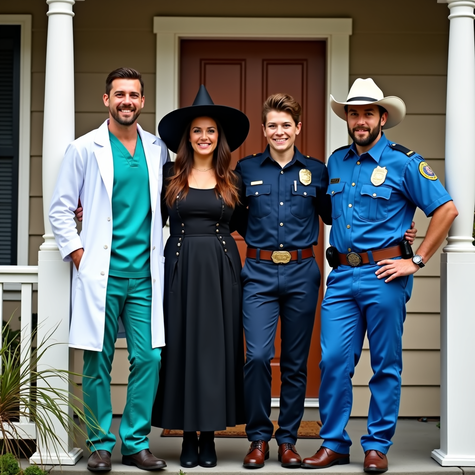}
	    
            \vspace*{-1mm}
            
            \small \counttext{$\blacksquare$} \attrtext{$\square$} \spatialtext{$\square$} 
            
	    \vspace{1mm}

	\end{minipage}
	\begin{minipage}{0.16\linewidth}
	    \centering
	    \includegraphics[width=\linewidth]{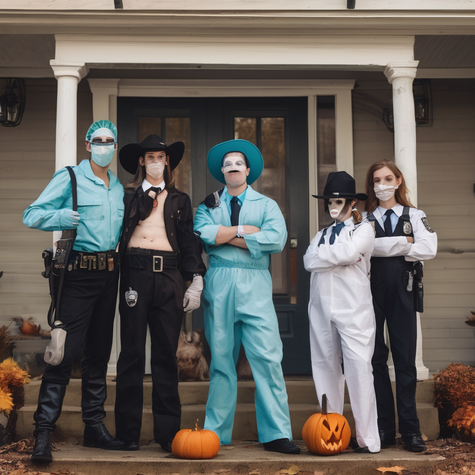}
	    
            \vspace*{-1mm}
            
            \small \counttext{$\square$} \attrtext{$\square$} \spatialtext{$\square$} 
            
	    \vspace{1mm}

	\end{minipage}
	\begin{minipage}{0.16\linewidth}

            \centering
	    \includegraphics[width=\linewidth]{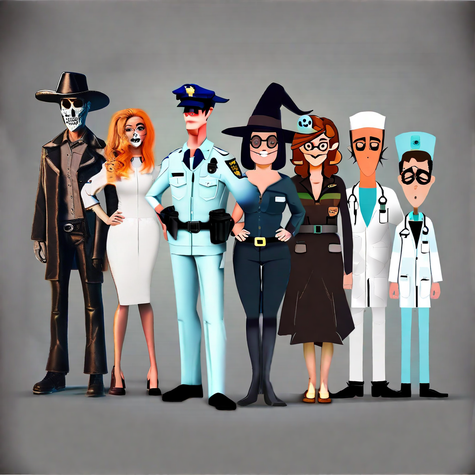}
	    
            \vspace*{-1mm}
            
            \small \counttext{$\square$} \attrtext{$\square$} \spatialtext{$\square$} 
            
	    \vspace{1mm}

	\end{minipage}
	\begin{minipage}{0.16\linewidth}

            \centering
	    \includegraphics[width=\linewidth]{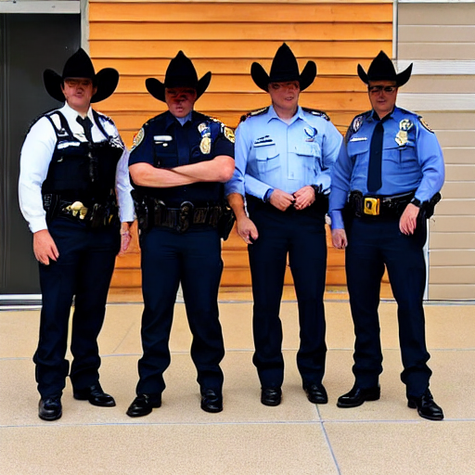}
	    
            \vspace*{-1mm}
            
            \small \counttext{$\blacksquare$} \attrtext{$\square$} \spatialtext{$\square$} 
            
	    \vspace{1mm}

	\end{minipage}

    \normalsize{\emph{``\counttext{four} friends dressed up for Halloween,
                    from \spatialtext{right to left} they are dressed as a \attrtext{surgeon},
                    a \attrtext{witch}, a \attrtext{police officer}, and a \attrtext{cowboy}.''}}
    \vspace{3mm}

	\begin{minipage}{0.16\linewidth}
	    \centering
	    \includegraphics[width=\linewidth]{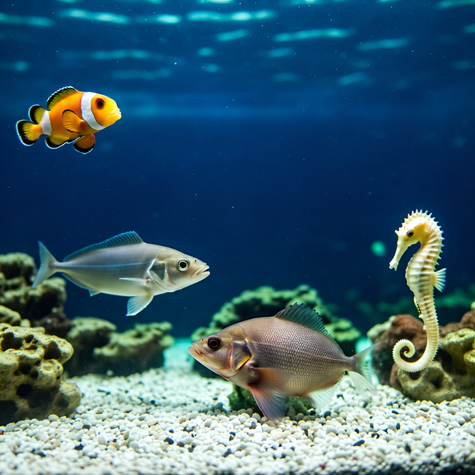}
	    
            \vspace*{-1mm}
            
            \small \counttext{$\blacksquare$} \attrtext{$\blacksquare$} \spatialtext{$\blacksquare$} 
            
	    \vspace{1mm}

	\end{minipage}
	\begin{minipage}{0.16\linewidth}
	    \centering
	    \includegraphics[width=\linewidth]{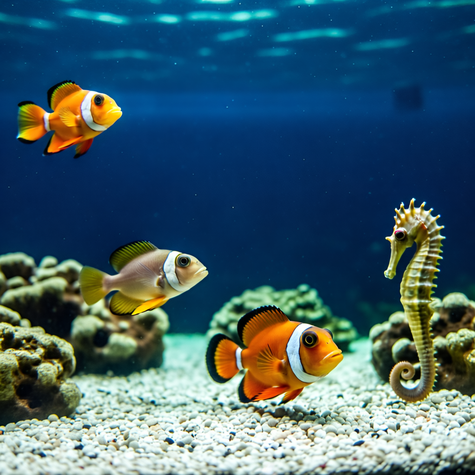}
	    
            \vspace*{-1mm}
            
            \small \counttext{$\blacksquare$} \attrtext{$\square$} \spatialtext{$\square$} 
            
	    \vspace{1mm}

	\end{minipage}
	\begin{minipage}{0.16\linewidth}
	    \centering
	    \includegraphics[width=\linewidth]{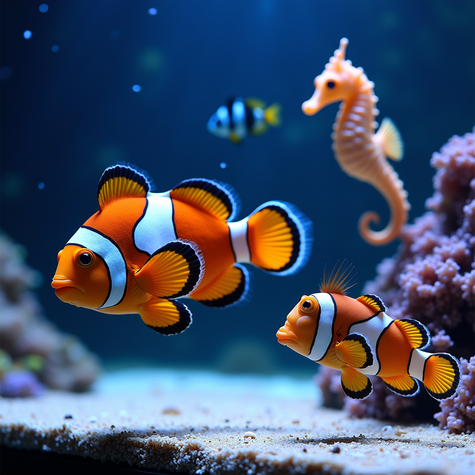}
	    
            \vspace*{-1mm}
            
            \small \counttext{$\blacksquare$} \attrtext{$\square$} \spatialtext{$\square$} 
            
	    \vspace{1mm}

	\end{minipage}
	\begin{minipage}{0.16\linewidth}
	    \centering
	    \includegraphics[width=\linewidth]{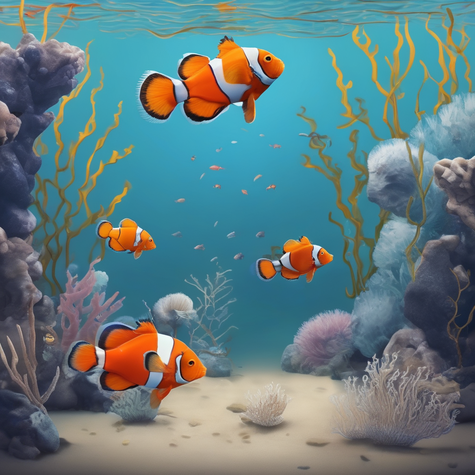}
	    
            \vspace*{-1mm}
            
            \small \counttext{$\square$} \attrtext{$\square$} \spatialtext{$\square$} 
            
	    \vspace{1mm}

	\end{minipage}
	\begin{minipage}{0.16\linewidth}

            \centering
	    \includegraphics[width=\linewidth]{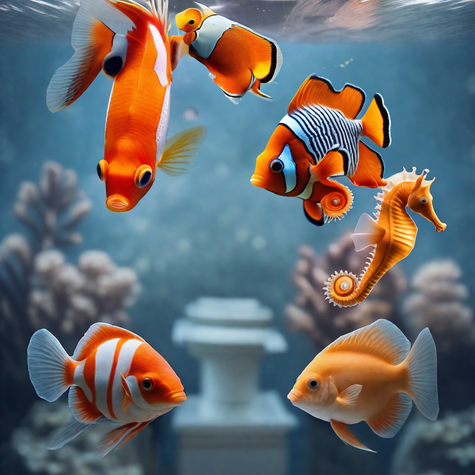}
	    
            \vspace*{-1mm}
            
            \small \counttext{$\square$} \attrtext{$\square$} \spatialtext{$\square$} 
            
	    \vspace{1mm}

	\end{minipage}
	\begin{minipage}{0.16\linewidth}

            \centering
	    \includegraphics[width=\linewidth]{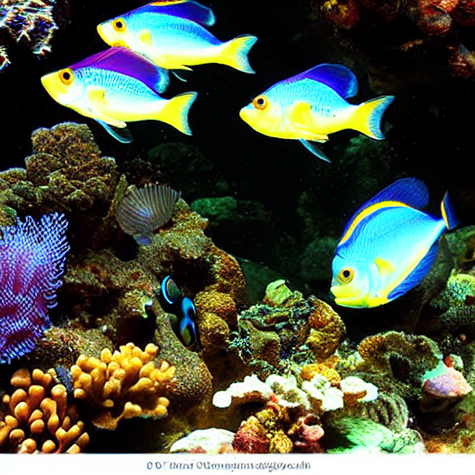}
	    
            \vspace*{-1mm}
            
            \small \counttext{$\square$} \attrtext{$\square$} \spatialtext{$\square$} 
            
	    \vspace{1mm}

	\end{minipage}
    \normalsize{\emph{``\counttext{three} fish and \counttext{a} seahorse in an aquarium, 
                    only the fish \spatialtext{closest to the surface} is a \attrtext{clown fish}.''}}

\end{minipage}

\end{center}
\caption{
        We present a method that incorporates instance-level instructions in image generation, capable of adhering to prompts with various
        \counttext{object counts~($\blacksquare$)}, 
        \attrtext{instance-level attributes~($\blacksquare$)}, 
        and \spatialtext{spatial relations~($\blacksquare$)}.
        Above we compare with several recent diffusion models (Emu~\cite{dai2023emuenhancingimagegeneration}, Flux~\cite{flux2023}, SDXL~\cite{podell2023sdxlimprovinglatentdiffusion}), as well as prior techniques that use bounding boxes to guide the generation (Bounded Attention~\cite{dahary2025yourself}, Attention Refocusing~\cite{phung2024grounded}). Filled squares represent adherence to the prompt, and empty squares represent a mismatch with the prompt.
   }
   \Description{A demonstration of our method's performance when compared against competing baselines.}
   \label{fig:teaser}

\end{teaserfigure}

\maketitle

\section{Introduction}
\label{sec:intro}

Text-to-image generation methods have shown continuous improvement in image quality and text fidelity, i.e., how closely the produced image follows the details described in the text prompt. However, current methods still struggle with text prompts which are composed of multiple objects with separate attributes associated with each object, including \emph{quantities} (e.g. "three cupcakes"), \emph{physical attributes} (e.g. "chocolate cupcake"), and \emph{spatial relations} (e.g. "chocolate cupcake on the right and vanilla cupcake on the left").
Examples of typical failure cases for current methods are shown in Figure \ref{fig:teaser}. \ignorethis{\esc{These include counting errors (too many fish in second image from the right on bottom row), \textit{semantic leakage} (surgeon wearing cowboy hat in second image from the left in the middle row) or \textit{catastrophic neglect}, where the model fails to generate one or more of the subjects from the input prompt (no cake in second image from the left on the top row).}\ignorethis{\hec{Figure, not table}}.} 
The failure is often due to a mismatch between each object instance and its associated attributes, where an object instance is either missing a specified attribute, erroneously associated with attributes of different objects, or has the wrong count.

To address these type of failures, existing work typically employs layout-based generation, where the coarse layout is either provided by the user~\cite{dahary2025yourself, chen2024training, xie2023boxdiff}, or generated as bounding boxes by an LLM~\cite{chen2023reason, phung2024grounded}. However, bounding box layouts generated this way tend to be inaccurate and unrealistic, providing a relatively weak structural signal %
to the image generation model for generating an image that adequately follows the complex multi-object prompt. 
A natural question therefore arises: Can one devise a text-to-image generation technique that utilizes a more fine-grained structural guidance, \emph{without} requiring any additional inputs?

In this work, we answer this question affirmatively by making the observation that current image generation models can provide a rough initialization which is more precise and realistic than bounding boxes, and can be used as an anchor for generating an image which faithfully adhere to the target prompt. We present an inference-time text-to-image generation technique that couples this image-based structural guidance with instance-level instructions extracted using Large Language Models (LLMs), leveraging the relative strengths of image-based and text-based techniques for producing fine-grained layout augmented with instance-level instructions. For example, the instance-level instructions may indicate that a certain segment should contain a chocolate cupcake while another segment should contain a vanilla cupcake with sprinkles. %
Moreover, we propose an attention-based optimization technique guided by our fine-grained structural layout, yielding a final generated image that accurately represents the text prompt.

Existing benchmarks for evaluating text fidelity~\cite{saharia2022photorealistic, wu2024conceptmix, bakr2023hrs} are often too simplified or focused on out-of-distribution surreal scenes.
Conversely, little attention 
is
directed towards more complex prompts with multiple instance-level instructions. 
Therefore, to facilitate a quantitative evaluation, we introduce a benchmark of \benchmarkTotal{} prompts with instance-level instructions which combine objects of different quantities, semantic attributes, and spatial relations. Our benchmark is divided to three complexity levels: object counts, instance-level attributes, and spatial relations between instances. 
We evaluate our method on the new benchmark and existing benchmarks~\cite{saharia2022photorealistic,ghosh2024geneval}.
The evaluation shows that our method achieves stronger performance across multiple metrics quantifying to what extent the generated image conveys the target text prompt.

In summary, our contributions in this paper are:
\begin{itemize}
    \item 
    A text-to-image generation method which generates a layout with per-segment instance-level instructions  from a text prompt, and conditions the image generation on the generated layout, with no additional training.
    \item 
    A benchmark for evaluating prompts of three complexity levels with multiple objects and instance-level attributes.
    \item 
    State-of-the-art results that show our method's ability to tackle particularly challenging use cases which are not well addressed by current methods.
\end{itemize}

\ignorethis{
\begin{table*}
    \centering
    \setlength{\tabcolsep}{0.01\textwidth}
    \begin{tabular}{p{\dimexpr 0.5\linewidth-2\tabcolsep} 
                    p{\dimexpr 0.5\linewidth-2\tabcolsep}}
        \toprule
        Prompt & Typical errors \\
        \midrule
        ``a box of crayons with five crayons in total, their colors are red,  & Incorrect number of crayons \\
        orange, yellow, green, and blue, from left to right'' & Incorrect colors of crayons \\
        & Incorrect order of crayons \\
        \midrule
        ``five eggs in a carton, two of which are broken'' & Incorrect total number of eggs \\
        & Incorrect number of broken eggs \\
        & Broken eggs are outside of the carton \\
        \midrule
        ``four action figures of batman, superman, spiderman, and wolverine'' & Incorrect number of action figures \\
          & Two action figures of the same character \\
          & Characters with a mixture of two superheroes costume elements \\
        \bottomrule
    \end{tabular}
    \captionof{table}{
    Typical failure cases of image generation methods on multi-category. \esn{This should be a figure with images right?}
    }
    \label{table:prompts}
\end{table*}
}

\begin{figure*}
    \centering
    \includegraphics[width=0.99\linewidth]{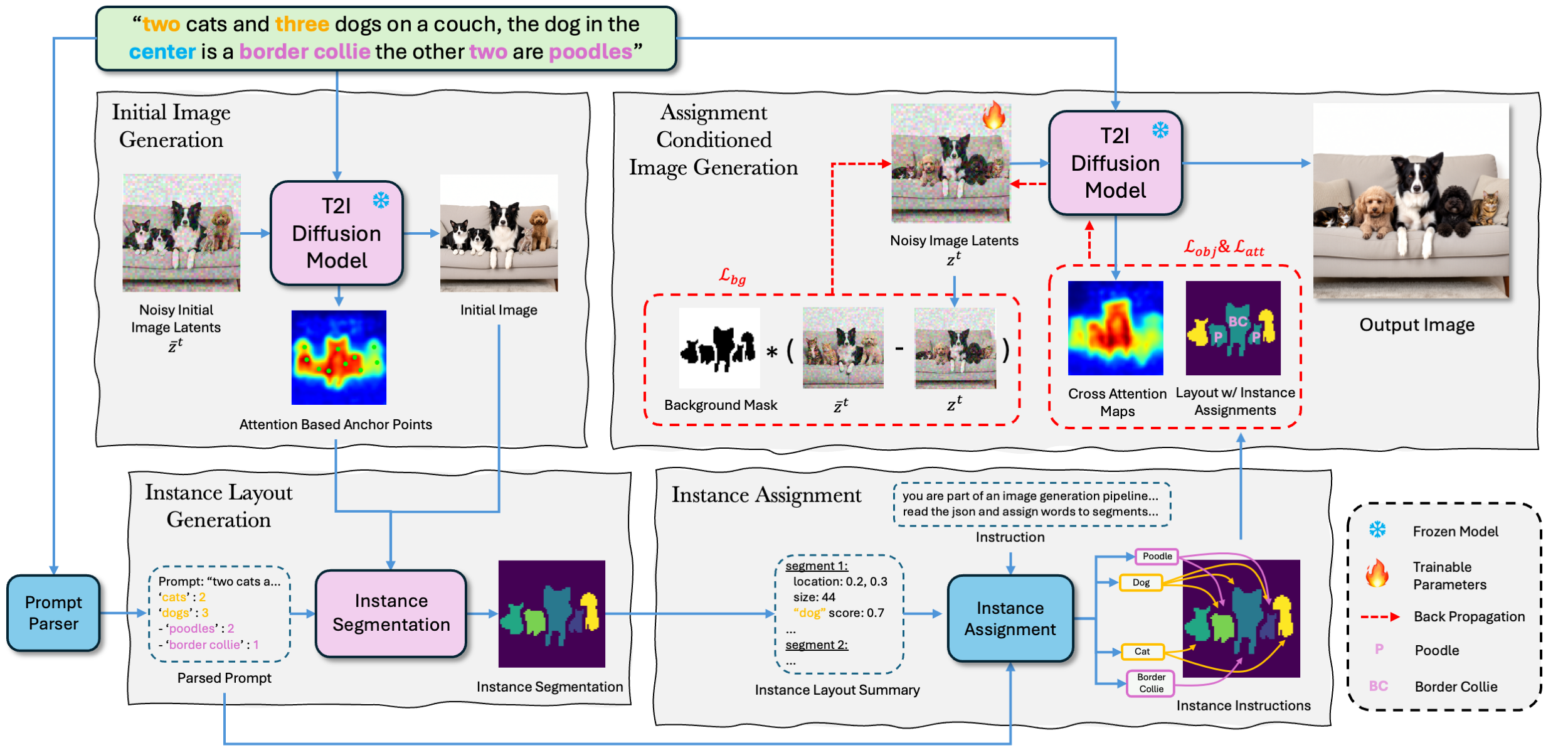}
    \caption{\textbf{Method Overview.} Given a text prompt (possibly) containing \counttext{object counts}, \attrtext{instance-level attributes}, 
        and \spatialtext{spatial relations}, our approach combines image-based and text-based components (visualized in pink and blue boxes above) for generating an output image (illustrated on the right). We first generate an initial image using a pretrained text-to-image diffusion model. Given the image and attention information from the initial diffusion process, we extract an instance segmentation layout, and assign object and instance-level instructions using an LLM. We then use the layout and instructions, as well as the initial image's latents, to incur losses, mask the attention, and optimize the latent in the assignment conditioned image generation stage which produces the output image. Note that for simplicity attention masking is not displayed in the figure.
    }
    \label{fig:method_overview}
\end{figure*}

\section{Related Works}
\label{sec:related_works}
\subsection{Text to Image Generation}
\ignorethis{Text-to-image generation has seen remarkable advancement in recent years. Several key technical innovations have driven this progress. The introduction of Generative Adversarial Networks (GANs) \cite{goodfellow2014generativeadversarialnetworks}, which employ a discriminator network to provide adversarial feedback to the generator, catalyzed numerous influential works \cite{arjovsky2017wasserstein, mirza2014conditional, brock2018large}. Subsequently, autoregressive approaches that treat images as sequences of visual tokens enabled models like DALL-E \cite{ramesh2021zero} to generate images by predicting these tokens one at a time \cite{chang2023muse, ding2022cogview2, lee2022autoregressive}. Most recently, diffusion models \cite{ho2020denoising}, which gradually denoise random Gaussian noise into coherent images through an iterative process, have dramatically improved output quality.} 
The emergence of diffusion models \cite{ho2020denoising} has been a huge leap forward for text-to-image generation, setting new standards in terms of both visual quality and text fidelity \cite{dhariwal2021diffusion}. These models generate samples by gradually denoising random Gaussian noise into coherent images through an iterative process, optionally incorporating encoded text via an attention mechanism \cite{vaswani2017attention} to guide it. Advanced variants such as Latent Diffusion Models (LDMs) \cite{rombach2022high} and Flow Matching \cite{lipman2022flow} approaches are now considered state-of-the-art \cite{podell2023sdxlimprovinglatentdiffusion, esser2024scaling}, with notable high performance models such as SDXL \cite{podell2023sdxlimprovinglatentdiffusion}, Flux-1~\cite{flux2023} and Emu \cite{dai2023emuenhancingimagegeneration}. However, despite improvements in image quality, current models still struggle with following complex, detailed prompts, a limitation largely attributed to the difficulty of acquiring large-scale, high-quality ground truth data for such specific prompts. Our work specifically addresses this challenge.

\ignorethis{
\esn{modified the above, too short?}
\ykc{Comments on the above:
1. No way that GANs were introduced in 2020. I think it should be 2014 :)
2. Do we really need to go this far back?? I would just start with the diffusion models and go from there. We can say a few more words on recent work, but having a shorter paragraph is also fine.
3. Please mention the models by name (e.g. SD, SDXL) instead of just providing the citation, at least for the well-known models.
4. Add Emu of course: "Emu: Enhancing image generation models using photogenic needles in a haystack", can also mention "Emu edit: Precise image editing via recognition and generation tasks"
5. For Flux / Mistral, we can mention it as a commercial tool and add a citation which is a url to their website.}
\esn{got it will change!}
}

\subsection{Controllability in Text-to-Image Generation}
Controllability in text-to-image generation is crucial, as users fundamentally seek to accurately materialize their creative visions. To enhance control, numerous approaches have introduced additional forms of guidance beyond text. Methods like ControlNet \cite{zhang2023adding}, GLIGEN \cite{li2023gligen} and InstanceDiffusion \cite{wang2024instancediffusion} enable various guidance types including bounding boxes, segmentation masks, and pose information, 
while DDPO \cite{black2023training} and DPO \cite{wallace2024diffusion} improve text fidelity by using reinforcement learning with MLLM or human feedback.
However, these approaches require specialized training and are not easily transferable across models. 

MultiDiffusion \cite{bar2023multidiffusion} offers an alternative by accepting segmentation masks and aspect ratio information without requiring training, instead incorporating an optimization task into the diffusion sampling procedure to coordinate multiple generation processes. Similarly, other training-free approaches \cite{chen2024training, dahary2025yourself, xie2023boxdiff} ground specific objects to image regions by manipulating attention values during the reverse diffusion process using bounding boxes. These methods all require explicit user input beyond text, reducing their intuitiveness. Some works \cite{chen2023reason, phung2024grounded} address this by using 
LLMs
to generate bounding box layouts from prompts, which then guide attention during inference. However, these LLM-generated layouts often prove unrealistic, and the ambiguity inherent in bounding box representations compromises accuracy.

Attend and Excite \cite{chefer2023attend} improves controllability by preventing attribute leakage between attention values of specific text tokens during inference, requiring no additional input but struggling with accurate object counting. CountGen \cite{binyamin2024make} specifically targets counting capabilities by extracting and modifying layouts from initial images to match prompt requirements, but is limited in handling multiple object types or attributes and relies heavily on model-specific layout extraction. 
In contrast,
our method leverages the base text-to-image model to extract realistic guidance, requiring no additional user input or training while handling complex prompts involving multiple objects, instance-level attributes, and spatial relationships.

Similar to our approach, SLD~\cite{wu2024self} derives corrected layouts by using an LLM and a segmentation model, and blend missing instances which are generated separately into the initial image. This approach improves object counting fidelity, but has limited ability to generate fine grained instance attributes due to its reliance on image segmentation models and separation between background and foreground generation.

\section{Method}
\label{sec:method}
In this work, we consider the problem of generating images with multiple objects, instance level attributes and spatial arrangements, with the goal of producing visually pleasing and highly realistic images that also have high text fidelity. To this aim, we generate an initial 
image via a text-to-image diffusion model.
The produced image typically has high visual quality and plausible composition, but may significantly diverge from the provided prompt. We leverage the attention maps extracted during the image generation process to segment the diffused image and produce an initial layout image (Sec.~\ref{sec:instance_segmentation}). We then provide this layout in textual form as input to an LLM,
which assigns instance-level instructions or attributes to each segment in the layout (Sec.~\ref{sec:instance_assignment}). Finally, the layout coupled with the instance-level assignments are used as guidance in our conditional image generation process (Sec.~\ref{sec:conditioned_image_generation}). Figure \ref{fig:method_overview} provides an overview of our approach. 

\subsection{Instance Layout Generation}
\label{sec:instance_layout_generation}

\subsubsection{Prompt Parsing}
\label{sec:prompt_breakdown}
As a precursor to our process, we break down the prompt into its core components, which include object-level quantities, instance-level attributes, and instance-level quantities. 
For example, the prompt ``three people and two dogs posing for a photo, one person is waving'' contains two objects, ``people'' and ``dogs'', with desired quantities of three and two respectively. The object ``people'' is associated with an instance-level attribute ``is waving'', which has a desired quantity of one, whereas the object ``dogs'' is not associated with any instance-level attributes.
To extract this information, we instruct an LLM (Llama 3.3~\cite{dubey2024llama}) to analyze the prompt and return this information in the form of a \texttt{json} format dictionary. An example prompt with its parsing output is provided in the supplementary material.

\subsubsection{Instance Segmentation}
\label{sec:instance_segmentation}
To guide our generation process, we begin by generating an initial image using a pretrained
text-to-image diffusion model.
The image generation model also produces cross-attention maps between the prompt and the image, which we use to guide a fine-grained segmentation of the initial image. 
We combine the signals from the cross-attention maps with off-the-shelf segmentation models applied to the image itself. This is in contrast with prior works that segment objects using attention information exclusively~\cite{patashnik2023localizing,namekata2024emerdiff,binyamin2024make}, an approach that makes it challenging to segment multiple instances that correspond to the same object token, and is prone to over-segmenting instances to semantic parts.
To guide the image segmentation models, we extract \emph{anchor points} in the form of local maxima of the aggregated attention maps, which indicate which segments correspond to specific object instances. We observe empirically that every object instance produces at least one such anchor point. Thus, we discard output segments that do not include anchor points, and we segment the image further as long as there are unassigned anchor points, using the anchor points as coordinate inputs. 
For the initial segmentation of instances, we use Mask R-CNN~\cite{he2017mask}, which tends to segment complete instances but misses some instances due to its limited vocabulary. Hence, we merge Mask-RCNN's instance segmentation with SAM2~\cite{kirillov2023segment} keypoint-based segmentations. While SAM2 is prone to over-segmentation of visually distinct parts, by combining it with the output of Mask-RCNN we can avoid partial segmentation while also ensuring all instances are accounted for, regardless of their type.
We provide additional details regarding this stage in the supplementary material.

\noindent \textbf{Attention Scores.} \quad
To allow our LLM agent to make logical instance assignments downstream, we pair each segment in our layout with an attention score for each object word and instance attribute. The attention scores for a given segment are calculated by averaging the cross attention map pixel values of object or instance attribute words over all pixels that belong to the segment.
The segments sizes, positions, and attention scores are written in \texttt{json} format and used as input for the instance assignment phase.

\noindent \textbf{Robust Initialization via Seed Search.}  \quad
Once we have an initial instance segmentation, we ensure that we have enough instances to accurately convey the desired prompt. For example if the initial layout contains four segments and our prompt requires six objects then the layout can not be used to accurately convey it. 
We first attempt to generate an initial image that contains enough instance segments using a different seed, and repeat this process up to five times. If a viable layout is not obtained after five attempts, we choose existing instance masks at random and copy them to a random location in the background.

\begin{figure}
    \setlength{\tabcolsep}{1pt}
    \begin{tabular}{cccc}
    \centering
    \includegraphics[width=0.24\linewidth]{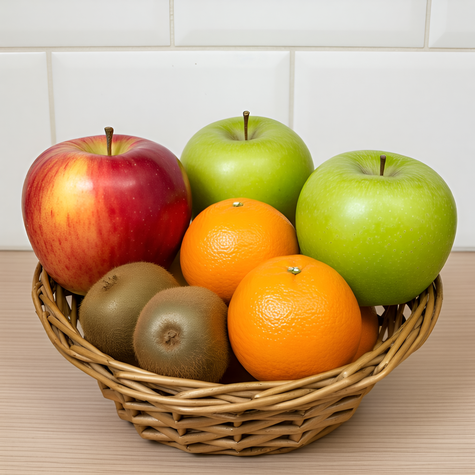}
    &
    \includegraphics[width=0.24\linewidth]{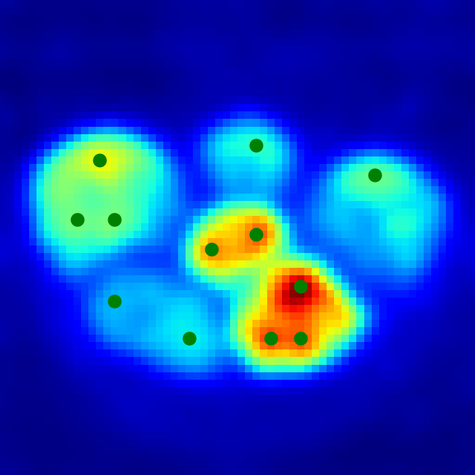}
    &
    \includegraphics[width=0.24\linewidth]{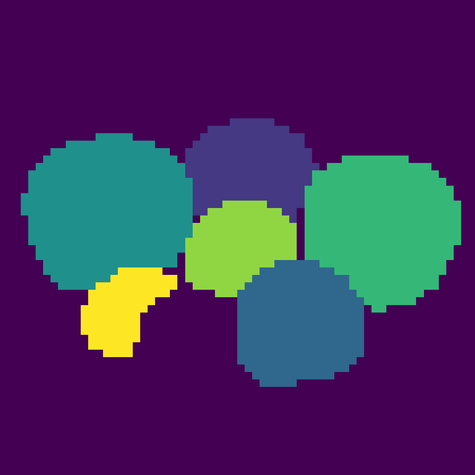}
    &
    \includegraphics[width=0.24\linewidth]{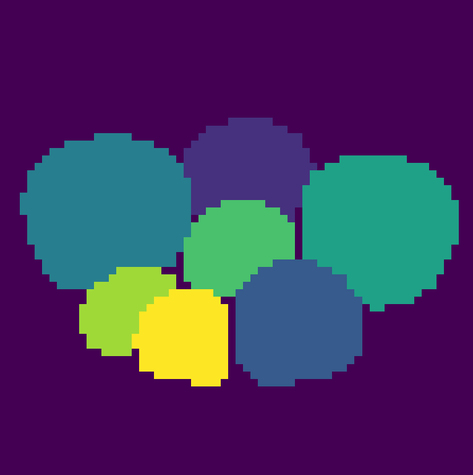}
    \\
    \includegraphics[width=0.24\linewidth]{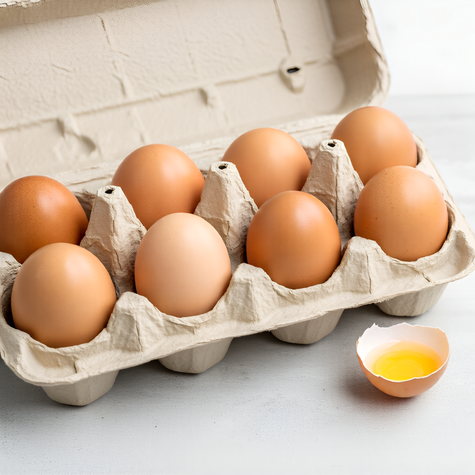}
    &
    \includegraphics[width=0.24\linewidth]{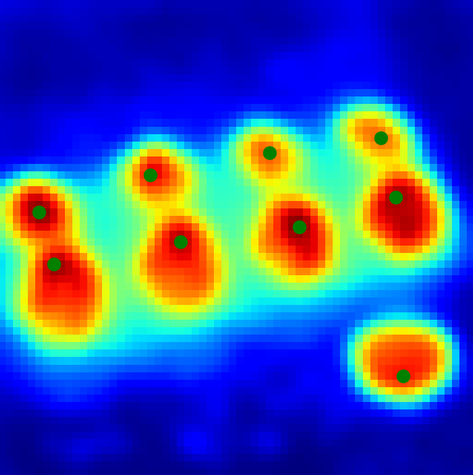}
    &
    \includegraphics[width=0.24\linewidth]{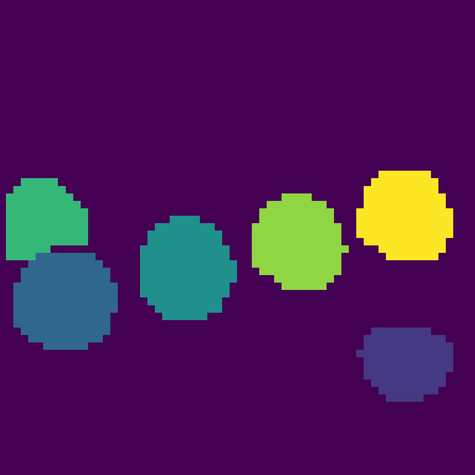}
    &
    \includegraphics[width=0.24\linewidth]{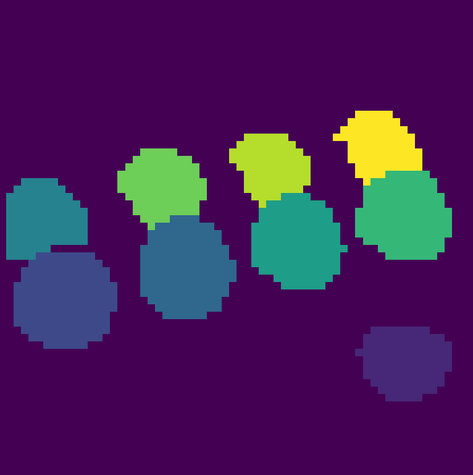}
    \\

    \begin{minipage}{0.24\linewidth}
        \footnotesize{(a) Initial Image}
    \end{minipage}
    &
    \begin{minipage}{0.24\linewidth}
        \footnotesize{(b) Cross Att. + Anchor Points}
    \end{minipage}
    &
    \begin{minipage}{0.24\linewidth}
        \footnotesize{(c) Initial Segmentation}
    \end{minipage}
    &
    \begin{minipage}{0.24\linewidth}
        \footnotesize{(d) Final Segmentation}
    \end{minipage}
    \end{tabular}
    \caption{\textbf{%
    Instance Layout Generation.} (a) The initial image generated by a pretrained text-to-image model. (b) The aggregated cross attention for all foreground words, with anchor points marked by green dots. (c) Initial segmentation after discarding segments without anchor points. (d) Final segmentation after adding segments for each unassigned anchor point.}
    \label{fig:layout}
\end{figure}

\subsection{Instance Assignment \ignorethis{via \textit{\agentName{}}}}
\label{sec:instance_assignment}
The instance assignment LLM takes as input the prompt, the parsed \texttt{json} dictionary (Sec.~\ref{sec:prompt_breakdown}), and the instance segmentation \texttt{json} (Sec.~\ref{sec:instance_segmentation}), and tasked with assigning each segment with instance-level instructions.
This can either be the \textit{delete} instruction, implying the segment should not be part of the foreground of the final image, or one of the desired object words, for example ``dogs'' or ``people''.
Additionally, the LLM optionally assigns one or more instance level attributes to each segment, for example ``is waving''.

In alignment with our overall goal, we instruct the LLM 
with
a primary and secondary objective. The primary objective is to output a layout which correctly depict the input prompt.
The secondary objective is to assign the layout in a way that results in a minimal visual difference between the output and initial image, since the initial image is likely to have a plausible composition and high visual quality, even if it is inaccurate in terms of prompt fidelity.
In practice, we instruct the LLM to first make assignments on the object level, ensuring spatial relationships and object counts are accurately addressed. Once these assignments are met, the LLM is instructed to follow a similar procedure for instance level attributes. When not bounded by specific constraints, such as a specific spatial arrangement, the LLM is instructed to assign each object or instance-level attribute to the segment which maximizes its attention score until the desired amount is met, such that these assignments agree most with the initial image. 

\ignorethis{
assign for a certain segment the words with maximal attention scores to each segment as we assume these assignments agree most with the initial image. \ykc{I don't think this is very clear - what do we do when we have X objects out of Y with some instance-level attribute? We could check the scores of all attributes for each segment but then we wouldn't necessarily end up with X of those (we can stop after we reach X instances). We can search for the segment with maximal score, then second highest, etc, until we have X. It is not clear which of these approaches is taken here.}
\ykc{Suggestions for phrasing the second option: the LLM is instructed to find the segment with the maximal attention score for the attributes of each desired instance, as we assume...}
}

The instruction prompt given to the LLM as input also contains a number of curated in-context examples depicting typical scenarios. \ignorethis{\esn{Expend on this a bit and emphasize robustness - After internal release}}
If we detect an error in the assignment output, for example if the assignment includes a wrong number of objects, or an attribute was assigned to a segment marked for deletion, we call the LLM again with the erroneous assignment appended to the instruction prompt as a negative example. 
We find that this process is less error-prone when the LLM is instructed to output its reasoning for the assignment as well as the \texttt{json} format output. However we do not use the reasoning part of the output in our pipeline.
An overview of the inputs given to the LLM in this stage is shown in Figure \ref{fig:segmentbrain}, and a detailed example including the LLM output is provided in the supplementary material.

\begin{figure}
    \centering
    \includegraphics[width=1\linewidth]{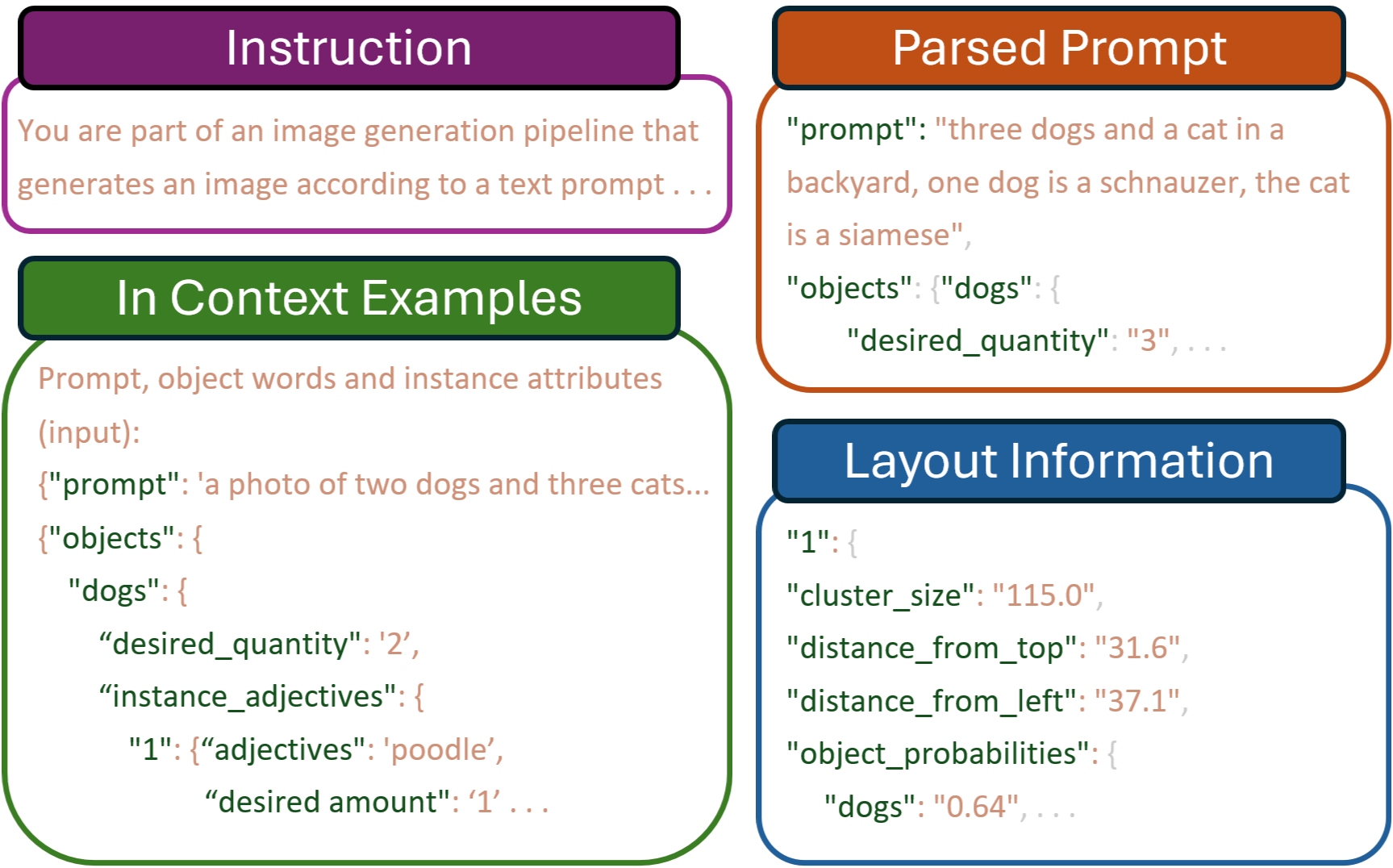}
    \caption{\textbf{Instance assignment inputs.}
    The input given to the LLM for instance assignment contains in-context examples, the parsed prompt, and the visual layout of each segment.}
    \label{fig:segmentbrain}
\end{figure}

\subsection{Assignment Conditioned Image Generation}
\label{sec:conditioned_image_generation}
Once assignments have been made, we omit the segments marked for deletion and are left with a set of binary masks, each associated with one of the object words and optionally one or more instance attribute word. %
The goal of this stage is to generate an image in which each assigned region 
accurately depicts its associated instructions, while maintaining high visual quality and plausibility. Inspired by prior work ~\cite{chefer2023attend, dahary2025yourself, binyamin2024make}, we approach this task by modifying the reverse diffusion process of the image generator in a manner that encourages appropriate attention values to correlate with their associated segments. Specifically, we modify the image generation process as follows.

\subsubsection{Cross Attention Losses} During inference, we optimize the intermediate image latents using the following attention based losses:

 \noindent \textbf{Object Attention Loss.}  \quad
Inspired by \cite{binyamin2024make} we incorporate a weighted binary cross entropy loss between cross attention values for an object word $C_w$ and a segment $m$ associated with the word:
\begin{equation}
    \mathcal{L}_{obj} = -\sum_i \lambda_i (m_i \cdot log(C_{w,i}) + (1-m_i)\cdot log(1-C_{w,i}))
\end{equation}
where is $C_{w,i}$ the cross-attention map value for word $w$ at pixel $i$, $m_i$ is the value of the binary mask $m$ at pixel $i$, and $\lambda_i$ is the weight assigned to each pixel $i$ where $\lambda_i = 1.5$ if $m_i = 1$, otherwise $\lambda_i = 1$. We set the weight $\lambda_i$ in this manner to better encourage object generation in the foreground. \ignorethis{\ykc{Please explain why we weight the negative examples differently in a short sentence. If this is a parameter of the method maybe better to just write "weight assigned to each pixel" and specify the 1.5 / 1 values in the appendix? Also I think weights are more commonly in lower case $w_i$} This loss encourages appropriate objects to be present in their assigned segments and discourages them from appearing elsewhere.
}

 \noindent \textbf{Attribute Attention Loss.}  \quad
While it is possible to treat attribute words in the same manner we treat object words and use the aforementioned \textit{Object Attention loss} to restrict them to their assigned segments, we find that this does not affect object creation in the background and often has an adverse affect on the overall image quality. As such we instead use a simple cross entropy loss to encourage attribute appearance in assigned segments:
\begin{equation}
    \mathcal{L}_{att} = -\sum_i m_i log(C_{w,i})
\end{equation}
where $C_{w,i}$ is the cross attention for attribute word $w$ at pixel $i$.
\ignorethis{
\ykc{
1. I would use the same parameter $w$ as the previous equation for the attribute word.
2. It is essentially the same as providing a different weight for attributes, right? Can we generalize both losses to the same equation with different weight parameters? 
}
}
\subsubsection{Attention Masking} As an additional measure for preventing semantic leakage between objects and attributes we \textit{mask} the attention of object words and attributes in segments that they are \textbf{not} assigned to:
\begin{equation}
    C_{w,i} = \begin{cases}
        \delta C_{w,i} & m_i = 1 \\
        C_{w,i} & m_i = 0 \\
    \end{cases}
\end{equation}
where $C_{w,i}$ is the cross attention value of object or attribute word $w$ at index $i$ that is \textbf{not} associated with mask $m$. $\delta$ is a hyper-parameter set to $\delta = -1.5$ in our implementation.
Additionally to further confine objects to the foreground and prevent objects from appearing in the background we use the self attention masking procedure used in \cite{binyamin2024make}:
\begin{equation}
    S_{i, j} = \begin{cases}
        0 & i \in \mathcal{F}\ and\ j \in \mathcal{B} \\
        0 & i \in \mathcal{B}\ and\ j \in \mathcal{F} \\
       S_{i, j} & otherwise
    \end{cases}
\end{equation}
where $S$ is a self attention map, $i$ and $j$ pixel indices and $\mathcal{F}$ and $\mathcal{B}$ is the foreground and background masks, defined as $\mathcal{F}=\bigcup_{i=0}^{|M|}m_i$ and $\mathcal{B}=\overline{\mathcal{F}}$. 

\subsubsection{Composition Preserving Regularization} Leveraging our initial image, we regularize our attention losses with a loss encouraging similarity between background latent pixels in the initial and diffused image:
\begin{equation}
    \mathcal{L}_{bg} = \frac{1}{|\mathcal{B}|} \sum_{i \in \mathcal{B}} (\bar{z}^{(t)}_i - z^{(t)}_i)^2
\end{equation}
where $\mathcal{B}$ is the background mask, $\bar{z}^{(t)}_i$ \ignorethis{\ykc{$\bar{z}^{(t)}$?}} is the latent representation of the initial image at timestep $t$ and $z^{(t)}_i$ is the latents of the optimized image at timesetep $t$. With this regularization objective, our overall loss becomes:
\begin{equation}
    \mathcal{L} = \mathcal{L}_{obj} + \lambda_{att}\mathcal{L}_{att} + \lambda_{bg}\mathcal{L}_{bg}
\end{equation}
with $\lambda_{att}$ set to $0.8$ and $\lambda_{bg}$ set to $0.3$ in our experiments. As a final note, we point out that the optimizing and masking does not occur at every timestep and layer of inference, the timesteps and layer used for each components are detailed in the supplementary material.

\section{Experiments}
In this section, we conduct both qualitative and quantitative experiments to assess the effectiveness of our method. In the supplementary material, we provide additional implementation details, discuss limitations, and show results and comparisons over the entire \benchmarkName{} benchmark, presented next in Section \ref{sec:benchmarks}. 

\subsection{Benchmarks}
\label{sec:benchmarks}
While several benchmarks exist for evaluating text-to-image models, each addresses different aspects of model performance with certain limitations. DrawBench~\cite{saharia2022photorealistic} and GenEval \cite{ghosh2024geneval} offer comprehensive evaluation of several image generation skills, but only a subset of the prompts specifically target object counting, spatial relationships and attribute fidelity.
We present an evaluation of our method on DrawBench in Section~\ref{sec:comparisons}, and on the GenEval dataset in the supplementary material.
Other datasets that address these aspects either employ oversimplified prompts for semantic segmentation evaluation (HRS~\cite{bakr2023hrs}) or lack instance-level attribute specification due to their procedurally generated nature (ConceptMix~\cite{wu2024conceptmix}).

 \noindent \textbf{\benchmarkName{}.} 
To address these limitations, we introduce a new benchmark, \benchmarkName{},
specifically designed to evaluate models' capabilities in numerical accuracy, spatial fidelity, and semantic attribute consistency. \benchmarkName{} provides prompts that compound precise object counting with complex spatial arrangements and detailed attribute specifications.

The \benchmarkName{} dataset was constructed with three difficulty tiers, with corresponding prompts for each tier.
The first tier (\emph{Tier A}) describes multiple instances of one or more object categories (e.g. ``an image of two dogs and three cats'').
The second tier (\emph{Tier B}) enhances these base prompts by incorporating instance-level attributes for specific objects (e.g., ``an image of two dogs and three cats, one of the cats is a sphinx while the other cats are tabbies''). The third tier (\emph{Tier C})  extends the prompts further by adding spatial relationships at both object and attribute levels (e.g., ``an image of two dogs and three cats, the cat on the far right is a sphinx while the other cats are tabbies''). %
In total, this dataset contains $540$ prompts, split across three difficulty tiers.

Following \cite{wu2024conceptmix}, we pair each unique prompt with a set of yes or no questions. Tier A prompts are accompanied by a question centered on counting (i.e. ``are there exactly two dogs and three cats in this image?''). Tier B prompts are paired with a counting question and a question centered on counting the instances with unique attributes (i.e. ``are there exactly two tabbies and a single sphinx cat?''). Tier C prompts are accompanied by a counting question, an attribute counting question and a question focused on spatial fidelity (``is the cat on the far right a sphinx?'').

\subsection{Metrics}
When evaluating on DrawBench we use the procedure used in \cite{phung2024grounded}.
When evaluating on the \benchmarkName{} benchmark we use the evaluation procedure used in ConceptMix~\cite{wu2024conceptmix}, i.e. using an MLLM (GPT-4o) to answer the yes or no questions accompanying each prompt. For an image to get a positive score, the MLLM must give a positive answer to all of its accompanying questions. We denote this metric as \emph{VQA Accuracy} or \emph{VQA Acc}. Additionally, we use the recently introduced VQAScore \cite{lin2025evaluating} metric to evaluate general image-text similarity, as it has been shown to have significantly stronger agreement with human judgments. To avoid confusion with \emph{VQA Acc}, we denote VQAScore as \emph{VQA Similarity} or \emph{VQA Sim}.

\begin{table}
    \centering
    \setlength{\tabcolsep}{0.003\textwidth}
    \begin{tabular}{l c c c c}
        \toprule
        Method & \multicolumn{3}{c}{Counting} & Spatial \\
        \cmidrule{2-5}
        & Precision & Recall & F1 & Accuracy \\
        \midrule
        SDXL & 0.74 & 0.78 & 0.76 & 0.19 \\
        Flux1-dev & 0.71 & 0.90 & 0.79 & 0.50 \\
        CountGen & 0.89 & 0.85 & 0.87 & 0.17 \\
        ReasonYourLayout & 0.78 & 0.84 & 0.81 & 0.39 \\
        BoundedAttention & 0.83 & 0.88 & 0.85 & 0.36 \\
        AttentionRefocusing & \textbf{0.90} & 0.86 & 0.88 & 0.64 \\
        DPO & 0.76 & 0.92 & 0.83 & 0.17 \\
        SLD & 0.84 & 0.96 & 0.90 & 0.39 \\
        Emu & 0.77 & 0.96 & 0.85 & 0.50 \\
        Ours & 0.87 & \textbf{0.96} & \textbf{0.91} & \textbf{0.67} \\
        \bottomrule
    \end{tabular}
    \captionof{table}{
    Quantitative evaluation
    on the DrawBench dataset.
    }
    \label{table:drawbench}
\end{table}

\begin{table}
    \centering
    \setlength{\tabcolsep}{0.006\textwidth}
    \begin{tabular}{l c c c c c}
        \toprule
        Method & \multicolumn{4}{c}{VQA Acc} & VQA Sim \\
        \cmidrule{2-5}
        & A & B & C & Avg. \\
        \midrule
        SDXL & 0.42 & 0.18 & 0.11 & 0.23 & 0.78 \\
        Flux1-dev & 0.58 & 0.37 & 0.38 & 0.44 & 0.85 \\
        CountGen & 0.45 & 0.18 & 0.10 & 0.24 & 0.75 \\
        Reason-your-Layout & 0.32 & 0.10 & 0.07 & 0.16 & 0.65 \\
        BoundedAttention & 0.39 & 0.14 & 0.10 & 0.21 & 0.77 \\
        AttentionRefocusing & 0.63 & 0.13 & 0.07 & 0.27 & 0.68 \\
        DPO & 0.33 & 0.16 & 0.07 & 0.19 & 0.80 \\
        SLD & 0.44 & 0.14 & 0.07 & 0.21 & 0.76 \\
        Emu  & 0.59 & 0.40 & 0.30 & 0.43 & 0.86 \\
        Ours & \textbf{0.72} & \textbf{0.57} & \textbf{0.50} & \textbf{0.60} & \textbf{0.89}  \\
        \bottomrule
    \end{tabular}
    \captionof{table}{
    Quantitative evaluation
    on the \benchmarkName{} dataset.
    }
    \label{table:asamo}
\end{table}

\subsection{Comparisons with baselines}
\label{sec:comparisons}
We compare our method with baseline generic image generation methods (Emu~\cite{dai2023emuenhancingimagegeneration}, SDXL~\cite{podell2023sdxlimprovinglatentdiffusion}, Flux~\cite{flux2023}), as well as untrained methods that specifically aim to produce images of multiple objects with high fidelity (Bounded Attention~\cite{dahary2025yourself}, Reason Your Layout~\cite{chen2023reason}, CountGen~\cite{binyamin2024make}, SLD~\cite{wu2024self}). We compare with Attention Refocusing~\cite{phung2024grounded}, which uses a \emph{supervised} grounded text-to-image method (GLIGEN~\cite{li2023gligen}) that accepts bounding box layouts as an additional input. Finally, we compare with DPO ~\cite{wallace2024diffusion} which relies on human feedback during training to improve prompt fidelity.
Below we detail quantitative and qualitative results of the comparison.

 \noindent \textbf{Quantitative results.}
\label{sec:quantitative}
Table~\ref{table:drawbench} presents a quantitative comparison 
on the DrawBench benchmark. 
As illustrated in the table, our method outperforms all baselines in both counting and spatial accuracy, with a significant improvement over our baseline method (Emu).
In Table~\ref{table:asamo} we present the results of a quantitative comparison conducted on the \benchmarkName{} benchmark. This table presents VQA Accuracy results across the three prompt types in \benchmarkName{}. While performance is somewhat comparable for type A prompts, our method significantly outperforms competing methods on type B (multiple objects with instance-level attributes) and type C prompts (multiple objects with instance-level attributes \emph{and} spatial arrangements). This gap suggests competing methods struggle with instance-level attributes and spatial arrangements. Furthermore, our method also outperforms all baselines in VQA Similarity, a metric assessing overall composition fidelity.

\begin{figure}
    \centering
    \small
    A: \emph{"\counttext{a} man and \counttext{a} woman sitting down with their \counttext{two} dogs"}
    \\
    \jsubfig{\includegraphics[width=0.193\linewidth]{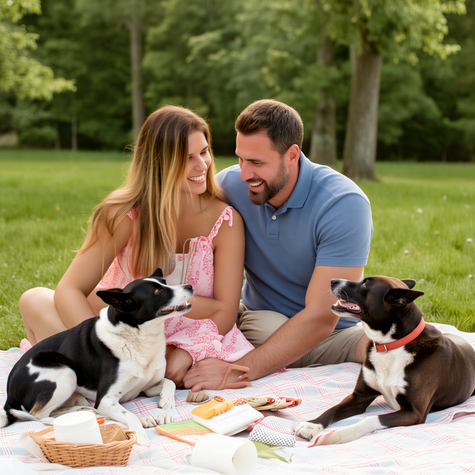}{}} 
    \hfill
    \jsubfig{\includegraphics[width=0.193\linewidth]{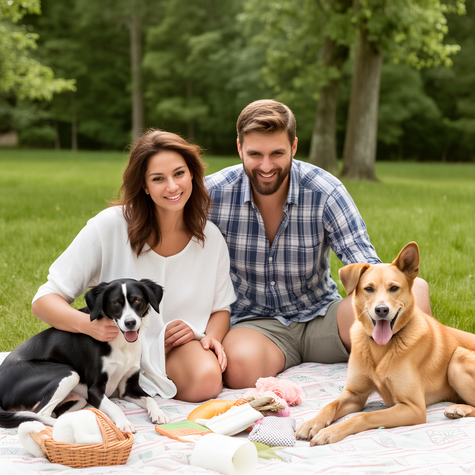}{}} 
    \hfill
    \jsubfig{\includegraphics[width=0.193\linewidth]{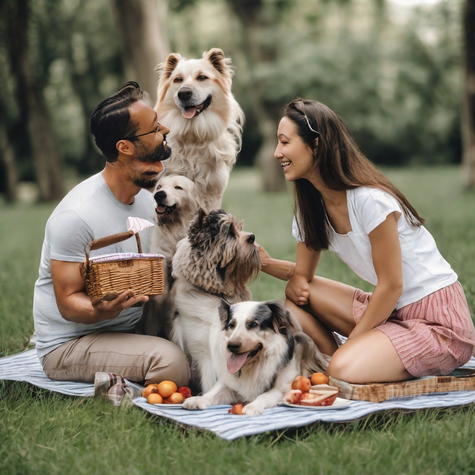}{}} 
    \hfill
    \jsubfig{\includegraphics[width=0.193\linewidth]{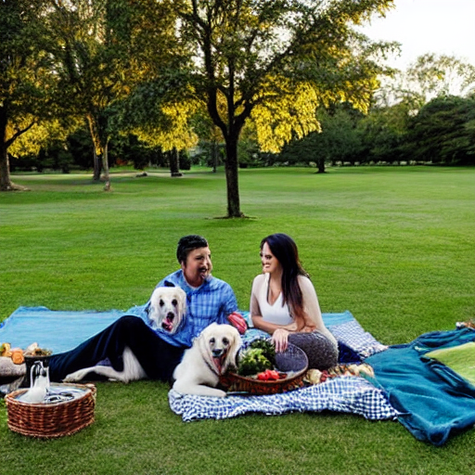}{}} 
    \hfill
    \jsubfig{\includegraphics[width=0.193\linewidth]{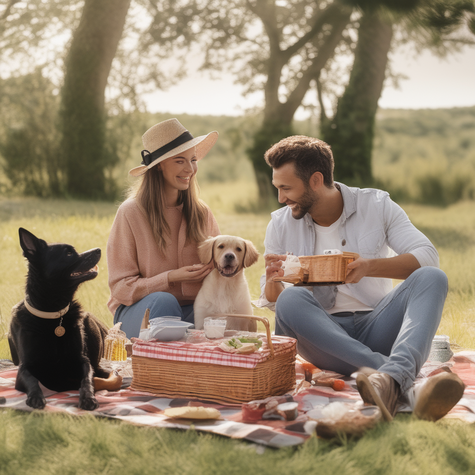}{}} 
    \hfill
    \\ %
    B: \emph{"\counttext{a} man and \counttext{a} woman sitting down with their \counttext{two} dogs, \attrtext{one} of the dogs is a \attrtext{golden retriever}"}
    \\
    \jsubfig{\includegraphics[width=0.193\linewidth]{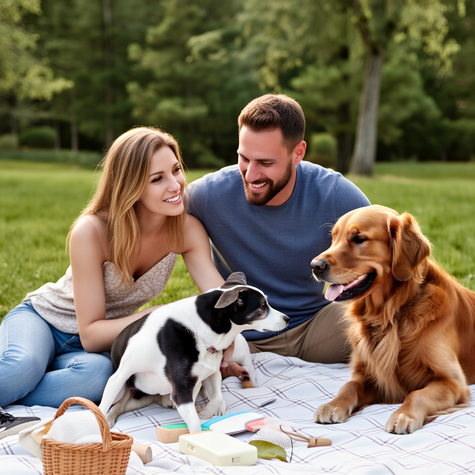}{}} 
    \hfill
    \jsubfig{\includegraphics[width=0.193\linewidth]{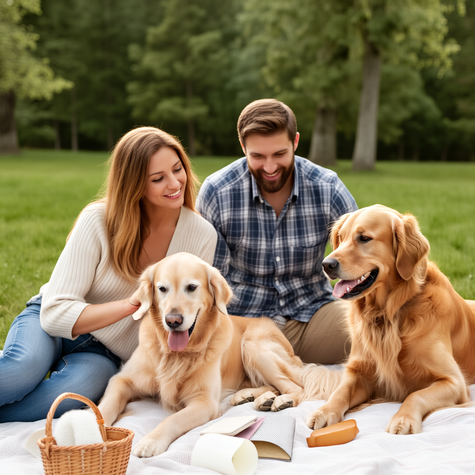}{}} 
    \hfill
    \jsubfig{\includegraphics[width=0.193\linewidth]{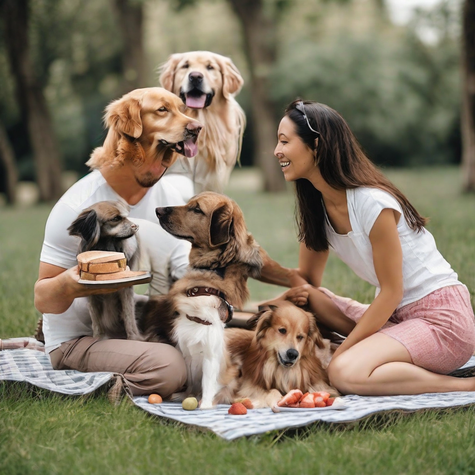}{}} 
    \hfill
    \jsubfig{\includegraphics[width=0.193\linewidth]{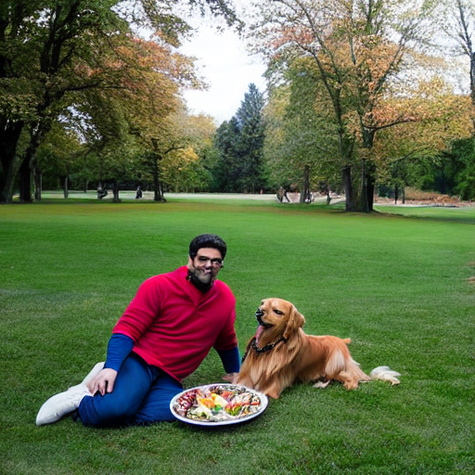}{}} 
    \hfill
    \jsubfig{\includegraphics[width=0.193\linewidth]{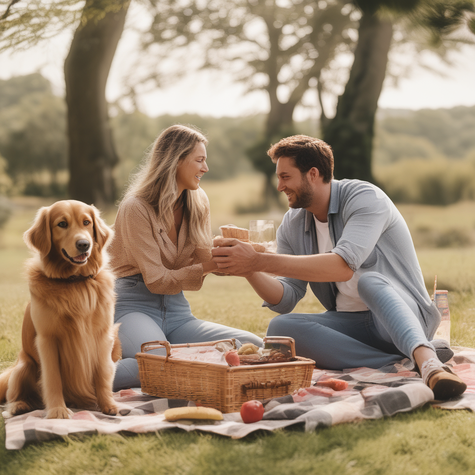}{}} 
    \hfill
    \\ %
    C:  \emph{"\counttext{a} man and \counttext{a} woman sitting down with their \counttext{two} dogs, the dog \spatialtext{between the man and the woman} is a \attrtext{golden retriever}"}
    \\
    \jsubfig{\includegraphics[width=0.193\linewidth]{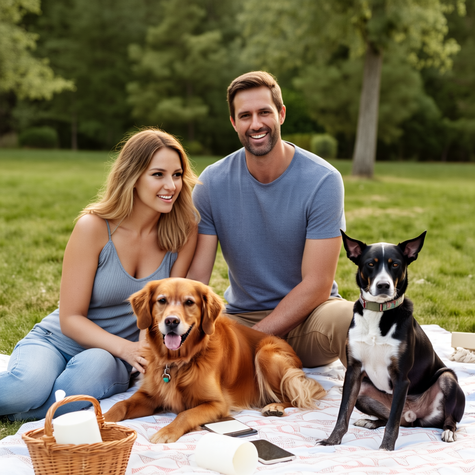}{}} 
    \hfill
    \jsubfig{\includegraphics[width=0.193\linewidth]{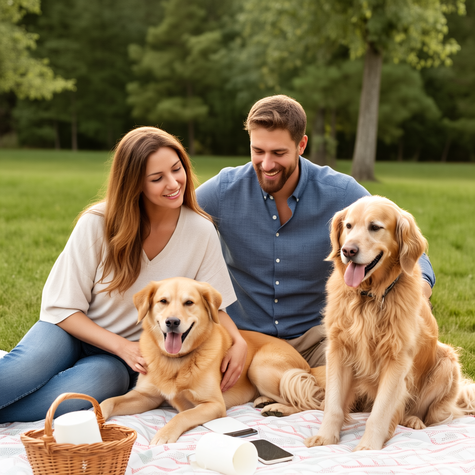}{}} 
    \hfill
    \jsubfig{\includegraphics[width=0.193\linewidth]{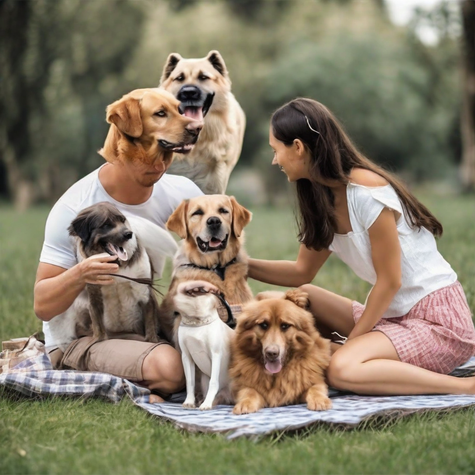}{}} 
    \hfill
    \jsubfig{\includegraphics[width=0.193\linewidth]{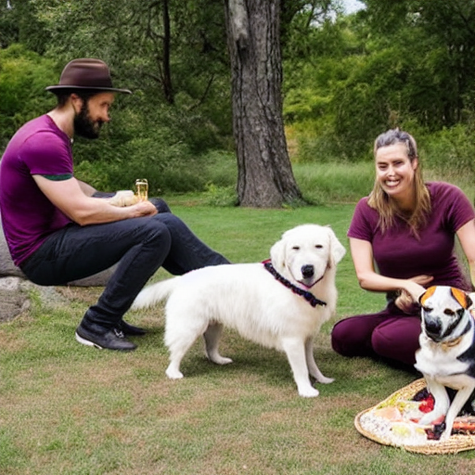}{}} 
    \hfill
    \jsubfig{\includegraphics[width=0.193\linewidth]{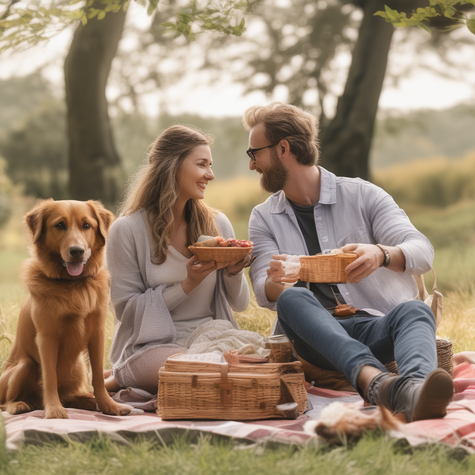}{}} 
    \hfill
    \\
    A: \emph{"\counttext{five} donuts in a box"}
    \\
    \jsubfig{\includegraphics[width=0.193\linewidth]{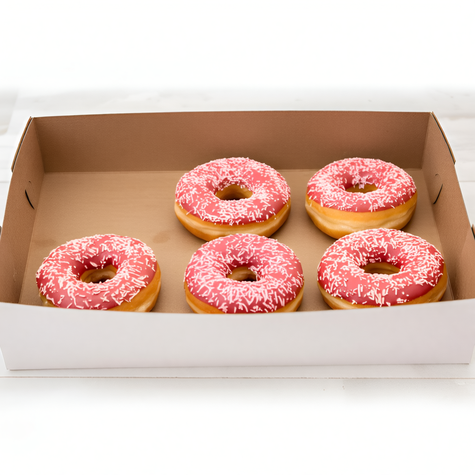}{}} 
    \hfill
    \jsubfig{\includegraphics[width=0.193\linewidth]{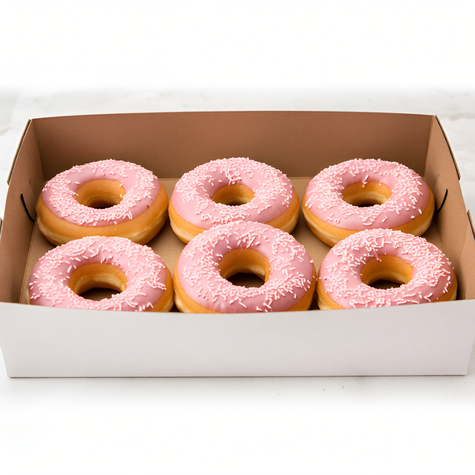}{}} 
    \hfill
    \jsubfig{\includegraphics[width=0.193\linewidth]{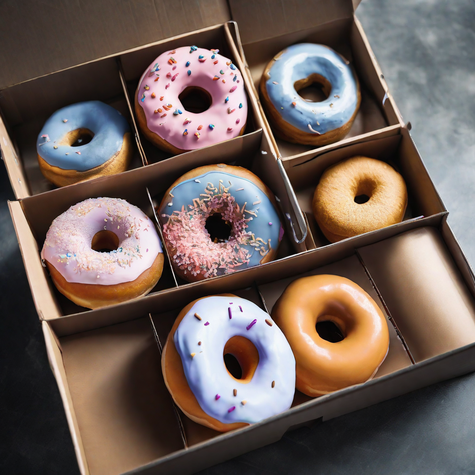}{}} 
    \hfill
    \jsubfig{\includegraphics[width=0.193\linewidth]{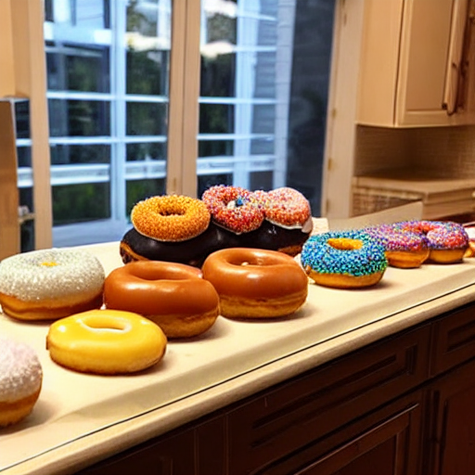}{}} 
    \hfill
    \jsubfig{\includegraphics[width=0.193\linewidth]{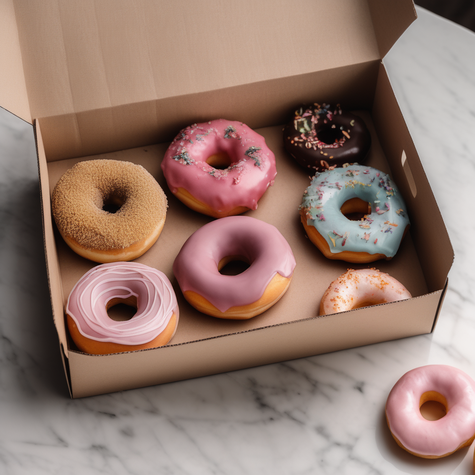}{}} 
    \hfill
    \\
    B: \emph{"\counttext{five} donuts in a box, \attrtext{two} are \attrtext{chocolate glazed}, the rest are \attrtext{plain}"}
    \\
    \jsubfig{\includegraphics[width=0.193\linewidth]{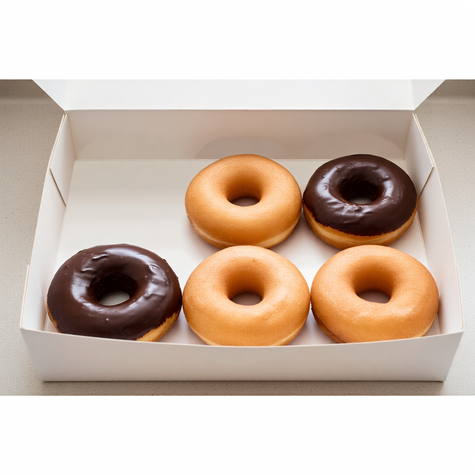}{}} 
    \hfill
    \jsubfig{\includegraphics[width=0.193\linewidth]{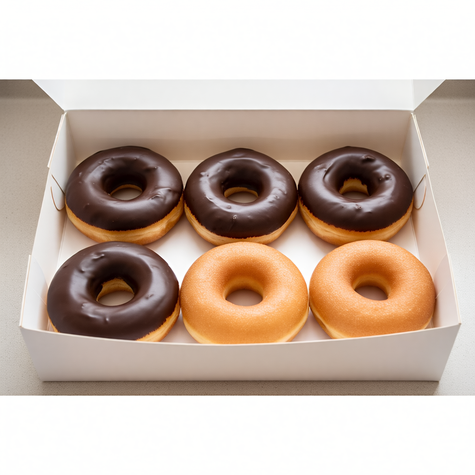}{}} 
    \hfill
    \jsubfig{\includegraphics[width=0.193\linewidth]{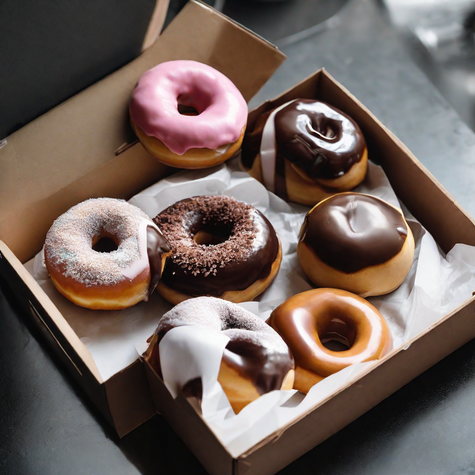}{}} 
    \hfill
    \jsubfig{\includegraphics[width=0.193\linewidth]{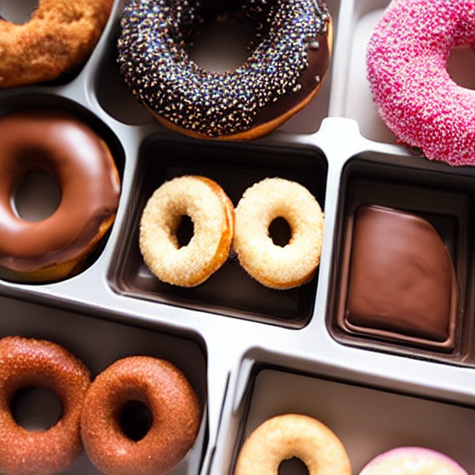}{}} 
    \hfill
    \jsubfig{\includegraphics[width=0.193\linewidth]{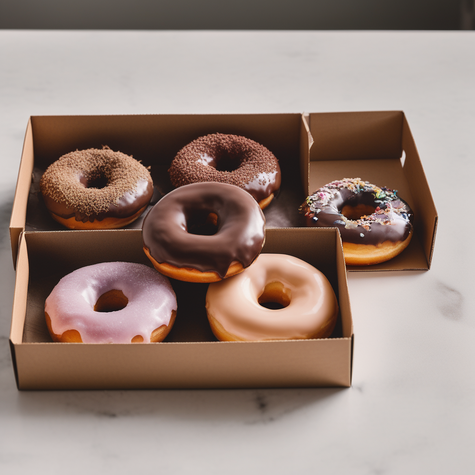}{}} 
    \hfill
    \\
    C: \emph{"\counttext{five} donuts in a box, the donut on \spatialtext{the top left} and \spatialtext{bottom right} are \attrtext{chocolate glazed}, the rest are \attrtext{plain}"}
    \\
    \jsubfig{\includegraphics[width=0.193\linewidth]{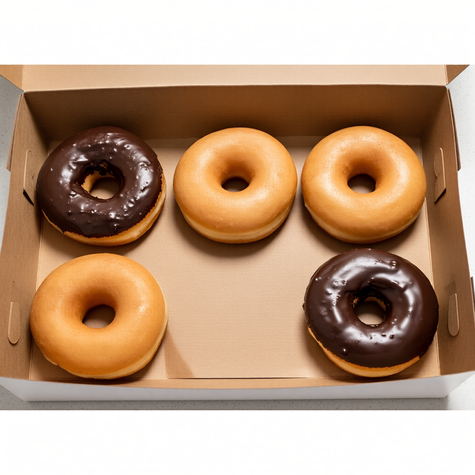}}{\footnotesize{Ours}} 
    \hfill
    \jsubfig{\includegraphics[width=0.193\linewidth]{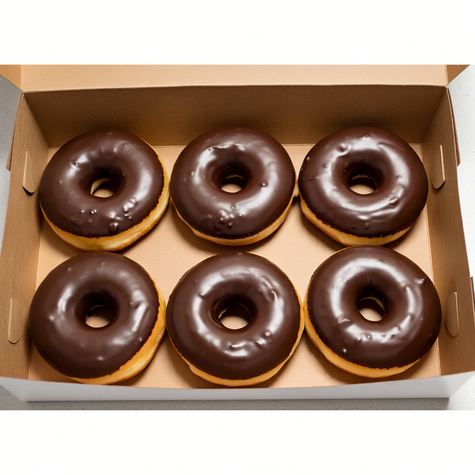}}{\footnotesize{Emu}} 
    \hfill
    \jsubfig{\includegraphics[width=0.193\linewidth]{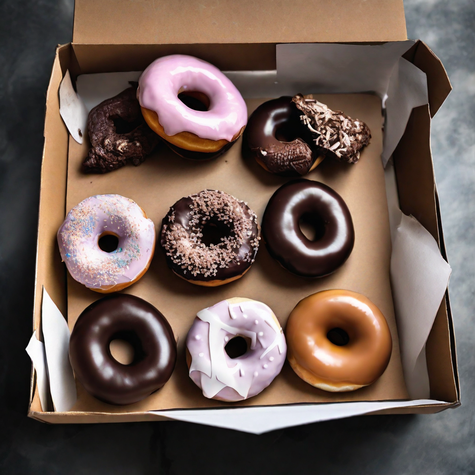}}%
    {\footnotesize{BA}} 
    \hfill
    \jsubfig{\includegraphics[width=0.193\linewidth]{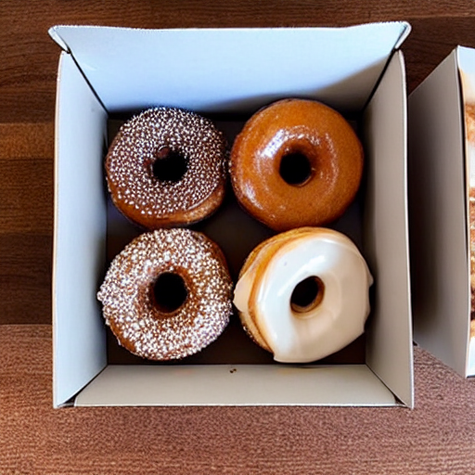}}%
    {\footnotesize{RYL}} 
    \hfill
    \jsubfig{\includegraphics[width=0.193\linewidth]{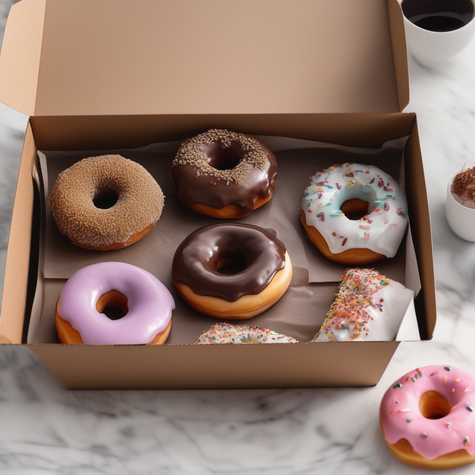}}{\footnotesize{SDXL}} 
    \hfill
    \\
\caption{\textbf{Qualitative results from the \benchmarkName{} benchmark.} We present results for two unique prompts from the \benchmarkName{} benchmark, presenting results for all three tiers (A, B and C) for each prompt.}
    \label{fig:baseline_comparison}
\end{figure}

 \noindent \textbf{Qualitative results.}
Qualitative comparisons with baselines methods is presented in Fig.~\ref{fig:baseline_comparison} and Fig.~\ref{fig:compound_w_questions}. Fig.~\ref{fig:baseline_comparison} presents images generated for two sets of prompts from the \benchmarkName{} dataset, each with the three difficulty tiers A, B, and C.
Note that while most baselines can generate a simple prompt in Tier A such as ``a man and a woman sitting down with their two dogs'', they cannot satisfy the instance-level attributes and instance-level counts which are introduced in tier B correctly. In addition, some baselines struggle with long prompts and generate very similar images for the different prompts in tiers B and C. 
As demonstrated by the second prompt (rows 4 to 6), baseline image generation methods have strong biases for some object counts, for example a bias towards generating an even number of donuts. This bias is not solved by previous methods such as Bounded Attention and Reason Your Layout, whereas our method ensures exactly five donuts are present in the final image. This bias also propagates to more complex prompts in tier B and C.

Fig.~\ref{fig:compound_w_questions} presents additional two sets of prompts from \texttt{Compound}-\texttt{Prompts}
benchmark with their three difficulty tiers, and comparison with additional methods (Flux, Attention Refocusing, and CountGen). The figure also displays the questions that are used for evaluating VQA Accuracy metric. While several methods can generate object counts correctly (particularly for lower counts), they rarely satisfy all of the conditions of the prompts in tiers B and C.

We show additional comparison with three prompts in Fig.~\ref{fig:teaser}. 
The results for all prompts in \benchmarkName{} can be found in the supplemental material.

\begin{table}
    \centering
    \setlength{\tabcolsep}{0.008\textwidth}
    \begin{tabular}{l c c c c c}
        \toprule
        Method & \multicolumn{4}{c}{VQA Acc} & VQA Sim \\
        \cmidrule{2-5}
        & A & B & C & Avg. \\
        \midrule
        w/o $\mathcal{L}_{bg}$& 0.68 & 0.60 & 0.54 & 0.61 & 0.84 \\
        w/o $\mathcal{L}_{obj}$ \& $\mathcal{L}_{att}$& 0.72 & 0.54 & 0.54 & 0.60 & 0.89 \\
        w/o Attn. Masking & 0.72 & 0.56 & 0.32 & 0.54 & 0.87 \\
        w/o M-RCNN & 0.70 & 0.60 & 0.50 & 0.60 & 0.88 \\
        w/o Seed Search & \textbf{0.84} & 0.56 & 0.56 & 0.65 & \textbf{0.90}  \\
        Ours & \textbf{0.76} & \textbf{0.72} & \textbf{0.60} & \textbf{0.69} & \textbf{0.90}  \\
        \bottomrule
    \end{tabular}
    \captionof{table}{
    Quantitative ablation experiment on a representative subset of the \benchmarkName{} dataset. 
    }
    \label{table:ablations}
\end{table}

\subsection{Ablations}
\label{sec:ablations}
We provide an ablation study on the \benchmarkName{} benchmark in Table \ref{table:ablations}; qualitative results are provided in the supplementary material. Specifically, we ablate key components in both the instance layout generation and assignment conditioned image generation phases of our framework. In the instance layout generation phase, we ablate the dual segmentation approach by using SAM (``w/o M-RCNN'') exclusively and ablate the seed search by using the initial seed and relying on our instance copying procedure exclusively to handle undergeneration  (``w/o Seed Search'').
In the assignment conditioned image generation phase, we ablate three key components: composition preserving regularization (``w/o $\mathcal{L}_{bg}$''), cross attention losses (``w/o $\mathcal{L}_{obj}$ \& $\mathcal{L}_{att}$'') and attention masking (``w/o Attn. Masking'').%

The results of this study are presented in Table \ref{table:ablations} and show that removing each of the ablated components causes some loss in performance most noticeably in terms of VQA Accuracy. This is especially evident on the more challenging Type B and Type C prompts, with attention masking having a particularly strong effect on Type C performance. In our assessment, this shows that the semantic leakage which this component addresses plays a major part in spatial errors on our benchmark. Another interesting observation is that while removing $\mathcal{L}_{bg}$ does not diminish VQA Acc values as much as removing other components does, its affect on VQA Sim is highly significant. This emphasizes the importance of this regularization in maintaining visual quality and plausibility. Finally, we observe that ablating the seed search component does not cause a major drop off in VQA Accuracy and does not decrease VQA Similarity, which demonstrates that this component is not the primary factor driving our increased performance compared to competing baselines.

\section{Conclusion}
In this work, we presented \methodName{}, a new method that enhances the ability of text-to-image models to handle complex prompts compounding multiple objects, attributes, and spatial relationships without requiring additional training. Through comprehensive experiments, we demonstrated \methodName{}'s superior performance in counting accuracy and spatial arrangement compared to existing zero-shot approaches. Additionally, we introduced \texttt{Compound}-\texttt{Prompts}, a new benchmark specifically designed to evaluate models' capabilities in handling intricate, multi-component prompts. Our benchmark not only validates our method's effectiveness but also provides a valuable resource for future research. As text-to-image synthesis continues to evolve, our work represents a step toward enabling these models to faithfully render increasingly complex and creative concepts, bringing us closer to the goal of translating any imaginable scenario into visual reality.

\bibliographystyle{ACM-Reference-Format}
\bibliography{main}
\clearpage

\newcommand{\minipageimage}[3][0.12]{\begin{minipage}[t]{#1\linewidth}\centering \small #3 \\ \includegraphics[width=\linewidth]{#2}\end{minipage}}

\begin{figure*}

\begin{center}

	\centering

	\begin{minipage}{0.12\linewidth}
            \textbf{Tier A:}

	\end{minipage}
	\begin{minipage}{0.86\linewidth}
            \small
            Prompt: \emph{``\counttext{Six} meerkats standing watch in the Savannah''}

            Question (A): \emph{``Do exactly \counttext{six} meerkats appear in this image?''}

	\end{minipage}

        \vspace{1mm}

        \minipageimage
        {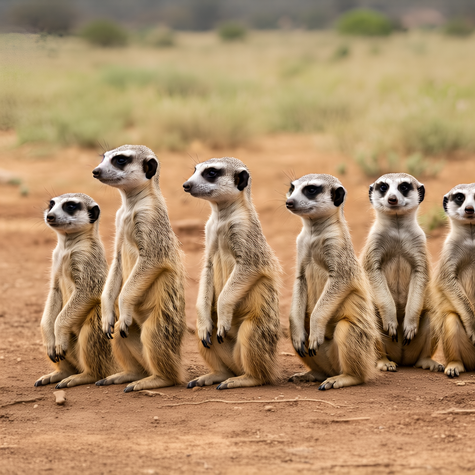}
        {\pass{} \quad \counttext{$\blacksquare$}}
        \minipageimage
        {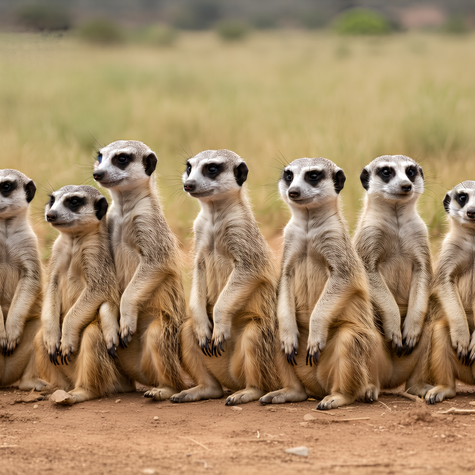}
        {\fail{} \quad \counttext{$\square$}}
        \minipageimage
        {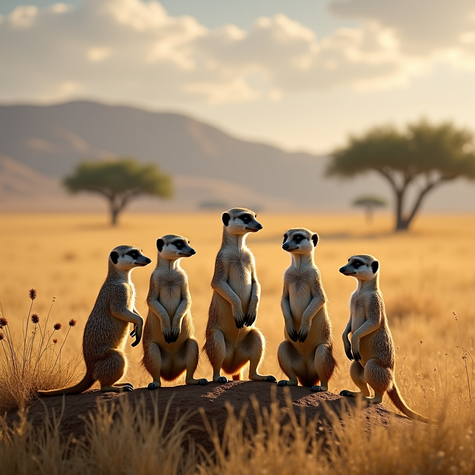}
        {\fail{} \quad \counttext{$\square$}}
        \minipageimage
        {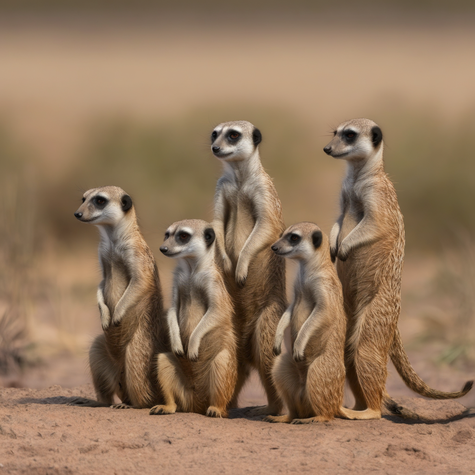}
        {\fail{} \quad \counttext{$\square$}}
        \minipageimage
        {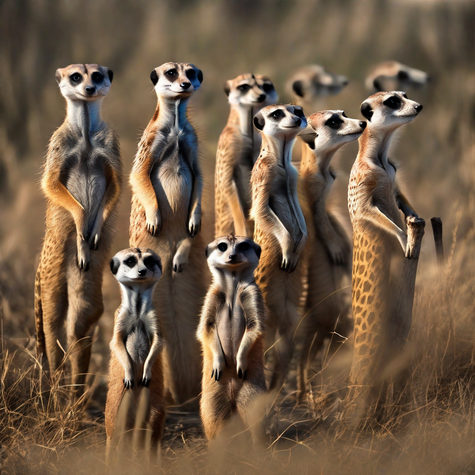}
        {\fail{} \quad \counttext{$\square$}}
        \minipageimage
        {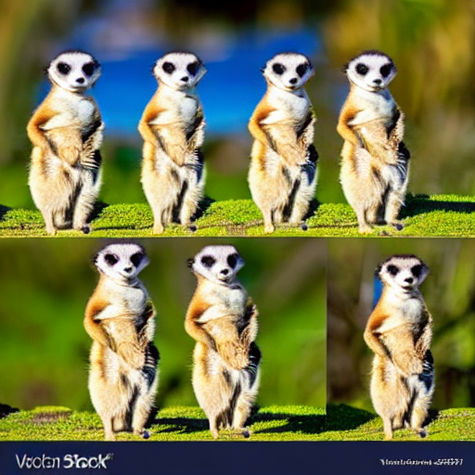}
        {\fail{} \quad \counttext{$\square$}}
        \minipageimage
        {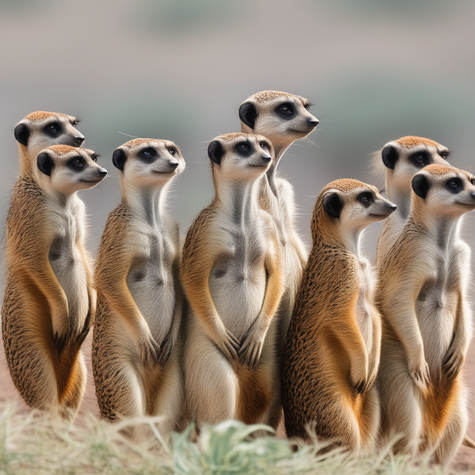}
        {\fail{} \quad \counttext{$\square$}}
        \minipageimage
        {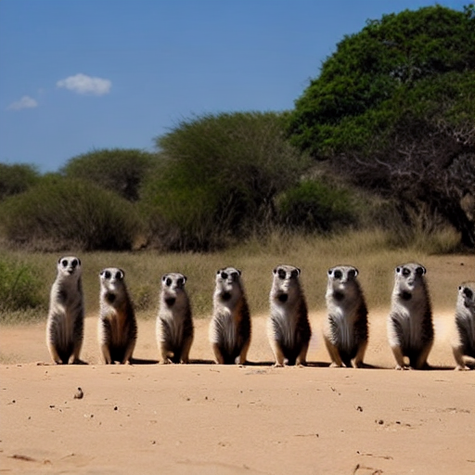}
        {\fail{} \quad \counttext{$\square$}}

        \vspace{1mm}

	\begin{minipage}{0.1\linewidth}
            \textbf{Tier B:}

	\end{minipage}
	\begin{minipage}{0.86\linewidth}
            \small
            Prompt: \emph{``\counttext{Six} meerkats standing watch in the Savannah, only \attrtext{one} of them is \attrtext{yawning}''}

            Question (B): \emph{``Is exactly \attrtext{one} meerkat \attrtext{yawning} in this image?''}

	\end{minipage}

        \vspace{2mm}

        \minipageimage
        {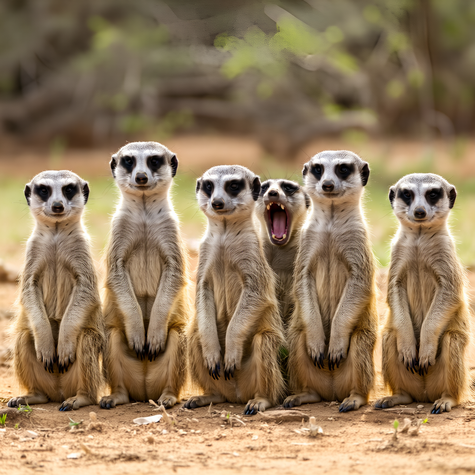}
        {\pass{} \quad \counttext{$\blacksquare$} \attrtext{$\blacksquare$}}
        \minipageimage
        {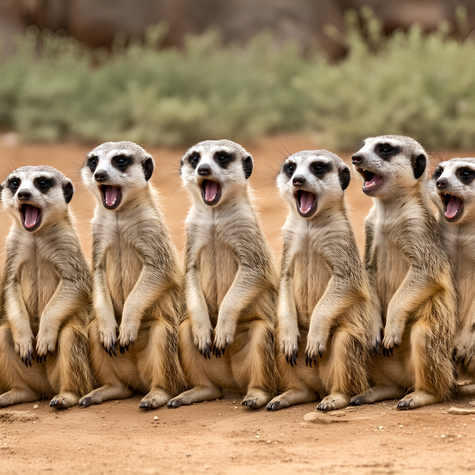}
        {\fail{} \quad \counttext{$\blacksquare$} \attrtext{$\square$}}
        \minipageimage
        {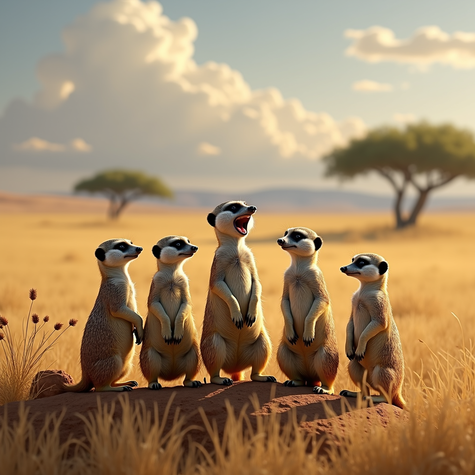}
        {\fail{} \quad \counttext{$\square$} \attrtext{$\square$}}
        \minipageimage
        {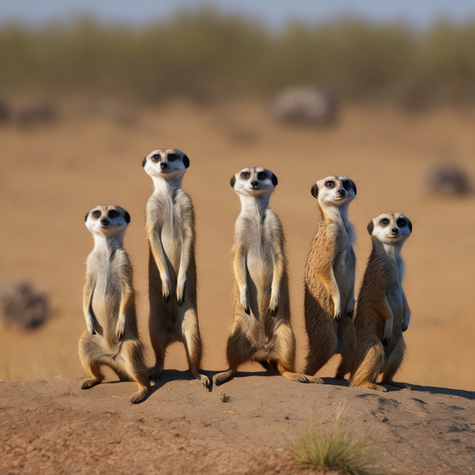}
        {\fail{} \quad \counttext{$\square$} \attrtext{$\square$}}
        \minipageimage
        {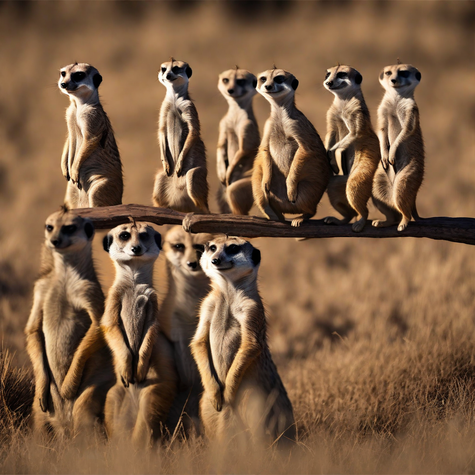}
        {\fail{} \quad \counttext{$\square$} \attrtext{$\square$}}
        \minipageimage
        {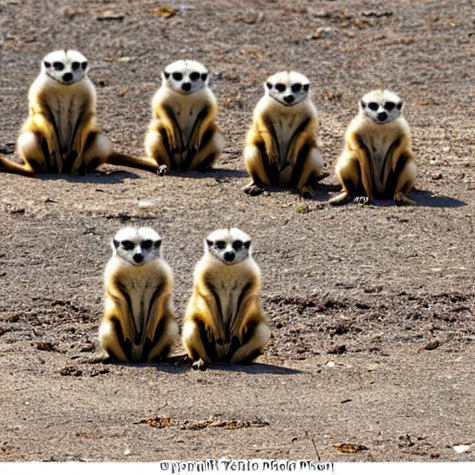}
        {\fail{} \quad \counttext{$\blacksquare$} \attrtext{$\square$}}
        \minipageimage
        {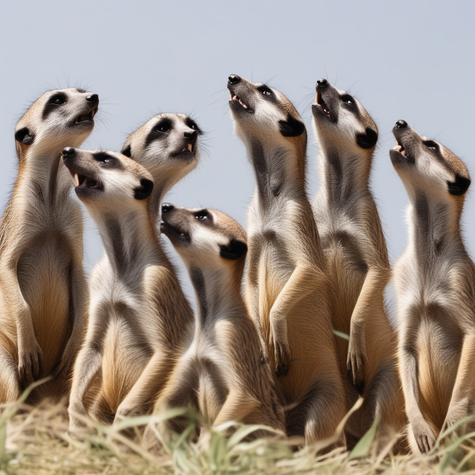}
        {\fail{} \quad \counttext{$\square$} \attrtext{$\square$}}
        \minipageimage
        {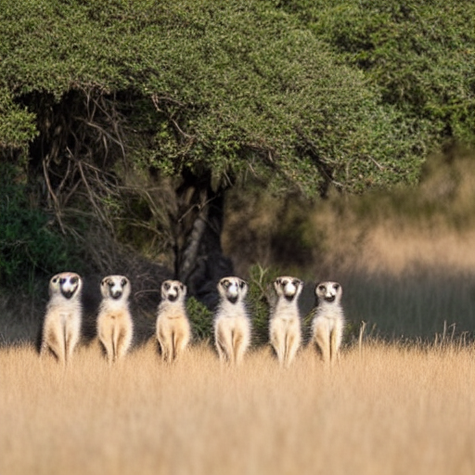}
        {\fail{} \quad \counttext{$\square$} \attrtext{$\square$}}

        \vspace{1mm}

	\begin{minipage}{0.1\linewidth}
            \textbf{Tier C:}
	\end{minipage}
	\begin{minipage}{0.86\linewidth}
             \small
           Prompt: \emph{``\counttext{Six} meerkats standing watch in the Savannah, only the \spatialtext{second meerkat from the right} is \attrtext{yawning}''}

            Question (C): \emph{``Is only the \spatialtext{second meerkat from the right} \attrtext{yawning} in this image?''}
	\end{minipage}
 
        \vspace{1mm}

        \minipageimage
        {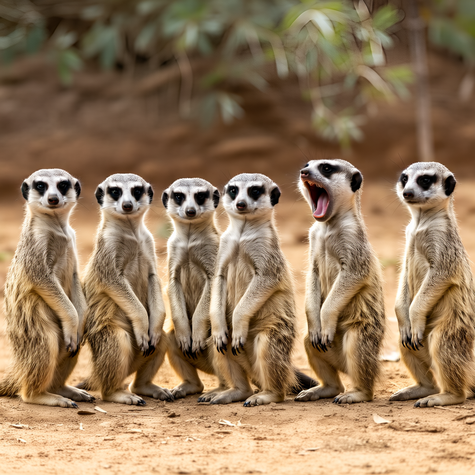}
        {\pass{} \quad \counttext{$\blacksquare$} \attrtext{$\blacksquare$} \spatialtext{$\blacksquare$}}
        \minipageimage
        {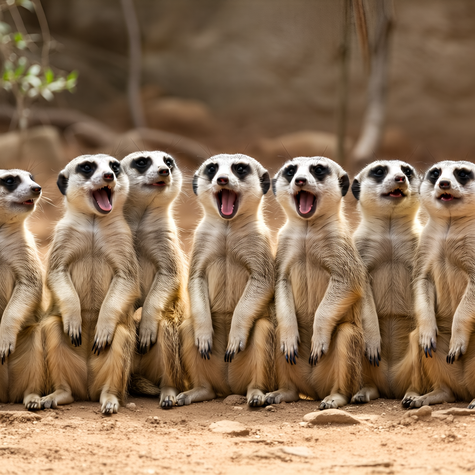}
        {\fail{} \quad \counttext{$\square$} \attrtext{$\square$} \spatialtext{$\square$}}
        \minipageimage
        {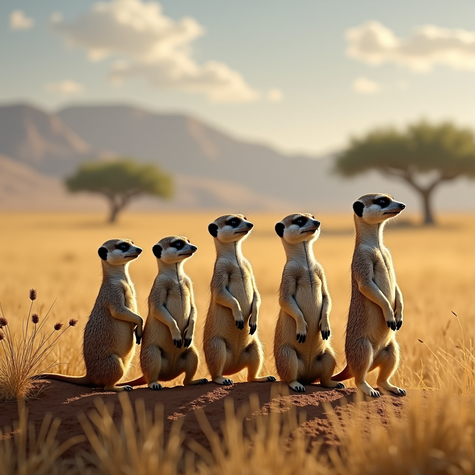}
        {\fail{} \quad \counttext{$\square$} \attrtext{$\square$} \spatialtext{$\square$}}
        \minipageimage
        {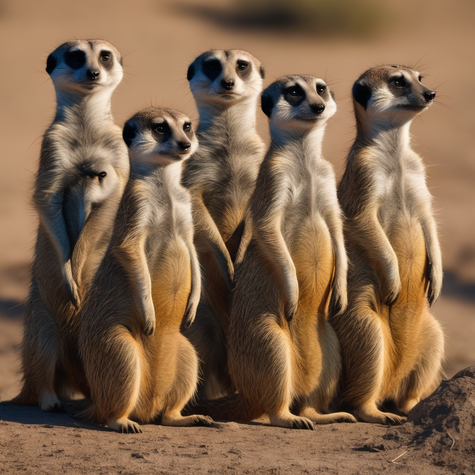}
        {\fail{} \quad \counttext{$\square$} \attrtext{$\square$} \spatialtext{$\square$}}
        \minipageimage
        {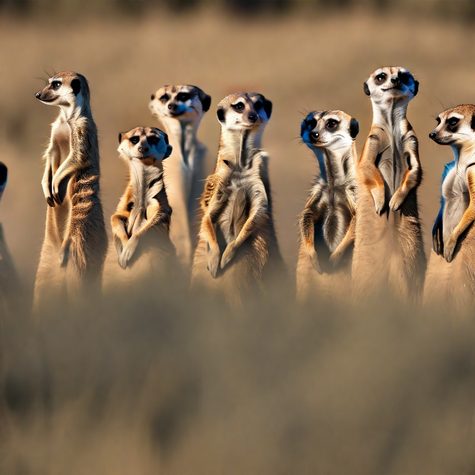}
        {\fail{} \quad \counttext{$\square$} \attrtext{$\square$} \spatialtext{$\square$}}
        \minipageimage
        {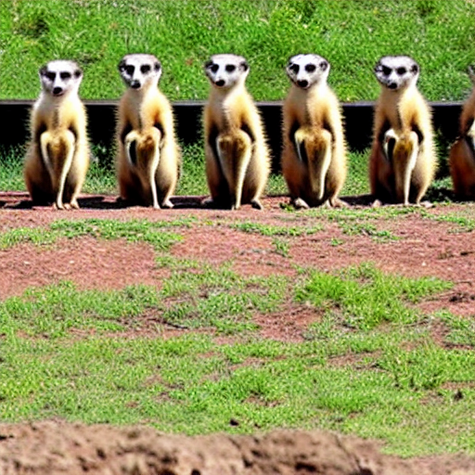}
        {\fail{} \quad \counttext{$\square$} \attrtext{$\square$} \spatialtext{$\square$}}
        \minipageimage
        {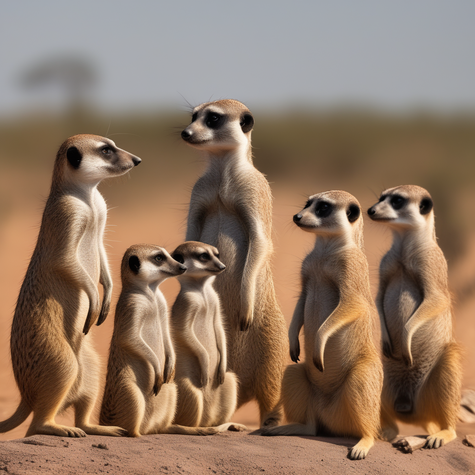}
        {\fail{} \quad \counttext{$\blacksquare$} \attrtext{$\square$} \spatialtext{$\square$}}
        \minipageimage
        {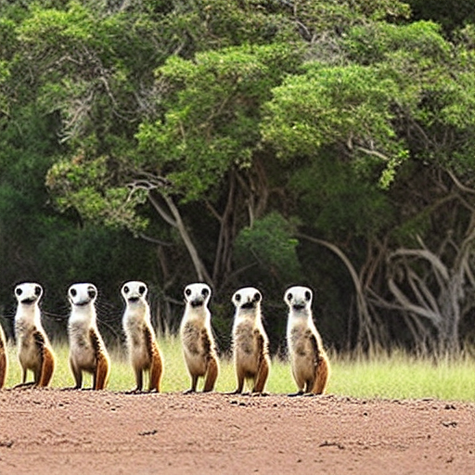}
        {\fail{} \quad \counttext{$\blacksquare$} \attrtext{$\square$} \spatialtext{$\square$}}

        \vspace{1mm}

        \vspace{1mm}

 	\begin{minipage}{0.12\linewidth}
            \textbf{Tier A:}

	\end{minipage}
	\begin{minipage}{0.86\linewidth}
             \small
           Prompt: \emph{``\counttext{Three} dogs and \counttext{a} cat in a backyard''}

            Question (A): \emph{``Do exactly \counttext{three} dogs and \counttext{one} cat appear in this image?''}

	\end{minipage}

        \vspace{1mm}

        \minipageimage
        {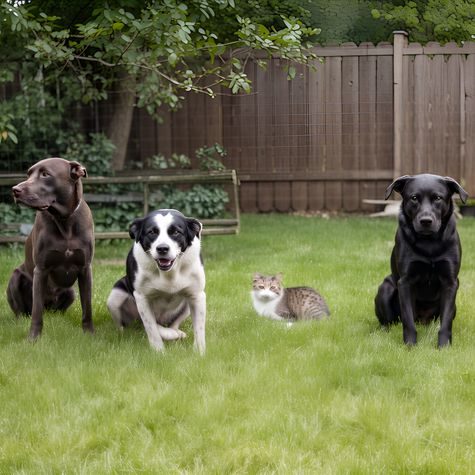}
        {\pass{} \quad \counttext{$\blacksquare$}}
        \minipageimage
        {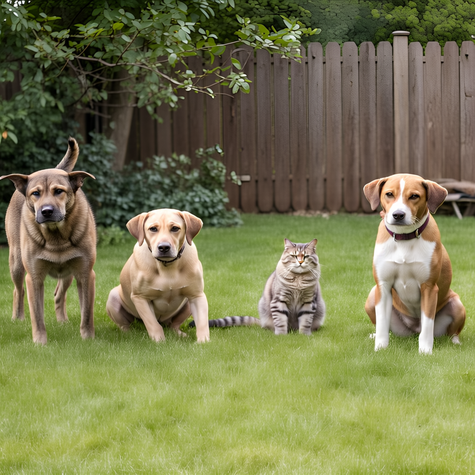}
        {\pass{} \quad \counttext{$\blacksquare$}}
        \minipageimage
        {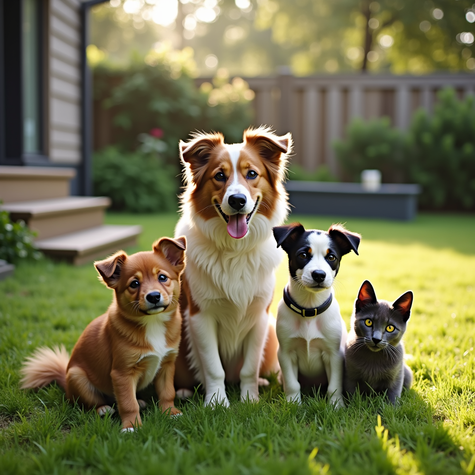}
        {\pass{} \quad \counttext{$\blacksquare$}}
        \minipageimage
        {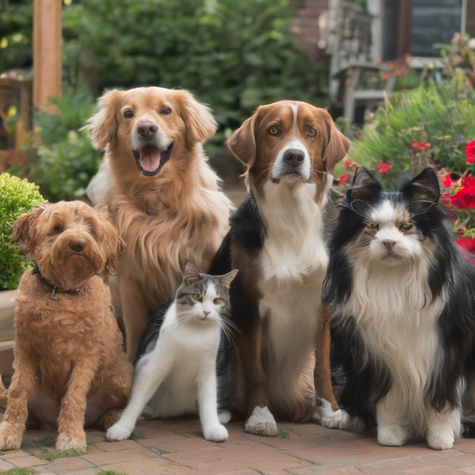}
        {\fail{} \quad \counttext{$\square$}}
        \minipageimage
        {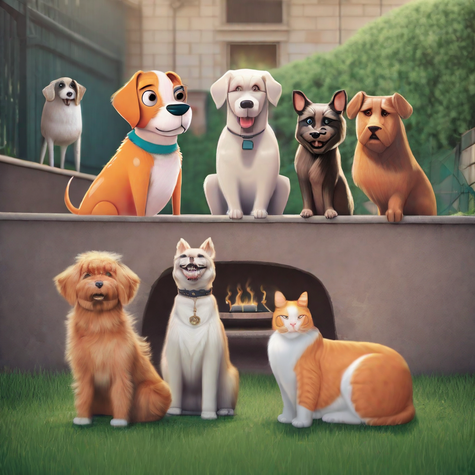}
        {\fail{} \quad \counttext{$\square$}}
        \minipageimage
        {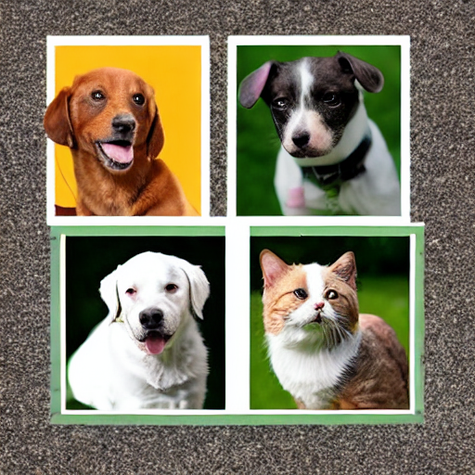}
        {\pass{} \quad \counttext{$\blacksquare$}}
        \minipageimage
        {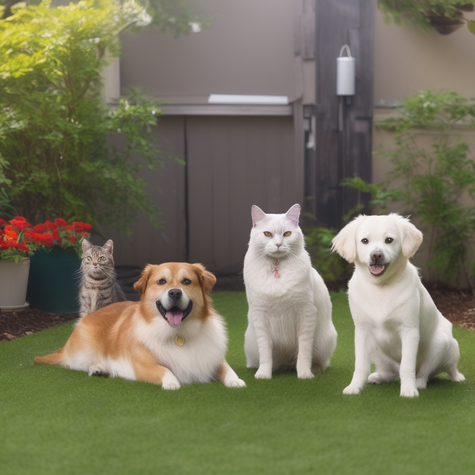}
        {\fail{} \quad \counttext{$\square$}}
        \minipageimage
        {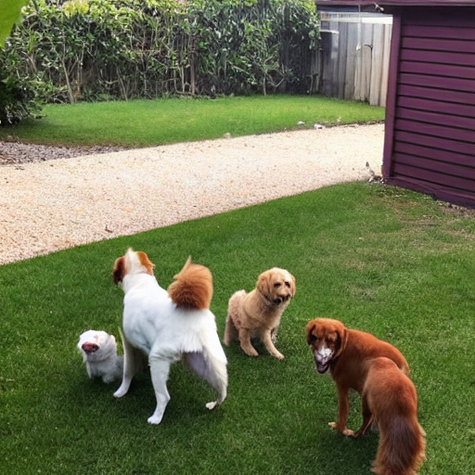}
        {\fail{} \quad \counttext{$\square$}}

        \vspace{1mm}

	\begin{minipage}{0.1\linewidth}
            \textbf{Tier B:}

	\end{minipage}
	\begin{minipage}{0.86\linewidth}
            \small
            Prompt: \emph{``\counttext{Three} dogs and \counttext{a} cat in a backyard, \attrtext{one} dog is a \attrtext{Schnauzer} and the cat is \attrtext{Persian}''}

            Question (B): \emph{``Is exactly \attrtext{one} dog a \attrtext{Schnauzer} and one cat a \attrtext{Persian} in this image?''}

	\end{minipage}

        \vspace{1mm}

        \minipageimage
        {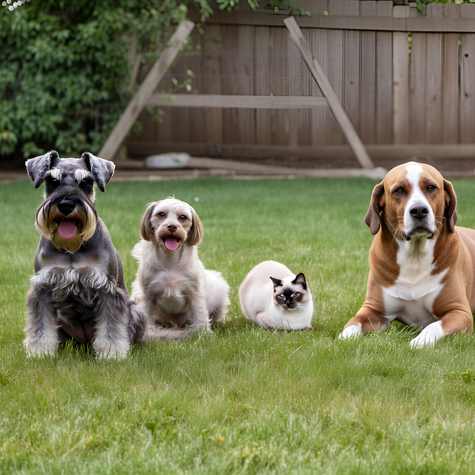}
        {\pass{} \quad \counttext{$\blacksquare$} \attrtext{$\blacksquare$}}
        \minipageimage
        {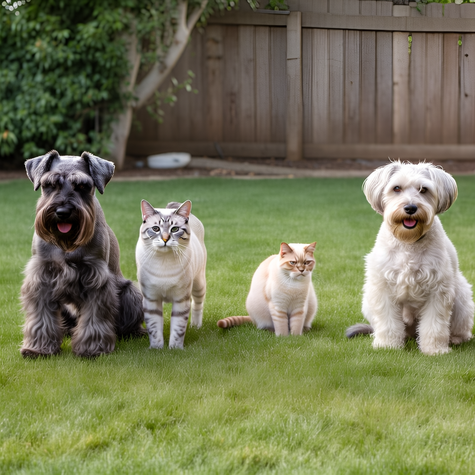}
        {\fail{} \quad \counttext{$\square$} \attrtext{$\square$}}
        \minipageimage
        {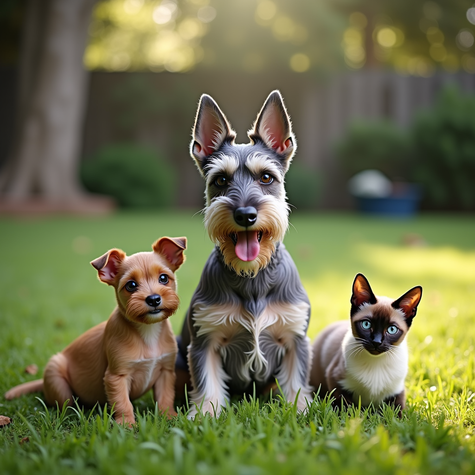}
        {\fail{} \quad \counttext{$\square$} \attrtext{$\blacksquare$}}
        \minipageimage
        {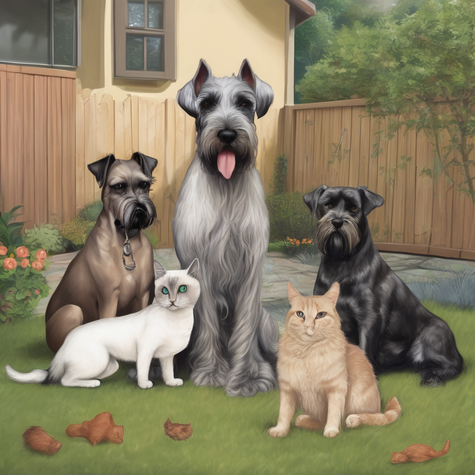}
        {\fail{} \quad \counttext{$\square$} \attrtext{$\square$}}
        \minipageimage
        {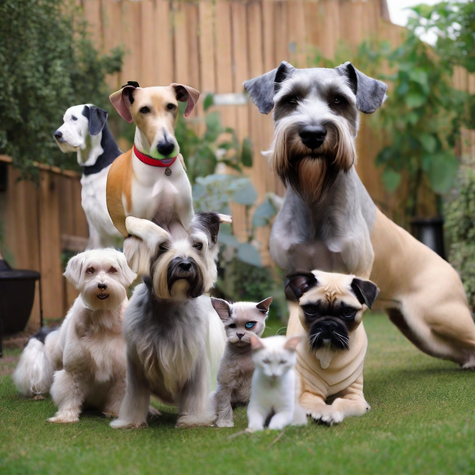}
        {\fail{} \quad \counttext{$\square$} \attrtext{$\square$}}
        \minipageimage
        {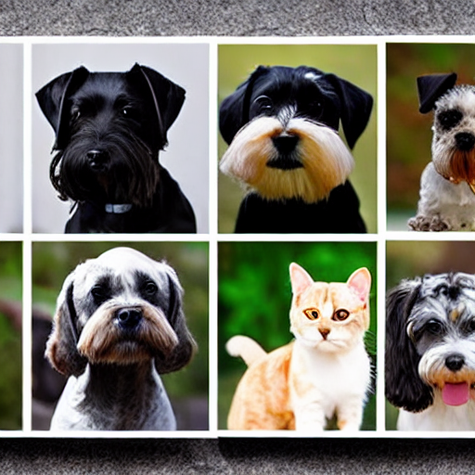}
        {\fail{} \quad \counttext{$\square$} \attrtext{$\square$}}
        \minipageimage
        {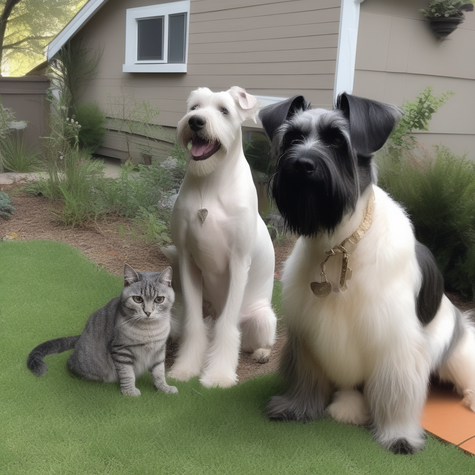}
        {\fail{} \quad \counttext{$\square$} \attrtext{$\square$}}
        \minipageimage
        {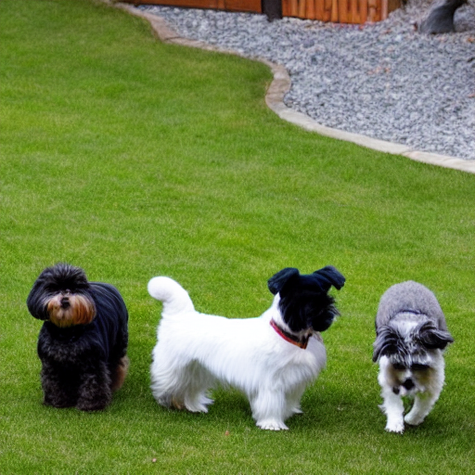}
        {\fail{} \quad \counttext{$\square$} \attrtext{$\square$}}

        \vspace{1mm}

	\begin{minipage}{0.1\linewidth}
            \textbf{Tier C:}
	\end{minipage}
	\begin{minipage}{0.86\linewidth}
            \small
            Prompt: \emph{``\counttext{Three} dogs and \counttext{a} cat in a backyard, \attrtext{one} dog is a \attrtext{Schnauzer} and is \spatialtext{on the far right}, the cat is \spatialtext{in the far left} and is \attrtext{Persian}''}

            Question (C): \emph{``Is only the dog \spatialtext{on the far right} a \attrtext{Schnauzer}, and only the cat \spatialtext{on the far left} a \attrtext{Persian} in this image?''}
	\end{minipage}
 
        \vspace{1mm}

        \minipageimage
        {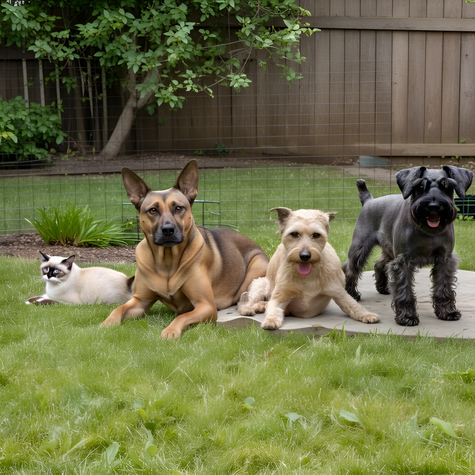}
        {\pass{} \quad \counttext{$\blacksquare$} \attrtext{$\blacksquare$} \spatialtext{$\blacksquare$}}
        \minipageimage
        {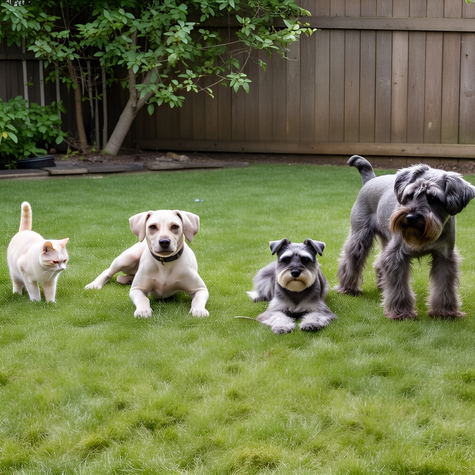}
        {\fail{} \quad \counttext{$\blacksquare$} \attrtext{$\square$} \spatialtext{$\square$}}
        \minipageimage
        {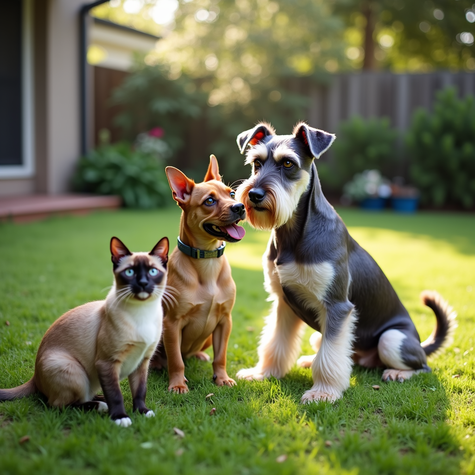}
        {\fail{} \quad \counttext{$\square$} \attrtext{$\blacksquare$} \spatialtext{$\blacksquare$}}
        \minipageimage
        {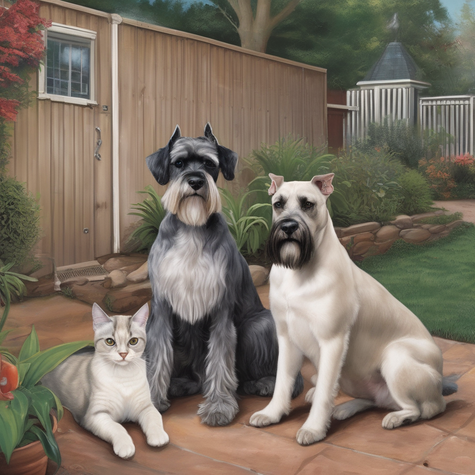}
        {\fail{} \quad \counttext{$\square$} \attrtext{$\square$} \spatialtext{$\square$}}
        \minipageimage
        {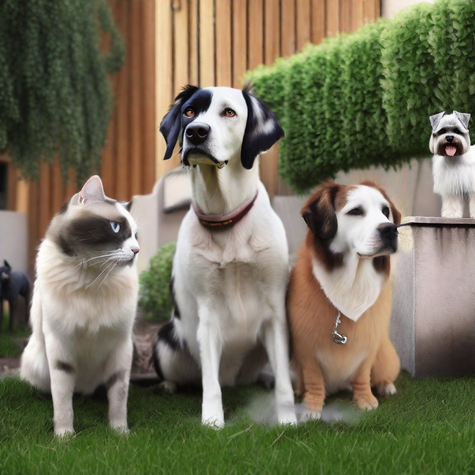}
        {\pass{} \quad \counttext{$\blacksquare$} \attrtext{$\blacksquare$} \spatialtext{$\blacksquare$}}
        \minipageimage
        {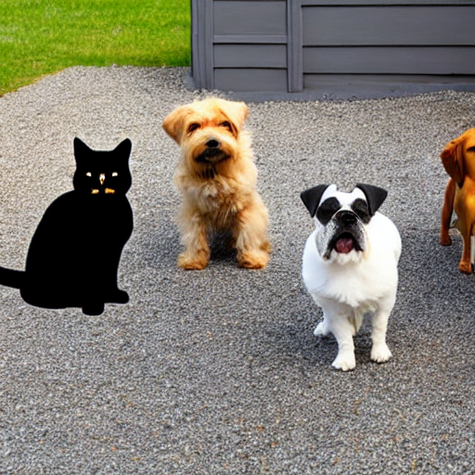}
        {\fail{} \quad \counttext{$\square$} \attrtext{$\square$} \spatialtext{$\square$}}
        \minipageimage
        {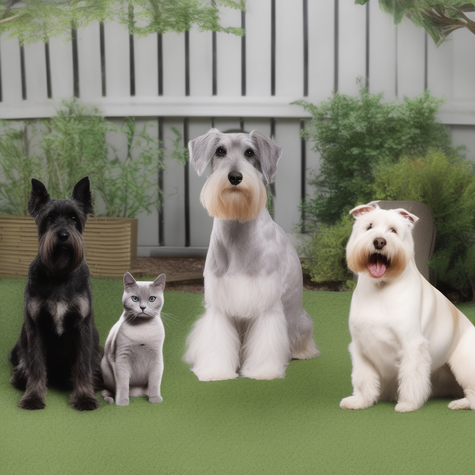}
        {\fail{} \quad \counttext{$\blacksquare$} \attrtext{$\square$} \spatialtext{$\square$}}
        \minipageimage
        {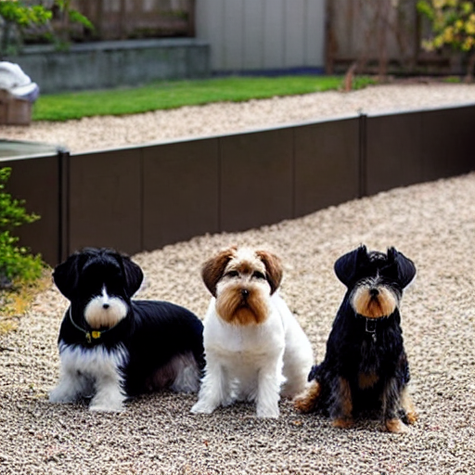}
        {\fail{} \quad \counttext{$\square$} \attrtext{$\square$} \spatialtext{$\square$}}

        \vspace{1mm}

        \input{figures/comparisons/extended_comparisons_method_titles}

\end{center}

    \caption{\textbf{Extended qualitative comparison.} Additional results for prompts sampled from \benchmarkName{}, along with VQA Acc. metric evaluation. For each tier, we consider an image \textit{correct} (\pass{}) if it receives positive answers from an MLLM (GPT4o) when prompted with all questions below its tier.
    We compare with Emu~\cite{dai2023emuenhancingimagegeneration}, Flux1-dev~\cite{flux2023}, SDXL~\cite{podell2023sdxlimprovinglatentdiffusion}, Bounded Attention (BA)~\cite{dahary2025yourself}, Attention Refocusing (GLIGEN + AR)~\cite{phung2024grounded}, and Reason Your Layout (RYL)~\cite{chen2023reason}.}
    \label{fig:compound_w_questions}
\end{figure*}

\begin{figure*}
    \setlength{\tabcolsep}{1.18pt}
    \begin{tabular}{ccccccc}
        \multicolumn{3}{c}{"\small{\counttext{four} action figures on a kid's shelf, from \spatialtext{left to right}}  }   
        &
        &
        \multicolumn{3}{c}{\small{"\counttext{three} people and a dog on the beach, the dog is a \attrtext{dalmatian},"}  }
        \\
        \multicolumn{3}{c}{\small{they are \attrtext{batman, superman, wolverine and spiderman"}}}
        &
        &
        \multicolumn{3}{c}{\small{and \attrtext{only} the person \spatialtext{next to it} is \attrtext{wearing a red hat"}} }\\
        \includegraphics[width=0.155\linewidth]{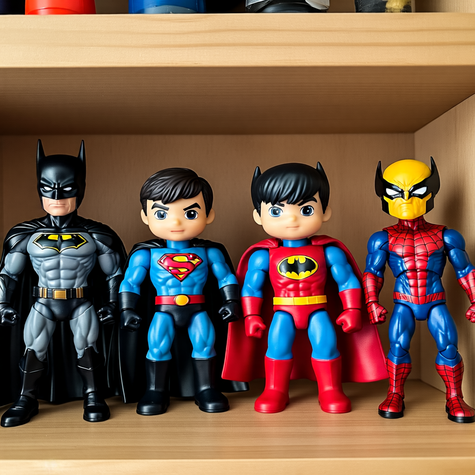}
        &
        \includegraphics[width=0.155\linewidth]{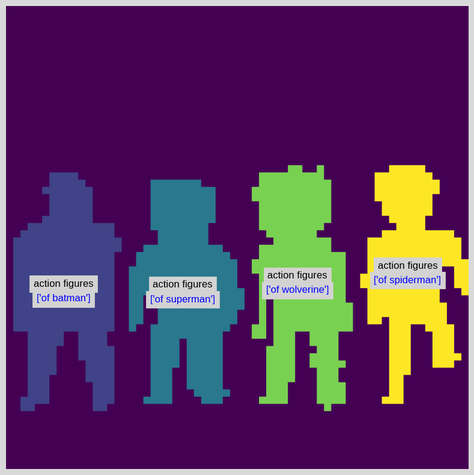}
        &
        \includegraphics[width=0.155\linewidth]{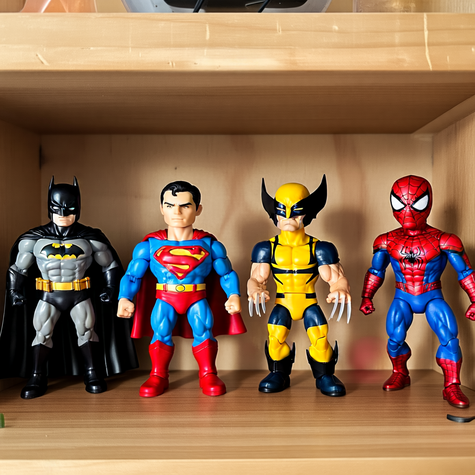}
        &
        &
        \includegraphics[width=0.155\linewidth]{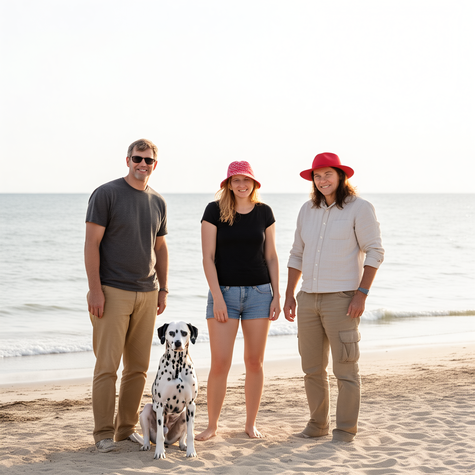}
        &
        \includegraphics[width=0.155\linewidth]{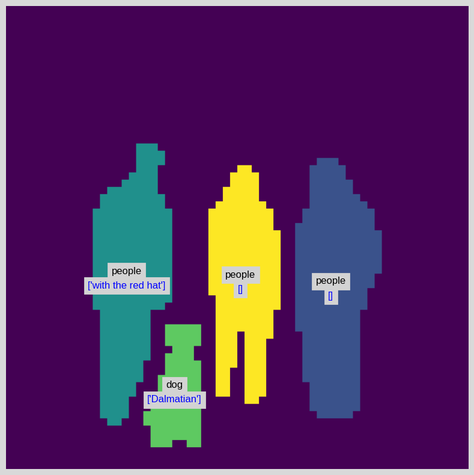}
        &
        \includegraphics[width=0.155\linewidth]{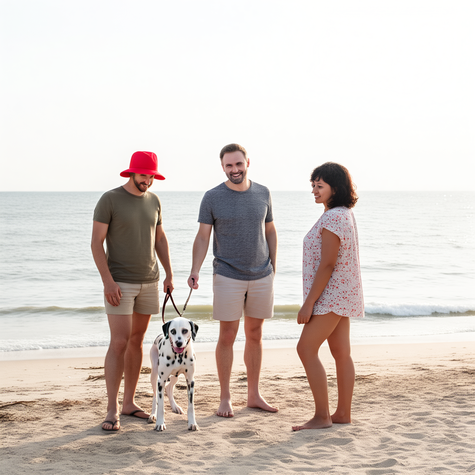}
        \\
        \footnotesize{Initial Image}
        &
        \footnotesize{Instance Assignments}
        &
        \footnotesize{\textbf{Ours}}
        &
        &
        \footnotesize{Initial Image}
        &
        \footnotesize{Instance Assignments}
        &
        \footnotesize{\textbf{Ours}}
        \vspace{2.0mm}
        \\

        \multicolumn{3}{c}{"\small{\counttext{three} people with \counttext{two} motorcycles on a rainy night, the one \spatialtext{on the}}}    
        &
        &
        \multicolumn{3}{c}{"\small{\counttext{five} birds on a tree branch in autumn, \attrtext{one} bird is a \attrtext{cardinal},} } 
        \\
        \multicolumn{3}{c}{\small{\spatialtext{right} is a \attrtext{blue chopper}, the one \spatialtext{on the left} is a \attrtext{yellow sportsbike}"}}
        &
        &
        \multicolumn{3}{c}{\small{\spatialtext{to the left of it} is a \attrtext{blue jay}, the rest are \attrtext{ravens}"}} \\
        \includegraphics[width=0.155\linewidth]{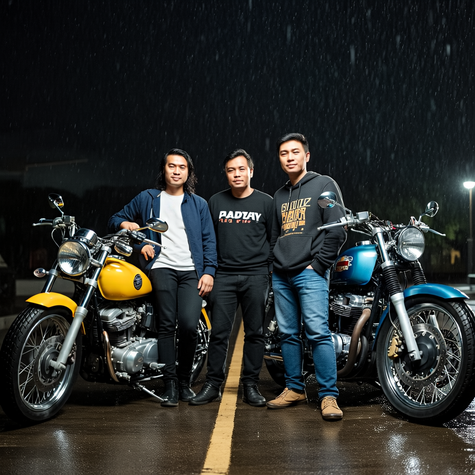}
        &
        \includegraphics[width=0.155\linewidth]{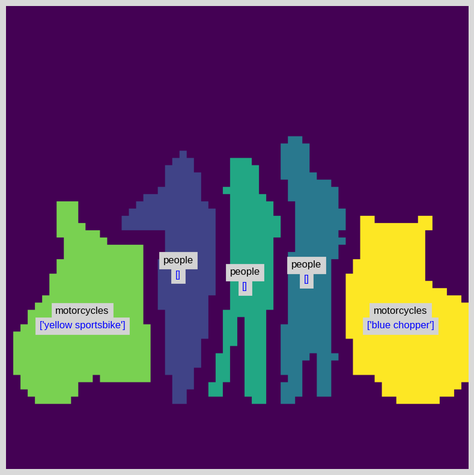}
        &
        \includegraphics[width=0.155\linewidth]{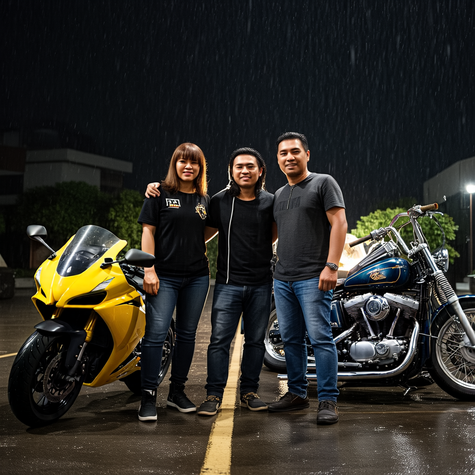}
        &
        &
        \includegraphics[width=0.155\linewidth]{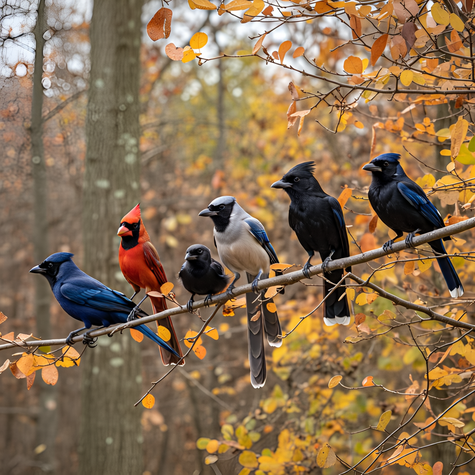}
        &
        \includegraphics[width=0.155\linewidth]{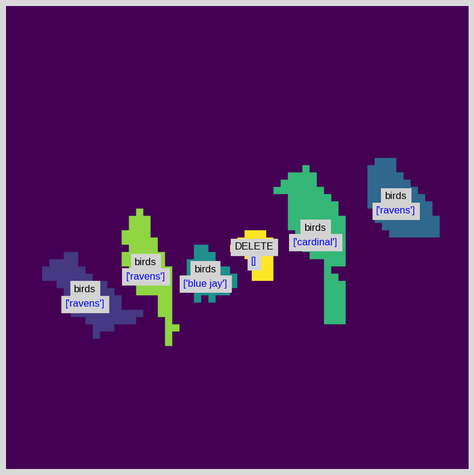}
        &
        \includegraphics[width=0.155\linewidth]{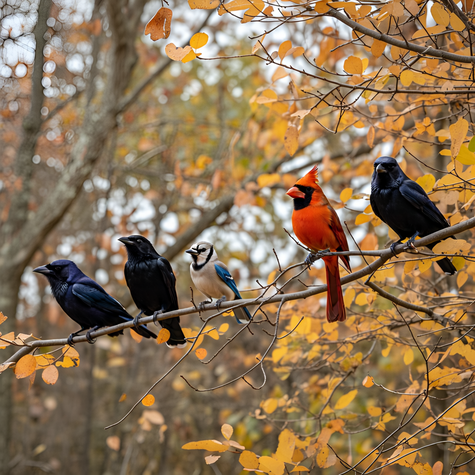}
        \\
        \footnotesize{Initial Image}
        &
        \footnotesize{Instance Assignments}
        &
        \footnotesize{\textbf{Ours}}
        &
        &
        \footnotesize{Initial Image}
        &
        \footnotesize{Instance Assignments}
        &
        \footnotesize{\textbf{Ours}}
        \vspace{2.0mm}
        \\

        \multicolumn{3}{c}{\small{"\counttext{four} pillows stacked on top of eachother, from \spatialtext{bottom to top}}   }  
        &
        &
        \multicolumn{3}{c}{\small{"\counttext{two} guitars and a saxophone in a music store, the guitar \spatialtext{on the}}  }
        \\
        \multicolumn{3}{c}{\small{they are \attrtext{pink, tiger pattern, covered in blue velvet and pinstriped"}}}
        &
        &
        \multicolumn{3}{c}{\small{\spatialtext{right} is a \attrtext{white fender}, the one \spatialtext{on the left} is a \attrtext{rickenbacker bass"}}} \\
        \includegraphics[width=0.155\linewidth]{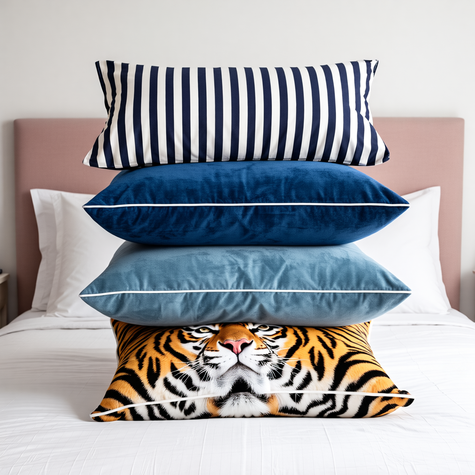}
        &
        \includegraphics[width=0.155\linewidth]{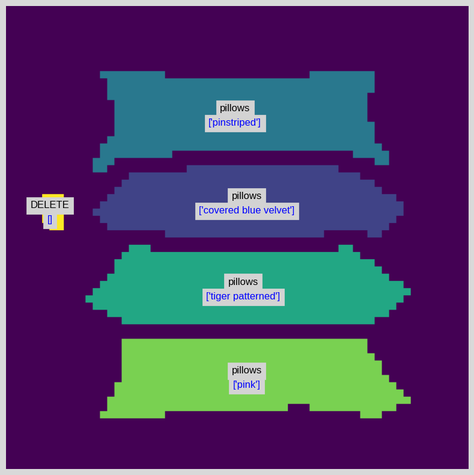}
        &
        \includegraphics[width=0.155\linewidth]{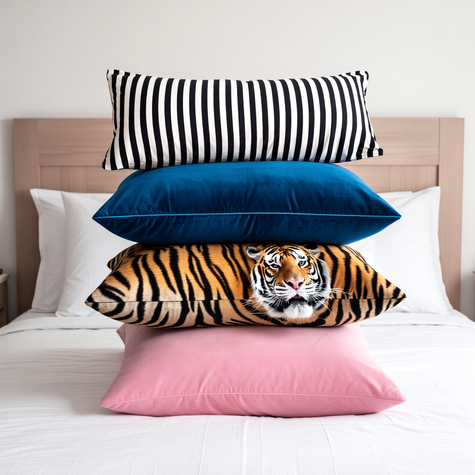}
        &
        &
        \includegraphics[width=0.155\linewidth]{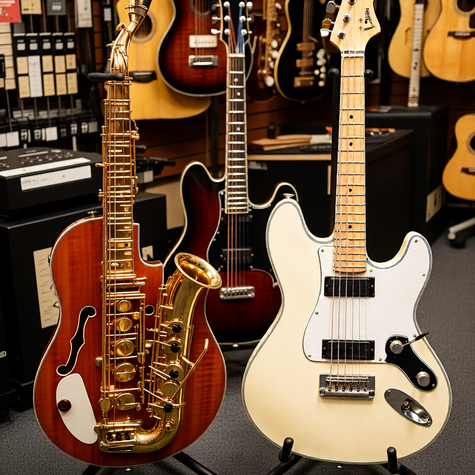}
        &
        \includegraphics[width=0.155\linewidth]{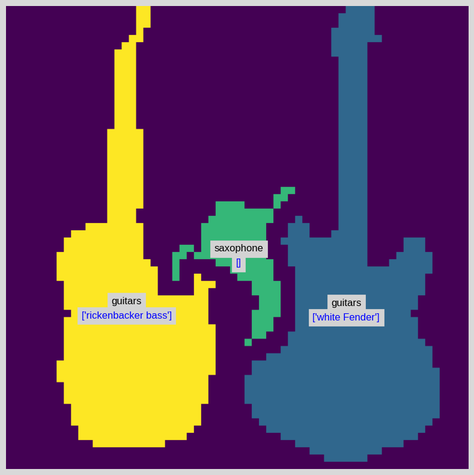}
        &
        \includegraphics[width=0.155\linewidth]{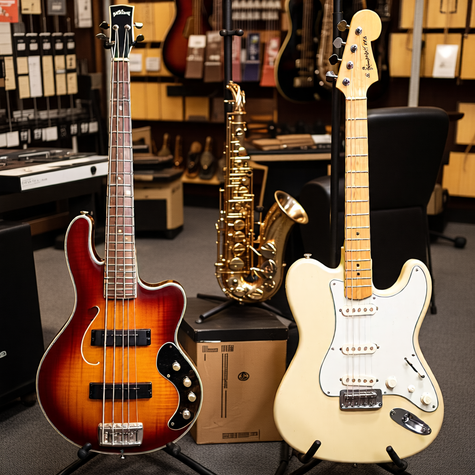}
        \\
        \footnotesize{Initial Image}
        &
        \footnotesize{Instance Assignments}
        &
        \footnotesize{\textbf{Ours}}
        &
        &
        \footnotesize{Initial Image}
        &
        \footnotesize{Instance Assignments}
        &
        \footnotesize{\textbf{Ours}}
        \vspace{2.0mm}
        \\

        \multicolumn{3}{c}{\small{"\counttext{four} pints of beer lined up at a bar, from \spatialtext{right to left} they are}    } 
        &
        &
        \multicolumn{3}{c}{\small{"\counttext{seven} crayons in a box, from \spatialtext{left to right} their colors are}  }
        \\
        \multicolumn{3}{c}{\small{\attrtext{a dark stout, a bright lager, a cloudy wheatbeer and an amber ale"}}}
        &
        &
        \multicolumn{3}{c}{\small{\attrtext{red, orange, yellow, green, blue, purple and black"}}} \\
        \includegraphics[width=0.155\linewidth]{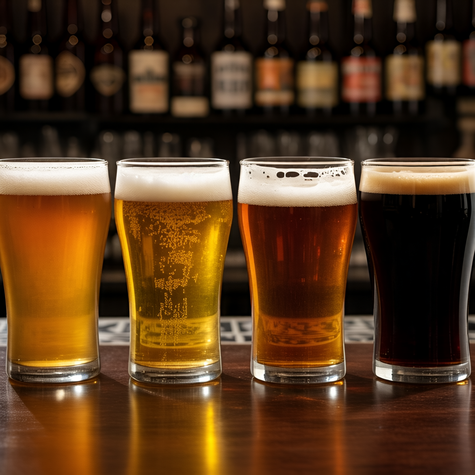}
        &
        \includegraphics[width=0.155\linewidth]{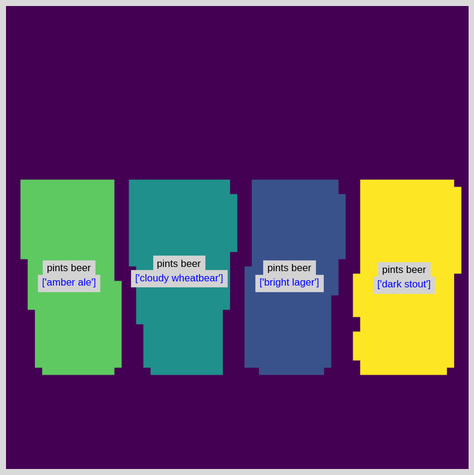}
        &
        \includegraphics[width=0.155\linewidth]{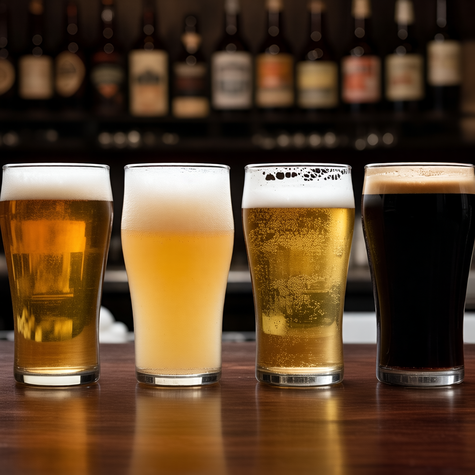}
        &
        &
        \includegraphics[width=0.155\linewidth]{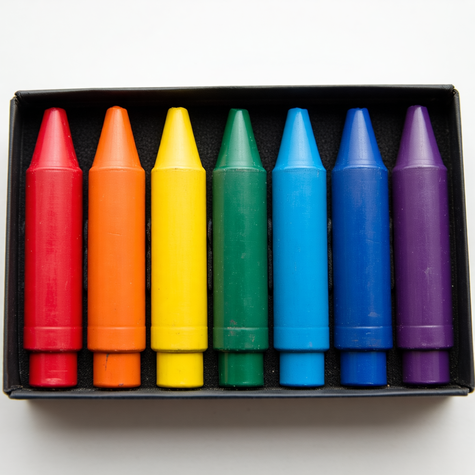}
        &
        \includegraphics[width=0.155\linewidth]{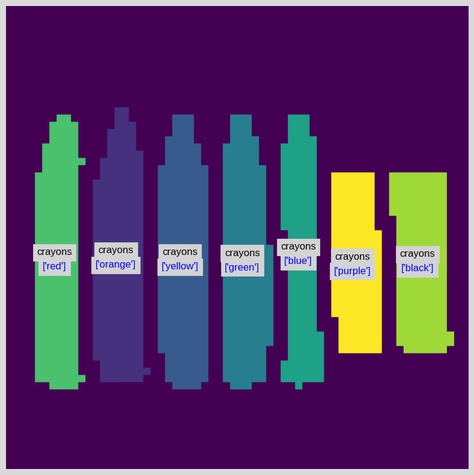}
        &
        \includegraphics[width=0.155\linewidth]{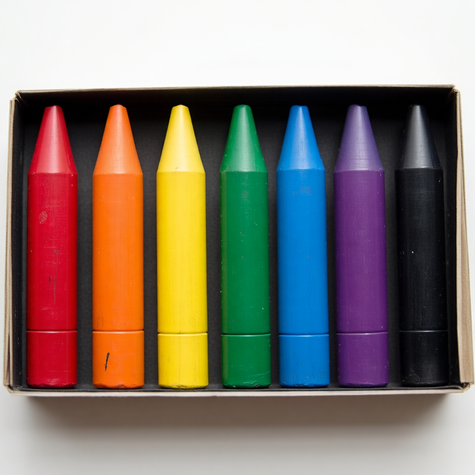}
        \\
        \footnotesize{Initial Image}
        &
        \footnotesize{Instance Assignments}
        &
        \footnotesize{\textbf{Ours}}
        &
        &
        \footnotesize{Initial Image}
        &
        \footnotesize{Instance Assignments}
        &
        \footnotesize{\textbf{Ours}}
        \vspace{2.0mm}
        \\

        \multicolumn{3}{c}{\small{"\counttext{five} ballerinas performing at a studio, the \spatialtext{second ballerina from}}}     
        &
        &
        \multicolumn{3}{c}{\small{"\counttext{a} porcupine, \counttext{a} raccoon and \counttext{a}  squirrel in the forest,}}  
        \\
        \multicolumn{3}{c}{\small{\spatialtext{the left} is \attrtext{wearing black and smiling}, the rest are \attrtext{wearing white}"}}
        &
        &
        \multicolumn{3}{c}{\small{the squirrel is \spatialtext{in the middle} \attrtext{holding a nut"}}} \\
        \includegraphics[width=0.155\linewidth]{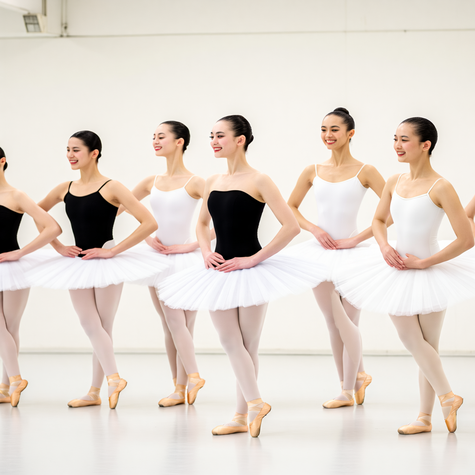}
        &
        \includegraphics[width=0.155\linewidth]{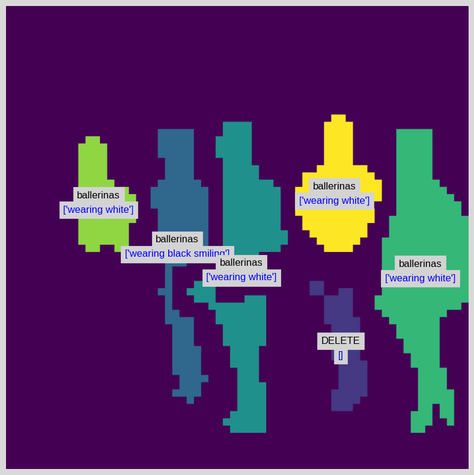}
        &
        \includegraphics[width=0.155\linewidth]{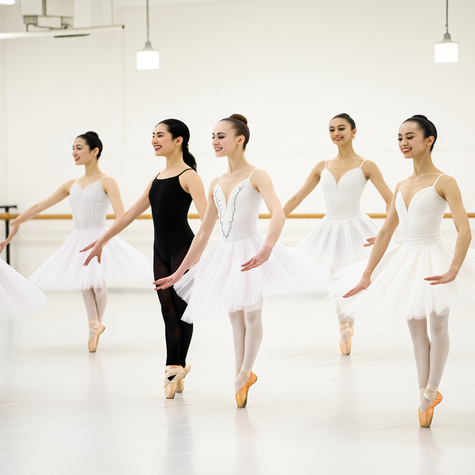}
        &
        &
        \includegraphics[width=0.155\linewidth]{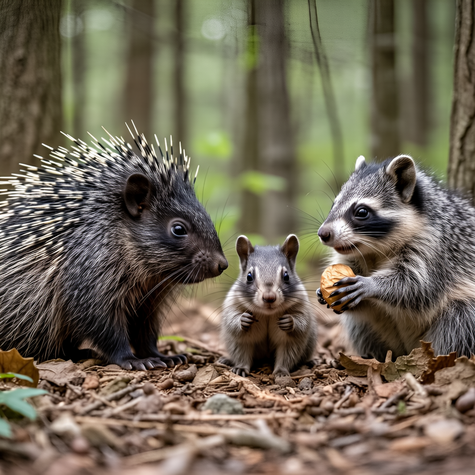}
        &
        \includegraphics[width=0.155\linewidth]{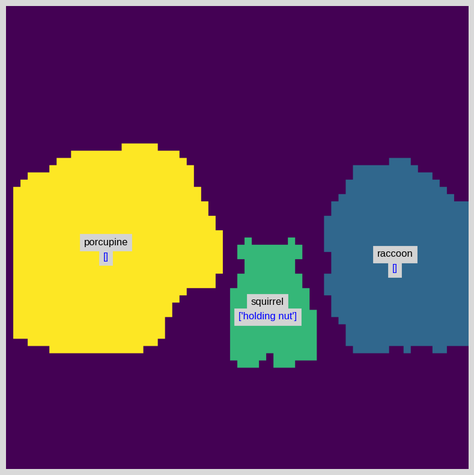}
        &
        \includegraphics[width=0.155\linewidth]{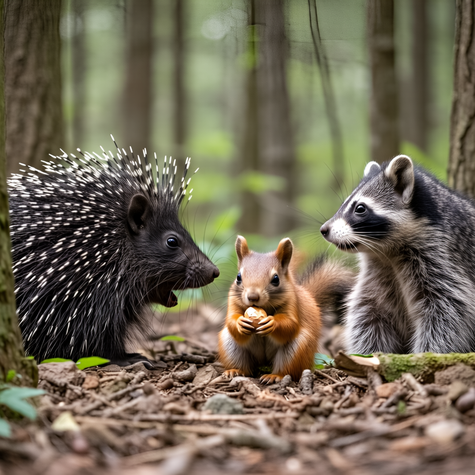}
        \\
        \footnotesize{Initial Image}
        &
        \footnotesize{Instance Assignments}
        &
        \footnotesize{\textbf{Ours}}
        &
        &
        \footnotesize{Initial Image}
        &
        \footnotesize{Instance Assignments}
        &
        \footnotesize{\textbf{Ours}}
        \vspace{-2.0mm}
        \\
    \end{tabular}
    \caption{\textbf{Assorted Tier C prompt results:} Initial Images shown here are post our robust initialization stage.}
    \label{fig:results}
\end{figure*}

\clearpage

\appendix
\noindent {\LARGE\textbf{\methodName{} Supplementary Material}}
\section{Additional Results and Information}
We refer readers to the ``supplementary results'' folder available alongside this document for additional results. This folder contains output images for all $150$ prompts available in the \benchmarkName{} dataset for our method as well as competing baselines. Also available alongside this document are the instruction prompts given to the LLM in the prompt parsing stage (``\texttt{prompt\_parser\_instruct- ion.txt}'') and the instance assignment stage (``\texttt{instance\_assignm- ent\_instruction.txt}'').

\section{Technical Details}
\label{sec:details}

\subsection{\methodName{} Implementation Details}
\label{sec:imp_details}

\subsubsection{Base Text to Image model}
The baseline text-to-image model used for both generating an initial image and generating the instance assignment guided output image is the latest version of Emu \cite{dai2023emuenhancingimagegeneration}. This particular model is a DiT \cite{peebles2023scalablediffusionmodelstransformers} based latent diffusion model, made up of 22 DiT layers and generating images in $1024 \times 1024$ resolution with attention dimensions $64 \times 64$. In all our experiments, we use this model with $T = 26$ inference steps and a classifier-free-guidance scale of $4.0$.

\subsubsection{Prompt Parsing}
As described in Section 3.1.1 of the main paper, in this step we task an LLM (Llama 3.3) with parsing a prompt into its key components in terms of subject instances. Specifically, the LLM is instructed to return a \texttt{json} format dictionary that lists all objects described in the prompt (and not describing the environment or setting), the amount of times they are required to appear and any instance attributes they might require with desired quantities for those as well. Below we present an example parser output for the prompt ``a porcupine, one squirrel and a raccoon in a forest, the squirrel is holding a nut''.

\begin{verbatim}
{
  "prompt": "a porcupine, one squirrel and a raccoon in 
  a forest, the squirrel is holding a nut",
  "objects": {
    "porcupine": {
      "desired_quantity": "1",
      "instance_adjectives": {}
    },
    "squirrel": {
      "desired_quantity": "1",
      "instance_adjectives": {
        "1": {
          "adjective": "holding nut",
          "desired amount": "1"
        }
      }
    },
    "raccoon": {
      "desired_quantity": "1",
      "instance_adjectives": {}
    }
  }
}
\end{verbatim}

Notice that the words ``nut'' or ``forest'' while certainly describing objects in the image are not treated as such by the parser as they either belong to an instance attribute (i.e. ``nut'' belongs to the ``holding a nut'' attribute) or a description of the environment (i.e. ``forest'' describing the setting and not an actual object instance).

\subsubsection{Instance Layout Generation}
In this section we provide details for the anchor point extraction and layout segmentation.

\medskip \noindent \textbf{Anchor Point Extraction.} \quad
To extract salient anchor points from the cross-attention maps, we aggregate the maps from various layers, timesteps, %
and text tokens in order to obtain a single cross-attention map for each object word. Specifically, we average the attention maps over timesteps [0-25] and layers [2-20] for each text token associated with object word $w$.
Then, we take the maximum
over all tokens associated with the word $w$ for each pixel,
resulting in a single aggregated cross attention map $C_w$. 

We extract anchor points from the cross attention map for each object word by
obtaining a foreground mask $M$ using the Otsu threshold,
$M_i = C_{w} > Otsu(C_{w})$~\cite{otsu1975threshold}. Then we compute local maxima of $C_w$ by running \texttt{skimage.feature.peak\_local\_max} and filter out the points that fall outside of the foreground mask $M_i$, to obtain anchor points $\{a_i\}_w$ associated with word $w$. 

\medskip \noindent \textbf{Layout Segmentation.} \quad
Given the initial image and the extracted anchor points, we produce the instance segmentation using a combination of an object instance segmentation model, Mask R-CNN~\cite{he2017mask}), and the more general segmentation model, SAM~\cite{kirillov2023segment}. 
First, we run Mask R-CNN on the initial image to obtain an initial set of instance segmentation masks $M$. Then, for each anchor point $a_{i,w}$ we find the the smallest mask $m \in M$ such that $a_{i,w}$ resides in $m$. If no such mask exists, we mark $a_{i,w}$ as unresolved. We discard masks that are not associated with any anchor points at this stage. Then, we extract additional segmentation masks for each unresolved point using SAM, by providing the unresolved point as an input positive coordinate. If there are significant overlaps, we merge the masks from the two phases. 
We end this process once each anchor point is associated with a segmentation mask. Note that multiple anchor points can be associated with a single mask.

\medskip \noindent \textbf{Segmentation Implementation Details.} \quad
For the cross-attention aggregation, we use $L_{agg} = [2, 20]$ and $T_{agg} = [0, 25]$ in all of our experiments.
For M-RCNN,
we use \href{https://github.com/facebookresearch/Detectron}{\texttt{Detectron}} with configuration \texttt{COCO-InstanceSegmentation/mask\textunderscore rcnn R\textunderscore 50\textunderscore FPN\textunderscore 1x.yaml}. \\
This segmentation method sometimes outputs a mask that is a union of multiple smaller masks and covers a vast region of the image. To avoid using these masks in our instance layout we filter out masks which cover more than $0.33\%$ of the total image area. After filtering out masks that are not assigned to any anchor point,
we are left with a set of masks that are not guaranteed to be non-overlapping. To mitigate this, we check for overlapping between every pair of masks and assign the overlapping region to the smaller mask between the two. 

Next, we
go over each unassigned anchor point and extract a segmentation mask with SAM, using the anchor point coordinate as a positive point. Specifically, we use SAM2 available \href{https://github.com/facebookresearch/sam2}{\emph{here}}. These segments are more prone to overlap as that is the desirable scenario for multiple anchor points belonging to the same instance. As such, we again go over each mask pair and look for overlaps. However, this time we check if the overlap between two masks cover $66.6\%$ of each masks area, if it does we merge the two masks, if it doesen't we assign the overlap region to the smaller mask.
After extracting this initial layout, we run a simple post processing steps in which we filter out masks that are smaller than $30$ pixels, and create a minimal margin of $2$ pixels between adjacent masks. 

\medskip \noindent \textbf{Instance Copying.} \quad
As described in the main paper, if a viable layout is not obtained after a set amount of seed search iterations due to under-generation of instances,
we copy existing instances from the mask and place them in the background in a random manner to guide the image generation.
We first extract a background mask by excluding all detected instances and passing it through 2 steps of binary erosion with a $3\times3$ kernel.
Then, we select a random existing instance mask and a random location withing the background mask, and create a new instance mask which has the same shape as the randomly selected instance, centered around the selected location in the background.
If there are overlaps between the new instance mask and the originals we remove the overlapping region from the new mask. We then run the postprocessing steps used in the original layout generation stage to ensure a $2$ pixel margin between the new mask and the originals. We repeat this process until there are enough instance masks to facilitate the prompt. To prevent a bias towards copying previously copied instance masks we sample only from the original instance masks when selecting instances for copying.

\medskip \noindent \textbf{\texttt{json} Output.} \quad
After post processing, we calculate attention scores for each object word and attribute word and summarize them all into a \texttt{json} format file. Below we present an example of this segmentation summary for a layout obtained for the above prompt:

\begin{verbatim}
{
  "1": {
    "cluster_size": "392.0",
    "distance_from_top": "34.3",
    "distance_from_left": "55.8",
    "object_probabilities": {
      "porcupine": "0.11",
      "squirrel": "0.14",
      "raccoon": "0.75"
    },
    "attribute_probabilities": {
      "holding nut": "0.068"
    }
  },
  "2": {
    "cluster_size": "225.0",
    "distance_from_top": "42.8",
    "distance_from_left": "40.7",
    "object_probabilities": {
      "porcupine": "0.21",
      "squirrel": "0.53",
      "raccoon": "0.26"
    },
    "attribute_probabilities": {
      "holding nut": "0.128"
    }
  },
  "3": {
    "cluster_size": "745.0",
    "distance_from_top": "34.3",
    "distance_from_left": "15.1",
    "object_probabilities": {
      "porcupine": "0.71",
      "squirrel": "0.14",
      "raccoon": "0.15"
    },
    "attribute_probabilities": {
      "holding nut": "0.066"
    }
  }
}
\end{verbatim}

\medskip \noindent \textbf{Instance Assignment}.
As described in Section 3.2 of the main paper, in this stage we use an LLM (Llama 3.3) to assign each segment in the extracted layout an instance instruction. The LLM receives an input text for completion consisting of an instruction, in-context examples, the parsed input prompt and the layout summary from the previous stage. We instruct the LLM to work according to the following steps:
\begin{enumerate}
    \item Figure out how many objects need to be set and how many need to be deleted.
    \item Figure out if objects require spatial arrangement.
    \item If an object does not require spatial arrangement, set it according to its probability (i.e. set it as the object with the highest 'object\textunderscore probabilities' value) until all required objects are set.
    \item After setting objects move on to setting attributes following the same logic.
    \item Go over your decisions and ensure you have set the correct amount of objects and attributes according to the parsed prompt.
\end{enumerate}
The LLM is instructed to output its instructions in \texttt{json} format and also provide a paragraph detailing the reasoning for its decisions, in accordance with the previously explained steps. An example of these outputs for the previously used prompt is given below:

\begin{verbatim}
** REASONING **
1. There are 3 total clusters in the layout dict,we are 
required to assign one cluster as 'porcupine', one clus-
ter as 'squirrel' and onecluster as 'raccoon' meaning we 
do not have to delete any clusters.
2. The 'porcupine' object does not require spatial arra-
ngment. The 'squirrel' object also does not require spa-
tial arrangment. The 'raccoon' object also does not req-
uire spatial arrangment.
3. 'porcupine' has the lowest desired quantity (1) so we 
start by assiging it to the cluster with the highest 
'porcupine' probability, which is cluster "3" with 'por-
cupine' probability (0.71). We then assign the 'raccoon' 
object to the cluster with the highest 'raccoon' probab-
ility, which is cluster "1" with 'raccoon' probability 
(0.75). This leaves us with cluster "2" which we mark as 
'squirrel'.
4.a. The 'squirrel' object requires one instances to be 
set as 'holding nut'.
4.b. The 'holding nut' adjective does not require spat-
ial arrangement.
4.c. We set cluster "2" as 'holding nut' because it has 
the highest 'holding nut' probability (0.128) out of 
the clusters we marked as 'squirrel'.
5. We have set one cluster as 'porcupine' ("3") as req-
uired. We have set one cluster as 'squirrel' ("2") as 
required. We have set one cluster as 'raccoon' ("1") as 
required. We have set one 'squirrel' cluster as 'holdi-
ng nut' ("2") as required.

** ASSIGNMENTS **
{
  "1": {
    "object": "raccoon",
    "attributes": []
  },
  "2": {
    "object": "squirrel",
    "attributes": [
      "holding nut"
    ]
  },
  "3": {
    "object": "porcupine",
    "attributes": []
  }
}
\end{verbatim}

As detailed in the main paper if this assignment is erroneous according to the parsed prompt (for instance if the attribute ``holding nut'' was assigned to a segment that has been assigned the ``racoon'' object, or if too many segments have been assigned the ``squirrel'' object word - as per the previous prompt), we append both the reasoning and assignment to the LLM's instruction text and generate another output.

\medskip \noindent \textbf{Assignment Conditioned Image Generation}.
As described in Section 3.3 of the main paper in this stage we generate an image that follows the assignemnts produced in the previous stage. The loss weights used in this stage are $1.0$ for $\mathcal{L}_{obj}$, $0.8$ for $\mathcal{L}_{att}$ and $0.3$ for $\mathcal{L}_{bg}$. These losses were used to optimize the latents from timestep $t_{start}=0$ to timestep $t_{end}=20$, with $15$ optimization iterations added to timesteps $t = 0$ and $t = 5$, and $5$ optimization iterations added to timestep $t = 10$. We use a fixed learning rate of $0.015$. Self attention masking was done from timestep $t=0$ to $t=3$ on layers $l=10$ to $l=21$. Cross attention masking was done from timestep $t = 0$ to timestep $t = 22$on layers $l=0$ to $l=21$. From end to end this stage takes roughly $3$ minutes on a single \texttt{NVIDIA H100 80GB} GPU.

\subsection{\benchmarkName{} Breakdown}

\benchmarkName{} is made up of $60$ unique prompts, each describing multiple objects in various settings. Each unique prompt has three versions, as described in the main paper, which include object counts (Tier A), instance-level attributes (Tier B), and spatial relationships (Tier C). In addition, we augment each of the three prompts with a lower count version (2-3 objects), a medium count version (4-5 objects), and a higher count version (6-7 objects). In total we have $9$ versions for each prompt or a total of $540$ prompts.
Prompts with different counts describe the same scene, for example, the prompt ``an image of \emph{two} dogs and \emph{three} cats, the cat on the far right is a sphinx'' corresponds to the lower count prompt ``an image of \emph{a} dog and \emph{two} cats, the cat on the far right is a sphinx''.

\subsection{Experimentation Details}
\label{sec:exp_details}

\subsubsection{Benchmarks}
\hfill\\
\noindent \textbf{Evaluation on DrawBench}.
To evaluate on this dataset, we first obtained the counting and spatial prompts available \href{https://huggingface.co/datasets/shunk031/DrawBench}{here}. Then we used the evaluation code in the ``Attention-Refocusing'' \cite{phung2024grounded} \href{https://github.com/Attention-Refocusing/attention-refocusing}{github repository}.

\noindent \textbf{Evaluation on \benchmarkName{}}.
As detailed in the main paper (Section 4.3) we used two different metrics when evaluating on \benchmarkName{}: VQA Sim and VQA Acc. To calculate VQA Sim we used the API available on the ``Evaluating text-to-visual generation with image-to-text generation'' \href{https://github.com/linzhiqiu/t2v_metrics}{github repository}. To calculate VQA Acc we used GPT4o \cite{openai2024gpt4technicalreport} to answer the questions associated with each prompt in the \benchmarkName{} benchmark and using the evaluation protocol detailed in the main paper.

\subsubsection{Baselines}
\hfill\\
\noindent \textbf{SDXL}.
We used the \texttt{stabilityai/stable-diffusion-xl-base-1.0} configuration for the SDXL pipeline available on \href{https://huggingface.co/stabilityai/stable-diffusion-xl-base-1.0}{huggingface}.

\medskip \noindent \textbf{Flux 1 dev}.
We used the default \texttt{black-forest-labs/FLUX.1-dev} configuration for the Flux 1 dev pipeline also available on \href{https://huggingface.co/black-forest-labs/FLUX.1-dev}{huggingface}.

\medskip \noindent \textbf{GLIGEN + Attention Refocusing}.
We used the default inference procedure available on the ``attention-refocusing'' \href{https://github.com/Attention-Refocusing/attention-refocusing}{github repository}, using GPT4o to generate layouts for the \benchmarkName{} prompts using the instruction also available in the repository. The GLIGEN \cite{li2023gligen} models were obtained via \href{https://huggingface.co/gligen/gligen-generation-text-box/blob/main/diffusion_pytorch_model.bin}{this link}.

\medskip \noindent \textbf{Bounded-Attention}.
We used the default SDXL based configuration available on the ``Bounded-Attention'' \href{https://github.com/omer11a/bounded-attention}{repository}. When evaluating this method we used the layouts generated for the ``GLIGEN + AR'' baseline.

\medskip \noindent \textbf{CountGen}.
We used the default inference configuration available on the ``CountGen'' \href{https://github.com/Litalby1/make-it-count}{github repository}.

\medskip \noindent \textbf{Reason-your-Layout}.
We used the default inference configuration available on the ``Reason-your-Layout'' \href{https://github.com/Xiaohui9607/LLM_layout_generator}{github repository}. Layouts were generated using GPT4o using the instruction also available on the repository.

\medskip \noindent {\textbf{Direct Preference Optimization (DPO)}.
We used the SDXL DPO weights available on \href{https://huggingface.co/mhdang/dpo-sdxl-text2image-v1}{huggingface}.

\medskip \noindent \textbf{Self-correcting
LLM-controlled Diffusion (SLD)}.
We used the default ``Image Correction'' configuration on images generated with SDXL using the GPT4o LLM. The code used to run this baseline is available on the SLD \href{https://github.com/tsunghan-wu/SLD}{github repository}.

\section{Further Discussion}

\begin{figure}
\begin{center}

\begin{minipage}{\linewidth}
	\centering

    \small{\emph{``a tray with \counttext{three} muffins, 
                    \counttext{two} vanilla éclairs, 
                    and 
                    \counttext{a} piece of cake on it, the
                    \spatialtext{top left} muffin is \attrtext{blueberry},
                    the other \attrtext{two} are \attrtext{chocolate}.''}}

	\begin{minipage}[t]{0.19\linewidth}
	    \centering
	    \includegraphics[width=\linewidth]{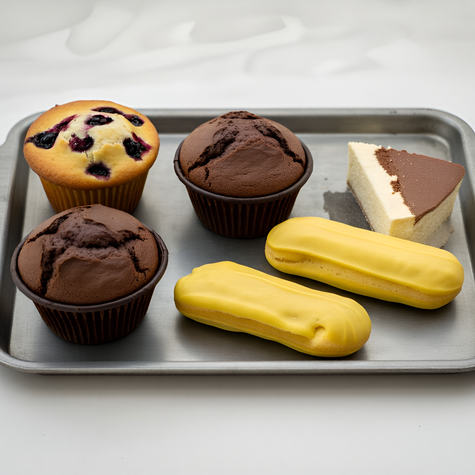}
	\end{minipage}
	\begin{minipage}[t]{0.19\linewidth}
	    \centering
            \includegraphics[width=\linewidth]{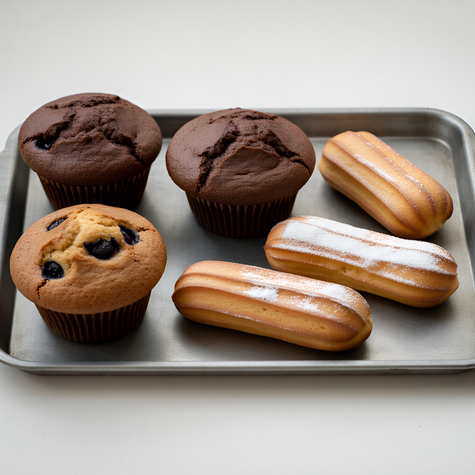}
	\end{minipage}
    \vrule
    \hspace{0.001\linewidth}
	\begin{minipage}[t]{0.19\linewidth}
	    \centering
	    \includegraphics[width=\linewidth]{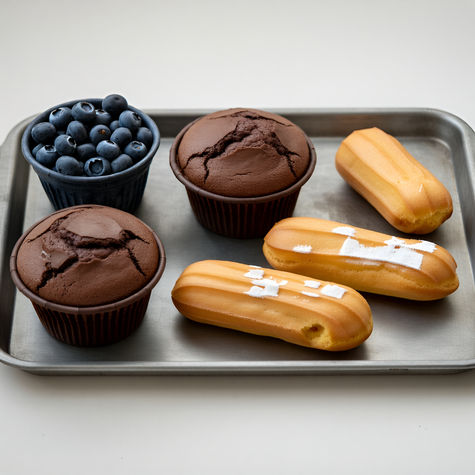}
	\end{minipage}
	\begin{minipage}[t]{0.19\linewidth}
	    \centering
	    \includegraphics[width=\linewidth]{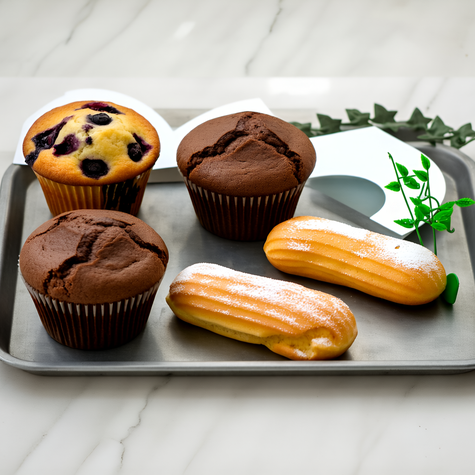}
	\end{minipage}
	\begin{minipage}[t]{0.19\linewidth}

            \centering
	    \includegraphics[width=\linewidth]{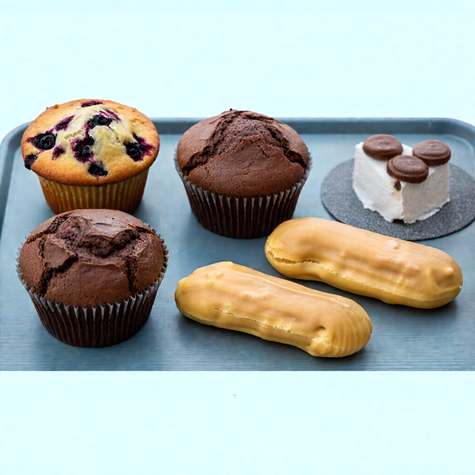}
	\end{minipage}

        \small{\emph{``\counttext{five} birds on a tree branch in autumn,  \attrtext{one} bird is a \attrtext{cardinal}, another is a \attrtext{blue jay},  the rest are \attrtext{ravens}.''}}
                    
    	\begin{minipage}[t]{0.19\linewidth}
	    \centering
	    \includegraphics[width=\linewidth]{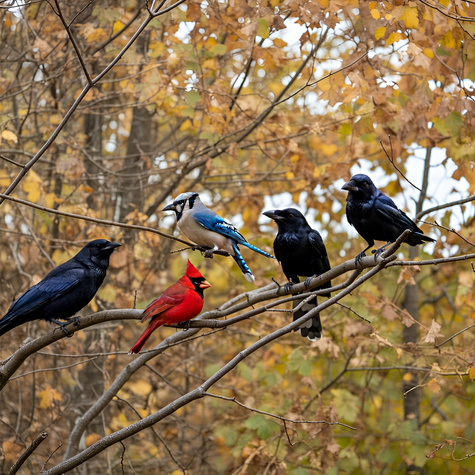}
	\end{minipage}
	\begin{minipage}[t]{0.19\linewidth}
	    \centering
            \includegraphics[width=\linewidth]{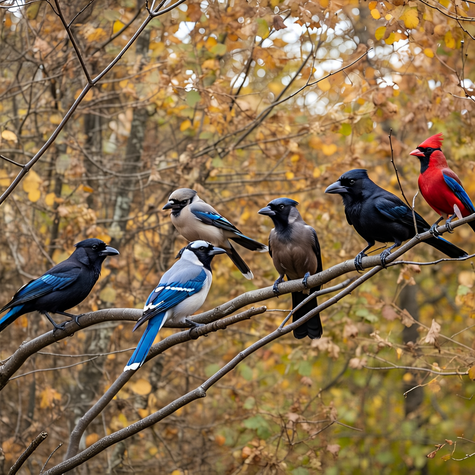}
	\end{minipage}
    \vrule
    \hspace{0.001\linewidth}
	\begin{minipage}[t]{0.19\linewidth}
	    \centering
	    \includegraphics[width=\linewidth]{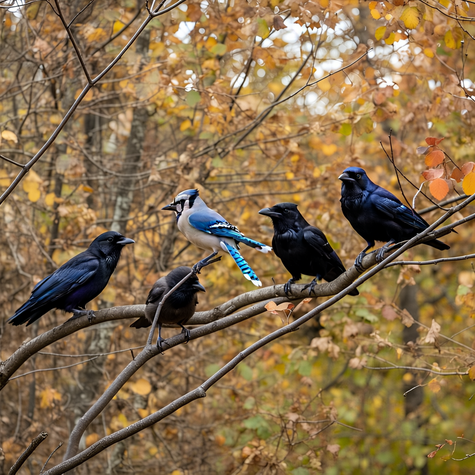}
	\end{minipage}
	\begin{minipage}[t]{0.19\linewidth}
	    \centering
	    \includegraphics[width=\linewidth]{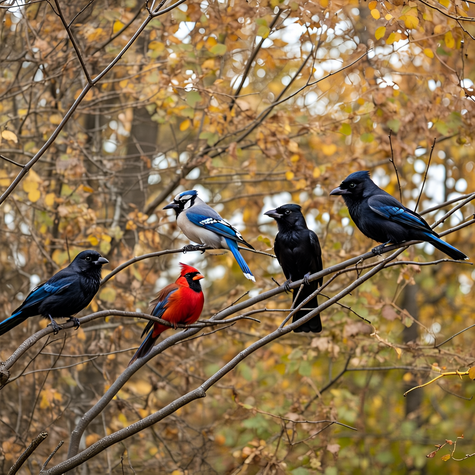}
	\end{minipage}
    \hspace{-4.2px}
	\begin{minipage}[t]{0.19\linewidth}

            \centering
	    \includegraphics[width=\linewidth]{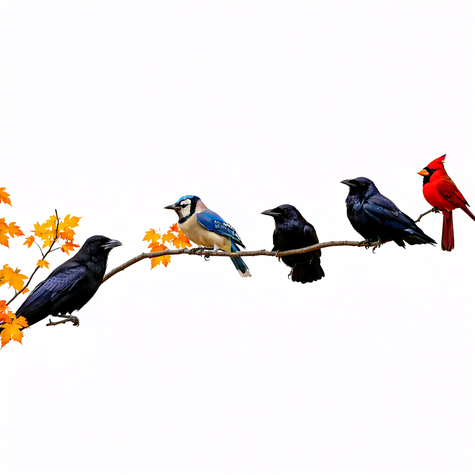}
	\end{minipage}

        \small{\emph{``\counttext{four} cupcakes and \counttext{two} pies on a     kitchen counter, the cupcake \spatialtext{in the center} has \attrtext{rainbow  sprinkles}.''}}
    
                	\begin{minipage}[t]{0.19\linewidth}
	    \centering
	    \includegraphics[width=\linewidth]{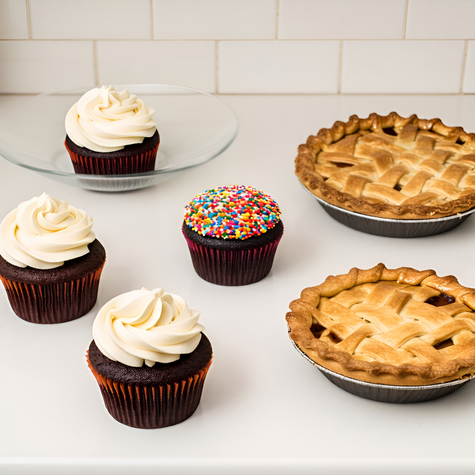}
	\end{minipage}
	\begin{minipage}[t]{0.19\linewidth}
	    \centering
            \includegraphics[width=\linewidth]{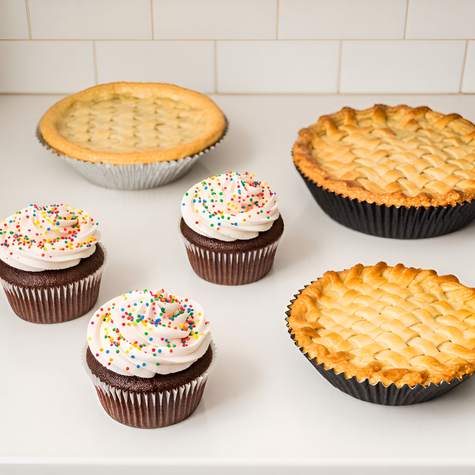}
	\end{minipage}
    \vrule
    \hspace{0.001\linewidth}
	\begin{minipage}[t]{0.19\linewidth}
	    \centering
	    \includegraphics[width=\linewidth]{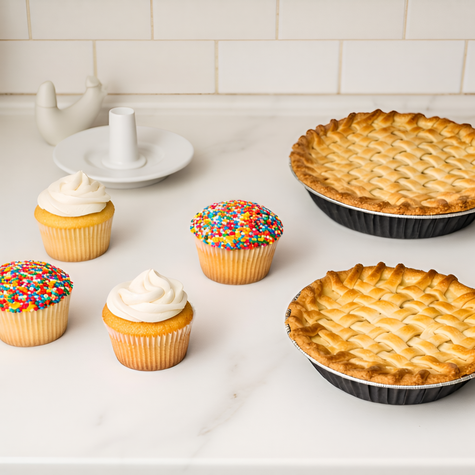}
	\end{minipage}
	\begin{minipage}[t]{0.19\linewidth}
	    \centering
	    \includegraphics[width=\linewidth]{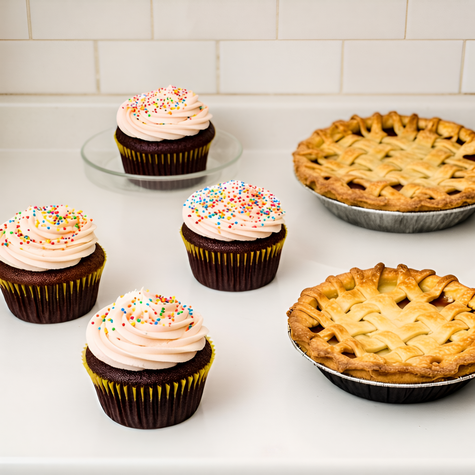}
	\end{minipage}
    \hspace{-4.2px}
	\begin{minipage}[t]{0.19\linewidth}

            \centering
	    \includegraphics[width=\linewidth]{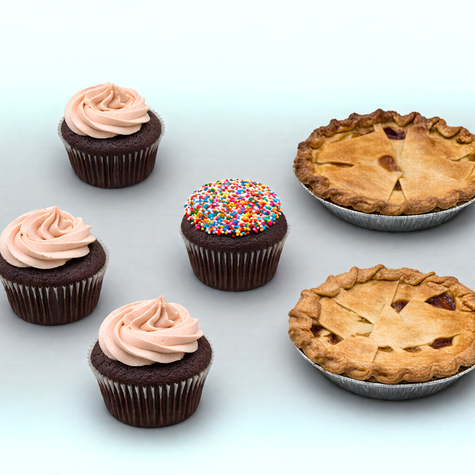}
	\end{minipage}

	\begin{minipage}{0.19\linewidth}
	    \centering
        \textbf{Ours}

	\end{minipage}
	\begin{minipage}{0.19\linewidth}
	    \centering
            Emu 1.6

	\end{minipage}
	\begin{minipage}{0.19\linewidth}
	    \centering
            w/o $\mathcal{L}_{obj}$ \& $\mathcal{L}_{att}$

	\end{minipage}
	\begin{minipage}{0.19\linewidth}
	    \centering
            w/o Attn. Masking

	\end{minipage}
    \hspace{-4.2px}
	\begin{minipage}{0.19\linewidth}
	    \centering
            w/o $\mathcal{L}_{bg}$

	\end{minipage}

\end{minipage}
\caption{\textbf{Qualitative ablations.} We present compare results produced by our method when ablating three different components: cross attention losses (``w/o $\mathcal{L}_{obj}$ \& $\mathcal{L}_{att}$''), attention masking (``w/o Attn. Masking'') and composition preserving regularization (``w/o $\mathcal{L}_{bg}$'')}

   \label{fig:ablations}

\vspace{-10pt}
\end{center}
\end{figure}

\subsection{Qualitative Ablation Study}
A qualitative ablation study accompanying the quantitative study presented in Section 4.4 is presented in Figure \ref{fig:ablations}. The figure showcases the types of failures that occur when different components are ablated from our method. Without attention losses (middle column) our method often fails to make meaningful changes to the initial image in the areas where they are required. For instance failing to turn one of the eclairs to a piece of cake (top row, middle column) or turning one of the ravens into cardinals (middle row, middle column). Without attention masking objects are often less distinct and semantic leakage is evident. This can be seen in the blue feathers that leak out into the ravens and cardinal (middle row, second column from the right) and sprinkles that appear on all four cupcakes (bottom row, second column from the right). The effect of removing the composition preserving regularization loss from our method (rightmost column) is perhaps most visually distinct, producing outputs that are much less visually pleasing and often seemingly ignore the parts of the input prompt that describe the environment (rightmost column, bottom and middle rows).

\subsection{Method Runtimes}
\begin{table}
    \centering
    \setlength{\tabcolsep}{0.008\textwidth}
    \begin{tabular}{l c}
        \toprule
        Method & Run Time \\
        \midrule
        SDXL &  \textbf{$\sim \text{7 seconds}$} \\
        Flux1-dev & $\sim \text{50 seconds}$ \\
        CountGen &  $\sim \text{2 minutes}$ \\
        Reason-your-Layout &  $\sim \text{15 seconds}$ \\
        BoundedAttention &  $\sim \text{10 minutes}$ \\
        AttentionRefocusing &  $\sim \text{40 seconds}$ \\
        DPO &  \textbf{$\sim \text{7 seconds}$} \\
        SLD &  $\sim \text{2 minutes}$ \\
        Emu  & $\sim \text{45 seconds}$ \\
        Ours &  $\sim \text{5 minutes}$ \\
        \bottomrule
    \end{tabular}
    \captionof{table}{
    Running times for all methods tested in this work.
    }
    \vspace{-20pt}
    \label{table:runtimes}
\end{table}

Running times for our method as well as competing methods that were compared against in this work are presented in Table \ref{table:runtimes}. All running times are calculated on a single NVIDIA A100 80GB GPU. As this table clearly shows, the general purpose image generation methods (SDXL, Emu, Flux1-dev) as well as DPO (which functions exactly like SDXL in inference) are fastest with SDXL and DPO being the fastest overall with a run time of $\sim \text{7 seconds}$. The optimization based methods are expectedly slower with Bounded-Attention being slowest with a runtime of $\sim \text{10 minutes}$.

\subsection{Extended Evaluation on \benchmarkName{}}

\begin{figure}
\begin{center}

\begin{minipage}{\linewidth}
	\centering

    \normalsize{\emph{``\counttext{four} pillows stacked on top of each other, from \spatialtext{bottom to top} they are \attrtext{pink, tiger pattern, covered in blue velvet and pinstriped}.''}}

	\begin{minipage}[t]{0.30\linewidth}
	    \centering
	    \includegraphics[width=\linewidth]{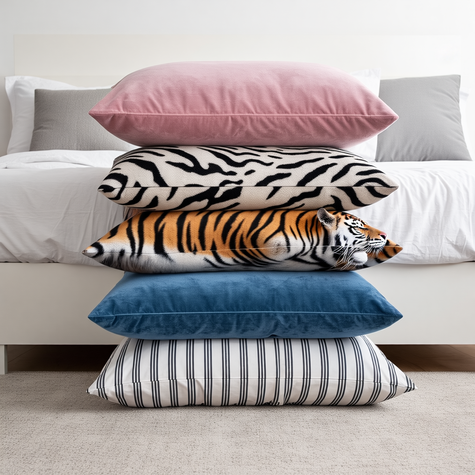}
	\end{minipage}
	\begin{minipage}[t]{0.30\linewidth}
	    \centering
            \includegraphics[width=\linewidth]{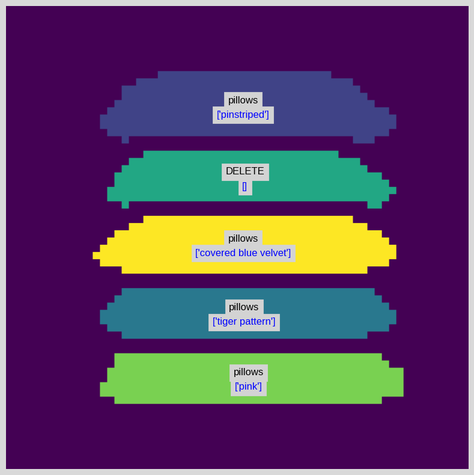}
	\end{minipage}
	\begin{minipage}[t]{0.30\linewidth}
	    \centering
	    \includegraphics[width=\linewidth]{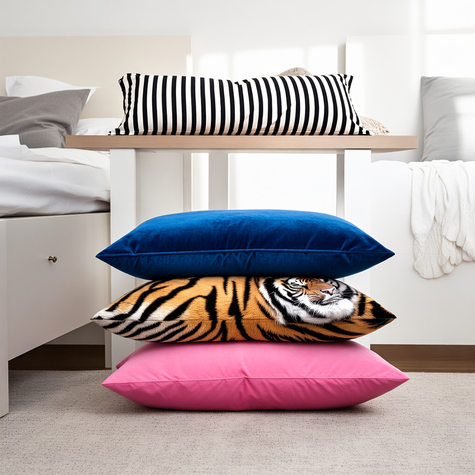}
	\end{minipage}

        \normalsize{\emph{``\counttext{four} cats and \counttext{a} dog in a living room, \attrtext{one} cat is \attrtext{ginger} another is a \attrtext{babydoll} the dog is \spatialtext{sitting between them} and is a \attrtext{poodle}.''}}
                    
    	\begin{minipage}[t]{0.30\linewidth}
	    \centering
	    \includegraphics[width=\linewidth]{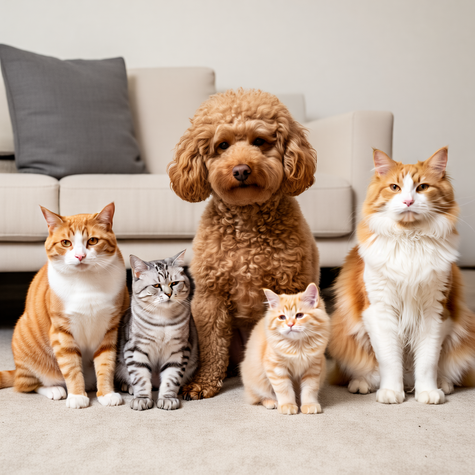}
	\end{minipage}
	\begin{minipage}[t]{0.30\linewidth}
	    \centering
            \includegraphics[width=\linewidth]{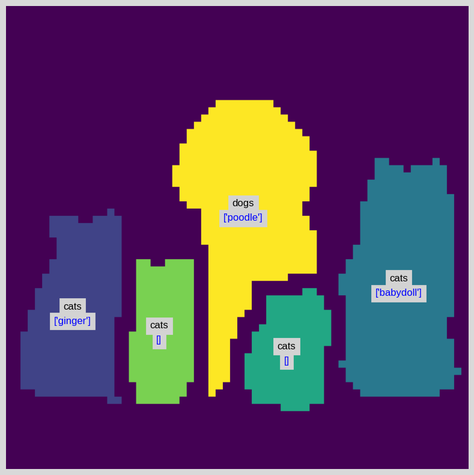}
	\end{minipage}
	\begin{minipage}[t]{0.30\linewidth}
	    \centering
	    \includegraphics[width=\linewidth]{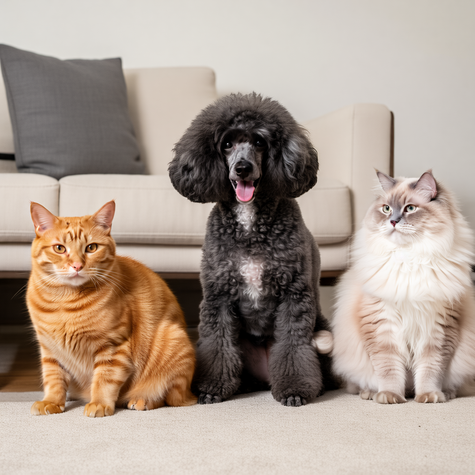}
	\end{minipage}

        \normalsize{\emph{``\counttext{four} balls in a playing field, \spatialtext{in front} is a \attrtext{volleyball}, \spatialtext{in the back} is a \attrtext{black basketball}, a \attrtext{soccer ball} is \spatialtext{on each side}, \attrtext{one} of which is \attrtext{golden}.''}}
    
        \begin{minipage}[t]{0.30\linewidth}
	    \centering
	    \includegraphics[width=\linewidth]{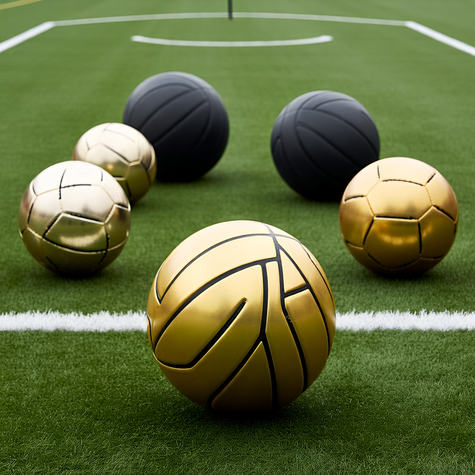}
	\end{minipage}
	\begin{minipage}[t]{0.30\linewidth}
	    \centering
            \includegraphics[width=\linewidth]{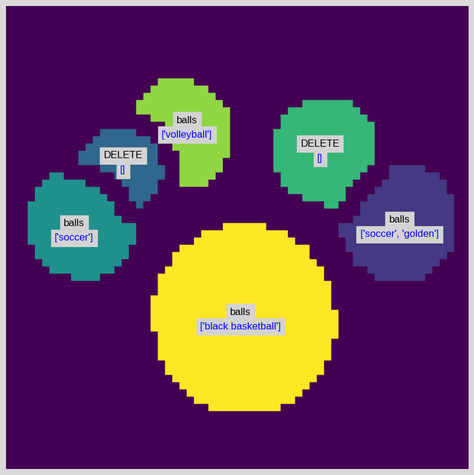}
	\end{minipage}
	\begin{minipage}[t]{0.30\linewidth}
	    \centering
	    \includegraphics[width=\linewidth]{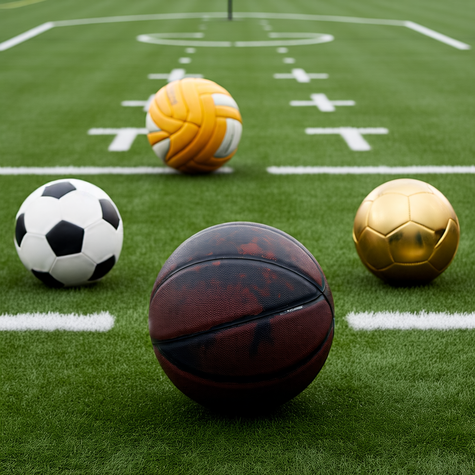}
	\end{minipage}

	\begin{minipage}{0.30\linewidth}
	    \centering
        Initial Image

	\end{minipage}
	\begin{minipage}{0.30\linewidth}
	    \centering
        Instance Assignments

	\end{minipage}
	\begin{minipage}{0.30\linewidth}
	    \centering
        \textbf{Ours}

	\end{minipage}

\end{minipage}
\caption{\textbf{Limitations.} We present three failure cases of our method. In the first example (top row) our method wrongfully creates a gap in the pillow stack. In the second (middle row), two pairs of cats are merged into one. In the final example (bottom row) our wrongfully places the volleyball in the back and the black basketball in the front. See Section \ref{sec:limitations} for an additional discussion over these failure cases.}

    \label{fig:limitations}
    
    \vspace{-20pt}
\end{center}
\end{figure}

\begin{table*}
    \centering
    \begin{tabular}{l | cccc | cccc | cccc | c}
        \toprule
        \multirow{3}{25mm}{Method} & \multicolumn{13}{c}{VQA Accuracy} \\
        \cmidrule{2-14}
        & & \multicolumn{2}{c}{2-3 Objects} & &  & \multicolumn{2}{c}{4-5 Objects} & & & \multicolumn{2}{c}{6-7 Objects} & & \multirow{2}{8mm}{Avg.} \\
        & A & B & C & Avg. & A & B & C & Avg. & A & B & C & Avg. & \\
        \midrule
        SDXL & 0.53 & 0.28 & 0.15 & 0.32 & 0.30 & 0.12 & 0.10 & 0.17 & 0.42 & 0.14 & 0.07 & 0.21 & 0.23 \\ 
        Flux1-dev & 0.68 & 0.53 & 0.53 & 0.58 & 0.65 & 0.37 & 0.42 & 0.48 & 0.42 & 0.20 & 0.20 & 0.27 & 0.44 \\ 
        CountGen & 0.55 & 0.28 & 0.17 & 0.33 & 0.42 & 0.13 & 0.07 & 0.21 & 0.37 & 0.13 & 0.07 & 0.19 & 0.24 \\ 
        ReasonYourLayout & 0.40 & 0.02 & 0.13 & 0.24 & 0.28 & 0.05 & 0.07 & 0.13 & 0.27 & 0.05 & 0.00 & 0.11 & 0.16 \\ 
        BoundedAttention & 0.45 & 0.18 & 0.23 & 0.29 & 0.35 & 0.13 & 0.03 & 0.17 & 0.37 & 0.11 & 0.05 & 0.18 & 0.21 \\ 
        AttentionRefocusing & 0.68 & 0.28 & 0.15 & 0.37 & 0.63 & 0.07 & 0.03 & 0.24 & 0.57 & 0.04 & 0.02 & 0.21 & 0.27 \\
        \esc{DPO} & 0.45 & 0.23 & 0.12 & 0.27 & 0.25 & 0.15 & 0.02 & 0.14 & 0.30 & 0.08 & 0.07 & 0.15 & 0.19 \\ 
        \esc{SLD} & 0.62 & 0.27 & 0.13 & 0.34 & 0.37 & 0.07 & 0.03 & 0.16 & 0.33 & 0.08 & 0.03 & 0.15 & 0.21 \\ 
        Emu & 0.78 & 0.60 & 0.52 & 0.63 & 0.40 & 0.32 & 0.22 & 0.31 & \textbf{0.58} & 0.28 & 0.17 & 0.34 & 0.43 \\ 
        Ours & \textbf{0.87} & \textbf{0.75} & \textbf{0.68} & \textbf{0.77} & \textbf{0.72} & \textbf{0.58} & \textbf{0.45} & \textbf{0.58} & \textbf{0.58} & \textbf{0.38} & \textbf{0.37} & \textbf{0.44} & \textbf{0.60} \\ 
        \bottomrule
    \end{tabular}
    \caption{Detailed quantitative evaluation
    on the \benchmarkName{} dataset.}
    \label{table:extended_cp}
    \vspace{-10pt}
\end{table*}

\begin{table*}
    \centering
    \begin{tabular}{l c c c c c c c}
        \toprule
        Method & Single Object & Colors & Counting & Position & Color Attr.  & Two Object & Overall \\
        \midrule
        SDXL & 0.98 & 0.85 & 0.39 & 0.15 & 0.23 & 0.74 & 0.55 \\
        Flux1-dev & \textbf{1.00} & 0.82 & \textbf{0.71} & 0.19 & 0.47 & 0.81 & 0.67 \\
        CountGen & 0.96 & 0.83 & 0.56 & 0.09 & 0.20 & 0.61 & 0.54 \\
        ReasonYourLayout & 0.95 & 0.64 & 0.44 & 0.34 & 0.07 & 0.55 & 0.50 \\
        BoundedAttention & 0.98 & 0.85 & 0.30 & 0.48 & 0.21 & 0.75 & 0.59 \\
        AttentionRefocusing & 0.97 & 0.66 & 0.62 & 0.58 & 0.15 & 0.73 & 0.66 \\
        DPO & \textbf{1.00} & \textbf{0.87} & 0.41 & 0.11 & 0.24 & 0.81 & 0.57 \\
        SLD & 0.88 & 0.80 & 0.51 & 0.24 & 0.18 & 0.70 & 0.55 \\
        Emu & 0.97 & 0.85 & 0.55 & 0.51 & 0.62 & 0.85 & 0.73 \\
        Ours & 0.97 & 0.85 & 0.68 & \textbf{0.64} & \textbf{0.73} & \textbf{0.87} & \textbf{0.79} \\
        \bottomrule
    \end{tabular}
    \caption{Quantitative evaluation
    on the GenEval benchmark.}
    \label{tab:geneval}
    \vspace{-10pt}
\end{table*}

Figure \ref{fig:compound_w_questions} in the main paper presents a comparison against all competing baselines which were evaluated in this work. The figure shows the results for two unique prompts 
in their three complexity versions (Tiers A, B and C). This figure also illustrates our evaluation process, specifically in calculating VQA Accuracy, displaying the prompts along with their accompanying questions and the answers each result received for it as evaluated by an MLLM (GPT4o \cite{openai2024gpt4technicalreport}). While several methods can generate object counts correctly (particularly for lower counts), they rarely satisfy all of the conditions of the prompts in tiers B and C.

In Table \ref{table:extended_cp} we present the results of a quantitative comparison against competing baselines on the \benchmarkName{} dataset, specifically presenting the VQA Accuracy metric across all difficulty tiers and object counts. While these results, somewhat unsurprisingly, show a general trend of VQA Accuracy being lower as the number of objects in the images increases, this is not always the case. SDXL, Emu and BoundedAttention for instance, achieve higher VQA Accuracy scores for images with 6-7 objects in comparison to images with 4-5 objects. The reason for this is not completely clear, but we hypothesize that this can be attributed to these specific baselines having a slight tendency to overgenerate.
Regardless, the results show that our method outperforms competing baselines across all difficulty tiers and object count tiers.

\subsection{Evaluation on GenEval}

In Table \ref{tab:geneval} we present the results of a quantitative comparison conducted on the GenEval \cite{ghosh2024geneval} benchmark. This benchmark is made up of $553$ prompts aimed at evaluating various compositional image properties. Specifically, these prompts are split into six categories: \textit{Counting, Colors, Position, Color Attribution, Single Object} and \textit{Two Object}. Most relevant for our problem setting are the \textit{Counting, Position, Color Attribution} and \textit{Two Object} prompts as they describe images with more than one object. Similarly to DrawBench, the evaluation process in GenEval uses a semantic segmentation model (Mask2Former \cite{cheng2022masked}) as well as a VLM (CLIP ViT-L/14 \cite{radford2021learning}) in order to compare the count, positioning, object type and color of instances in the image against a ground truth label. As the results show, our method outperforms competing baselines in all the multi-object categories apart from the \textit{Counting} category, in which we are slightly outperformed by Flux1-dev even though our method outperforms our baseline method (Emu) by a significant $13\%$. Also worth mentioning is the fact that the \textit{Counting} prompts in this benchmark involve only a single object appearing multiple times, this is in contrast to the other baselines used in the paper - DrawBench and \benchmarkName{} in which \textit{Counting} (or Type A prompts in \benchmarkName{}) can involve multiple types of objects in the same image. These multi object prompts are difficult for competing methods, and our method significantly outperforms competing baselines on these more complex counting prompts.

\subsection{Limitations}
\label{sec:limitations}
In Figure \ref{fig:limitations} we present three typical failure cases of our method, demonstrating issues resulting from both the instance assignment stage (top and bottom rows) as well as the assignment conditioned image generation stage (middle row). In the first example (top row) our method fails to accurately convey the text prompt as while the object count and spatial ordering of the pillows is correct they are not stacked on top of each other as requested. In this specific scenario the LLM opted to delete the second pillow from the top (most likely due to it having a lower attention score for all the attribute words) resulting in a gap that it addressed by placing the top pillow further back on a shelf. We believe these types of errors are caused by the LLM focusing on following its instructions to the point of neglecting basic logic, and might be solvable with better instruction prompt tuning or more powerful language models. 

The second example (middle) row, shows a common failure where two (or more) instances merge into one. This is most common in two instances close to one another with both having been assigned the same object words or attributes. This shows that while mask segments are certainly more precise than bounding boxes, a guidance-based optimization approach still leaves room for such errors. %

In the last example (bottom row) our method wrongfully assigns ``volleyball'' and ``black basketball'', placing the volleyball in the back and the black basketball in front. This is caused by one of the more significant limitations of our method which is that the LLM in charge of instance assignment assumes instances are on a 2D plane and as a result is unable to differentiate ``front'' from ``back''. An interesting direction for future work is to address this limitation by incorporating depth estimation outputs into the instance assignment procedure.

Our method handles undergeneration of instances by \emph{seed search} and an \emph{instance copying} algorithm. While effective, these approaches have some limitations. 
Seed search is computationally demanding and is not guaranteed to produce a viable layout with enough object instances. Instance copying, on the other hand, is guaranteed to produce a viable layout, however this layout is usually less aligned with the original attention distribution, which sometimes results in inaccurate or low visual quality results. This could potentially be improved by incorporating attention based guidance into the instance copying mechanism, or using a specifically trained model to add missing instances to layout maps (similarly to CountGen). We leave these potential improvements for future work.

\end{document}